\documentclass{article}

     \PassOptionsToPackage{square,sort,comma,numbers}{natbib}

     \usepackage[final]{neurips_2018}

\usepackage[utf8]{inputenc} 
\usepackage[T1]{fontenc}    
\usepackage{hyperref}       
\usepackage{url}            
\usepackage{booktabs}       
\usepackage{amsfonts}       
\usepackage{nicefrac}       
\usepackage{microtype}      
\usepackage{enumitem}
\usepackage{amsmath}
\usepackage{graphicx}
\usepackage{multirow,bigdelim}
\usepackage{booktabs}
\usepackage{textcomp}
\usepackage{adjustbox}
\usepackage{verbatim}
\usepackage{enumitem}
\usepackage{caption}
\usepackage{subcaption}
\usepackage{mathtools}
\usepackage{commands}
\usepackage{amsmath,amsfonts,amssymb}
\usepackage{algorithm}
\usepackage[noend]{algpseudocode}
\usepackage{xifthen}
\usepackage{tikz}
\usetikzlibrary{graphs, graphs.standard, quotes}
\usepackage{todonotes}
\usepackage{wrapfig}

\newcommand{\vect}[1]{\boldsymbol{\mathrm{#1}}}
\newcommand{\vectall}[1]{\vect{#1}_{1:T}}

\title{EM-like Learning Chaotic Dynamics from Noisy and Partial Observations}

\author{%
  Duong Nguyen,  Said Ouala, Lucas Drumetz, Ronan Fablet\\
  IMT Atlantique, Lab-STICC, UBL, 29238 Brest, France\\
  \texttt{\{van.nguyen1, said.ouala, lucas.drumetz, ronan.fablet\}@imt-atlantique.fr} \\
}

\begin{document}

\maketitle

\begin{abstract}
The identification of the governing equations of chaotic dynamical systems from data has recently emerged as a hot topic. While the seminal work by Brunton et al. reported proof-of-concepts for idealized observation setting for fully-observed systems, {\em i.e.} large signal-to-noise ratios and high-frequency sampling of all system variables, we here address the learning of data-driven representations of chaotic dynamics for partially-observed systems, including significant noise patterns and possibly lower and irregular sampling setting. Instead of considering training losses based on short-term prediction error like state-of-the-art learning-based schemes, we adopt a Bayesian formulation and state this issue as a data assimilation problem with unknown model parameters. To solve for the joint inference of the hidden dynamics and of model parameters, we combine neural-network representations and state-of-the-art assimilation schemes. Using iterative Expectation-Maximization (EM)-like procedures, the key feature of the proposed inference schemes is the derivation of the posterior of the hidden dynamics. Using a neural-network-based Ordinary Differential Equation (ODE) representation of these dynamics, we investigate two strategies: their combination to Ensemble Kalman Smoothers and Long Short-Term Memory (LSTM)-based variational approximations of the posterior. Through numerical experiments on the Lorenz-63 system with different noise and time sampling settings, we demonstrate the ability of the proposed schemes to recover and reproduce the hidden chaotic dynamics, including their Lyapunov characteristic exponents, when classic machine learning approaches fail.  
\end{abstract}

\section{Introduction}
\label{sec:introduction}

Dynamical systems are at the core of numerous scientific fields, among which we may cite geosciences, aerodynamics, fluid dynamics, etc.
Classically, the determination of the governing laws of a given system, usually stated as Ordinary Dynamical Equations (ODE) or Partial Differential Equations (PDE) combines mathematical derivation based on physical principles and some experimental validations \cite{lorenz_deterministic_1963, hilborn_chaos_2000, sprott_chaos_2003, hirsch_differential_2012}, this approach forms the discipline of data assimilation. Recently, advances in machine learning \cite{jordan_machine_2015}, together with the availability of more and more data, open an appealing means for the data-driven identification of dynamical systems. However, learning dynamical systems is an ``extremely difficult task" \cite{brunton_discovering_2016}. Although machine learning in general, and deep learning in particular has achieved illustrious results in many domains \cite{lecun_deep_2015}, their performance in learning dynamical systems may remain limited as many dynamical systems are non-linear, chaotic and associated with noisy and partial observations in practice. 

In this paper, we propose a novel methodology, which combine the advances of state-of-the-art machine learning---neural networks---and classical data assimilation schemes for the problem of learning dynamical systems. The advantages of the proposed methodology are demonstrated by experiments on two chaotic Lorenz--63 system \cite{lorenz_deterministic_1963}.

The paper is organized as follows. In Section \ref{sec:dynamicalSystem}, we formulate the problem of learning non-linear dynamical systems. The detail of the proposed methodology and its two schmemes are presented in Section \ref{sec:methodology}. Section \ref{sec:experiments} present the experiments and results. Finally, we reviews the related work then close with a discussion in Section \ref{sec:discussion}.

\section{Data-driven identification of non-linear dynamical systems}
\label{sec:dynamicalSystem}

Following the pioneering work of Lorenz \cite{lorenz_deterministic_1963, lorenz_predictability:_1996}, the identification of data-driven representations of dynamical systems amounts to determine a data-driven computational representation to approximate a dynamical model typically stated as the solution of an ODE (Ordinary Differential Equation): 
\begin{align}
    \frac{d\vect{x}_{t}}{dt} &= \mathcal{F}\big(\vect{x}_{t}\big)+\eta_{t}
    \label{eq:ode}
\end{align}
where $\mathcal{F}$ is the unknown dynamical model of a multi-dimensional state $\vect{x}_{t}$ and $\eta_{t}$ is zero-mean additive noise due to neglected physics and/or numerical approximations. State-of-the-art data-driven schemes typically exploits observation data $\vect{y}_{t_i}$ as  
\begin{align}
    \vect{y}_{t_i} &= \Phi_{t_i}\big(\mathcal{H}\big(\vect{x}_{t_i}\big)+\epsilon_{t}\big)
    \label{eq:obs}
\end{align}
where $\mathcal{H}$ is the observation operator, usually known, $\epsilon_{t}$ is a zero-mean noise process and $\{t_i\}_i$ refers to the time sampling, typically a regular high-frequency sampling such that $t_i = t_0 +i.\delta $ with respect to a time resolution $\delta$ and reference starting time $t_0$. We introduce a masking operator $\Phi_{t_i}$ to account for observation $\vect{y}_{t_i}$ may not be available at each time step $t_i$ ($\Phi_{t_i,j} = 0$ if the $j^{th}$ component of $\vect{y}_{t_i}$ is missing). Noise processes $\eta_{t}$ and/or $\epsilon_{t}$ are generally assumed to be zero-mean Gaussian noise processes.


While data assimilation methods \cite{ghahramani_parameter_1996, ghahramani_learning_1998} address linear or linearizable systems only, recent advances in machine learning has proved that complex and highly non-linear systems can also be learned from ideal observations \cite{brunton_discovering_2016, raissi_multistep_2018, pathak_using_2017}, that is to say noise-free observations available at all high-frequency time steps. However, when significant noise patterns are presented and/or the observation is partial, these methods are likely to fail, as the minimization of the short-term prediction error based on the propagation of noisy inputs through operator ${\mathcal{F}}$ cannot be guaranteed to lead to the true dynamical operator. 

Following a classic state-space formulation, the identification of operator ${\mathcal{F}}$ from a series of observations ${\vect{y}_{t_i}}_i$ amounts to a joint identification of the hidden states ${\vect{x}_{t_i}}_i$ and of operator ${\mathcal{F}}$. The combination of assimilation methods for the identification issue and machine learning frameworks (especially, neural network architectures) for the parameterization of operator ${\mathcal{F}}$ naturally appears as an appealing solution. As detailed in the next Section, inspired by Bayesian formulation and associated Expectation-Maximization algorithm, we propose and evaluate two alternative schemes.


\section{Proposed Expectation-Maximization-like Approach}
\label{sec:methodology}

In this Section, we introduce the proposed schemes for the joint of the hidden states ${\vect{x}_{t_i}}_i$ and operator ${\mathcal{F}}$, using neural-network-based representation for the latter. Formally, we consider a discrete state-space formulation. It amounts to reformulating Eqs. \ref{eq:ode} and \ref{eq:obs} as follows:
\begin{align}
    \vect{x}_{t+\delta} &= f\big(\vect{x}_{t}\big) + \omega_{t+\delta}
    \label{eq:dyn_x} \\
    \vect{y}_{t} &= \Phi_{t}\big(\mathcal{H}\big(\vect{x}_{t}\big) + \epsilon_{t}\big)
    \label{eq:obs_y}
\end{align}
Where $\vect{x}_{t+\delta}$ results from an integration of operator $\mathcal{F}$ from state $\vect{x}_{t}$: $f\big(\vect{x}_{t}\big) = \vect{x}_{t} + \int_{t}^{t+1}\mathcal{F}\big(\vect{x}_{t}\big)dt$. $\omega_{t}$ and $\epsilon_{t}$ represent the uncertainty of the model and the observation, respectively. For the sake of simplicity, $\delta$ is arbitrarily set to 1 without any loss of generality. Within this Bayesian setting, 
Eqs. \ref{eq:dyn_x} and \ref{eq:obs_y} relate respectively to the dynamical prior $p\big(\vect{x}_{t+1}|\vect{x}_{t}\big)$ and the observation likelihood $p\big(\vect{y}_{t}|\vect{x}_{t}\big)$. Under Gaussian assumption for noise processes,  $\omega_{t}$ and $\epsilon_{t}$, $p\big(\vect{x}_{t+1}|\vect{x}_{t}\big)$ and $p\big(\vect{y}_{t}|\vect{x}_{t}\big)$ are Gaussian distributions.


The Expectation-Maximization (EM) algorithm \cite{bishop_pattern_2006} is an iterative procedure to estimate model parameters, here the considered parameterization for operator $\mathcal{F}$ as well as noise process parameters, that maximize the likelihood of observed data $\vect{y}$. Let us denote by $\theta$ the set of all model parameters. Especially, we assume that operator $\mathcal{F}$ lies within a finite-dimensional space of operators $\mathcal{X}$. We describe in Section \ref{sec:EnKS-EM} and Section \ref{sec:VODEN} the considered parameterization using neural network architectures. Formally, the EM procedure comes to maximize likelihood $p_{\theta}(\vect{y})$:
\begin{align}
    \ln p_{\theta}(\vect{y}) = \ln \int p_{\theta}(\vect{y}, \vect{x})d\vect{x}
    \label{eq:ln_ll_EM}
\end{align}
For any arbitrary distribution $q$,  this function can be decomposed into:
\begin{align}
    \ln p_{\theta}(\vect{y}) = \mathcal{L}(\vect{y},p_{\theta},q) +\mathrm{KL}(q||p_{\theta})
    \label{eq:decomposition_EM}
\end{align}
where:
\begin{align}
    \mathcal{L}(\vect{y},p_{\theta},q) &= \int q(\vect{x}|\vect{y}) \ln{\frac{p_{\theta}(\vect{y}, \vect{x})}{q(\vect{x}|\vect{y})}} d \vect{x}
    \label{eq:elbo_em}
    \\
    \mathrm{KL}(q||p_{\theta}) &= -  \int q(\vect{x}|\vect{y}) \ln{\frac{p_{\theta}( \vect{x}|\vect{y})}{q(\vect{x}|\vect{y})}} d \vect{x}
    \label{eq:kl_em}
\end{align}
The EM algorithm alternately maximizes the Evidence Lower Bound (ELBO) $\mathcal{L}(\vect{y},p_{\theta},q)$ with respect to $q$ (the E-step) and $\theta$ (the M-step) in each iteration. The E-step comes to determine the posterior $q=p_\theta(\vect{x}|\vect{y})$ knowing $\theta$ and the M-step amounts to maximizing the log-likelihood of the observation conditionally to posterior $q$. Compared to the direct joint minimization of ELBO w.r.t. $q$ and $\theta$ \cite{graves_practical_2011, mnih_neural_2014, maddison_filtering_2017}, EM procedures usually lead to stable solutions \cite{ghahramani_learning_1999} and faster convergence \cite{hinton_wake-sleep_1995}. As stated in \cite{ghahramani_learning_1999}, they are also particularly appealing to deal with noise and missing data/irregularly sampling.

For the simplest case, analytic expressions can be derived both for the Expectation and Maximization steps, as for instance for Gaussian mixture models \cite{bishop_pattern_2006}. When dealing with non-linear dynamical systems as addressed here, neither the E-step nor the M-step can be solved analytically. Then, approximations or numerical solutions have to be considered. Restricting the problem to a family of distributions that can be factored over $t$, we can rewrite Eq. \ref{eq:elbo_em} as:
\begin{align}
    \mathcal{L}(\vectall{y},p_{\theta},q) &= \int q(\vectall{x}|\vectall{y}) \ln{\frac{p_{\theta}(\vectall{y}|\vectall{x})p_{\theta}(\vectall{x})}{q(\vectall{x}|\vectall{y})}} d\vect{x}  
    \\
    &=  \mathbb{E}_{\vectall{x} \sim q(\vectall{x}|\vectall{y})}\left[ \ln p_{\theta}(\vectall{y}|\vectall{x})\right] - \mathrm{KL}\big[q(\vectall{x}|\vectall{y})||p_{\theta}(\vectall{x})\big]
    \label{eq:elbo2}
\end{align} 
where:
\begin{equation}
    p_{\theta}(\vectall{x}) = p_{\theta}(\vect{x}_1)\prod_{t=1}^{T-1} p_{\theta}(\vect{x}_{t+1}|\vect{x}_t)
    \label{eq:px}
\end{equation}
\begin{align}
    q(\vectall{x}|\vectall{y}) &= \prod_{t=1}^T q(\vect{x}_t|\vectall{y})
    \\
    p_{\theta}(\vectall{y}|\vectall{x}) &= \prod_{t=1}^T p_{\theta}(\vect{y}_t|\vect{x}_t)
\end{align}

Regarding specifically the E-step, {\em i.e.} the approximation of true posterior $p_\theta\big(\vect{x}_t|\vect{y}_{1:T}\big)$ given model parameters $\theta$, we consider two alternative solutions. On the one hand, stochastic filtering approaches naturally apply, especially ensemble Kalman and particle filtering schemes \cite{evensen_ensemble_2000, doucet_tutorial_2009}. We here focus on Ensemble Kalman smoothers (EnKS), which are among the state-of-the-art approaches in data assimilation \cite{evensen_data_2009}. On the other hand, variational Bayesian approximation \cite{bishop_pattern_2006} exploits an explicit parameterization of posterior $q$. This variational setting has been recently investigated using neural-network-based parameterization and proved computational efficient for inference of latent state and dynamics \cite{graves_practical_2011, mnih_neural_2014}. We further detail these two alternative approaches, referred to respectively as EnKS-EM and VODEN (Variational ODE Networks), in Section \ref{sec:EnKS-EM} and Section \ref{sec:VODEN}.

Regarding the M-step, it relates to the learning of a parametric representation of dynamical operator $\mathcal{F}$ as considered in previous works \cite{brunton_discovering_2016,fablet_bilinear_2017} using inferred hidden states series rather than observation data, and may implement gradient-based schemes using neural network frameworks. Many ``pure" machine learning methods \cite{graves_practical_2011, mnih_neural_2014, maddison_filtering_2017, pu_variational_2016, chung_recurrent_2015, fraccaro_sequential_2016} use here a Monte Carlo approximation of the gradient of $\mathcal{L}$ for training. However, we empirically observe that when the dynamical system is chaotic, this approach turns out to be unstable. Recall that under Gaussian hypotheses, $p(\vect{x}_{t+1}|\vect{x}_{t})$ and $p(\vect{y}_{t}|\vect{x}_{t})$ have mean $f(\vect{x}_{t})$, $\mathcal{H}(\vect{x}_{t})$ and variance $\Sigma^{\vect{x}}_{t}$, $\Sigma^{\vect{y}}_{t}$ respectively. We can decompose $\theta$ into $\big[ \theta_f, \theta_{\Sigma^{\vect{x}}_{t}}, \theta_{\mathcal{H}}, \theta_{\Sigma^{\vect{y}}_{t}}\big]^T$. Maximizing $\mathcal{L}(\vectall{y},p_{\theta},q)$ means simultaneously maximizing $\mathbb{E}_{\vectall{x} \sim q(\vectall{x}|\vectall{y})}\left[ \ln p_{\theta}(\vectall{y}|\vectall{x})\right]$ over $\big[\theta_{\mathcal{H}}, \theta_{\Sigma^{\vect{y}}_{t}}\big]^T$ and minimizing $\mathrm{KL}\big[q(\vectall{x}|\vectall{y})||p_{\theta}(\vectall{x})\big]$ over $\big[ \theta_f, \theta_{\Sigma^{\vect{x}}_{t}} \big]^T$. Since we are more interested in $\mathcal{F}$, we can ignore the first term of Eq. \ref{eq:elbo2}. Because $q$ is fixed and $p_{\theta}(\vect{x}_{t+1}|\vect{x}_t)$ is Gaussian, suppose $p_{\theta}(\vect{x}_1)$ known, we now have to maximize:

\begin{equation}
    \mathbb{E}_q\left[ \ln \left( \sum_{t=1}^{T-1} \left((2\pi)^k|\Sigma^{\vect{x}}_{t}|\right)^{-1/2} \exp \left( -\frac{1}{2} (f(\vect{x}_t) - \vect{x}_{t+1})^T (\Sigma^{\vect{x}}_{t})^{-1} (f(\vect{x}_t) - \vect{x}_{t+1})\right) \right)\right]
    \label{eq:m_gaussian}
\end{equation}
with $|\Sigma^{\vect{x}}_{t}|$ is the determinant of $\Sigma^{\vect{x}}_{t}$.

Using another simplification $\Sigma^{\vect{x}}_{t} = \sigma^2 \vect{I}$, the M-step now amounts to minimizing the following loss over $\theta_f$:
\begin{equation}
    \mathbb{E}_q\left[\sum_{t=1}^{T-1}||f(\vect{x}_t) - \vect{x}_{t+1}||_2\right]
    \label{eq:before_MAP}
\end{equation}
Where the posterior $q$ is the current approximation of the true posterior $p_\theta\big(\vect{x}_t|\vect{y}_{1:T}\big)$, $||.||_2$ denotes the Euclidean norm. $f(\vect{x}_t)$, parameterized by $\theta_f$, is the forecast of the hidden state $\vect{x}_t$. Again we still have the expectation with respect to $q$ (computed in the E-step). One simple solution is to use Maximum A Posteriori (MAP) instead of Maximum Likelihood (ML) in the E-step. In other words, we restrict the family of distribution $q$ to Dirac distributions. \ref{eq:before_MAP} then becomes:
\begin{equation}
    loss_M = \sum_{t=1}^{T-1}||f(\vect{x}^*_t) - \vect{x}^*_{t+1}||_2
\end{equation}
With 
\begin{align}
    \vect{x}^*_t = \mathbb{E}_q\left[q(\vect{x}_t|\vectall{y})\right]
\end{align}
We refind here the short-term prediction error as the objective that widely used in machine learning \cite{pathak_using_2017, raissi_multistep_2018, raissi_forward-backward_2018, qin_data_2018}.

Although the hypothesis $\Sigma^{\vect{x}}_{t} = \sigma^2 \vect{I}$ is not true for most of systems, and the lower-bound computed using MAP is looser than the bound computed using ML,  we empirically observe that these approximations stabilize and significantly accelerate the training process.

\subsection{Ensemble Kalman Smoother--Expectation Maximization (EnKS-EM)}
\label{sec:EnKS-EM}

For non-linear dynamical systems, the Ensemble Kalman Smoother (EnKS) is one of the most popular and efficient data assimilation schemes. The EnKS was first introduced by Evensen \cite{evensen_ensemble_2000} and has rapidly become popular thanks to its simplicity, both in terms of conceptual formulation and implementation. It is an improvement of the Ensemble Kalman Filter \cite{evensen_sequential_1994}. Similar to EnKF, the backbone of the EnKS is a so-called Markov Chain Monte Carlo (MCMC) method to solve Fokker–Planck equation (also know as the the Kolmogorov forward equation) \cite{kolmogoroff_analytical_1931} which represents the evolution of $p\big(\vect{x}_t\big)$. Briefly speaking, the EnKS uses an ensemble of model states (``cloud points") to represent $p\big(\vect{x}_t\big)$. Any moment of this distribution can be calculated easily by integrating the ensemble of states forward in time. For a series of observed data points $\vect{y}_{1:T} = \{\vect{y}_{1},..,\vect{y}_{T}\}$, the posterior $p\big(\vect{x}_t|\vect{y}_{1:T}\big)$ can be obtained by applying the Kalman update formulas, the covariance of the hidden states being estimated by the sample covariance computed from the ensemble members. For the details of the formulation, the implementation as well as the application of the EnKS, we let the reader refer to \cite{evensen_ensemble_2000} and \cite{evensen_ensemble_2003}.

The Ensemble Kalman Smoother--Expectation Maximization (EnKS-EM) exploits the EnKS approximation $q_{EnKS}\big(\vect{x}_t|\vect{y}_{1:T}\big)$ for the true posterior $p_{\theta}\big(\vect{x}_t|\vect{y}_{1:T}\big)$ within the proposed EM framework. Overall, the considered algorithm is presented in Alg. \ref{alg:EnKS-EM}.   

\begin{algorithm}[t]
\caption{EnKS-EM procedure for learning dynamical systems}
\label{alg:EnKS-EM}
    \begin{algorithmic}[1]
    \State {\bfseries\scshape EnKS-EM}$(\vect{y}_{1:T}, \theta, N_M)$:
    \State Initialize $\theta$
    
    \While {not converge}
        \State \% E-step
        \For {$t \in \{1,..,T\}$}
            \State Estimate $p_\theta (\vect{x}_t|\vect{y}_{1:T})$
            \State Compute $\vect{x}^*_t = \mathbb{E}\left[q_{EnKS} \big(\vect{x}_t|\vect{y}_{1:T}\big)\right]$    
        \EndFor
        
        \State \% M step
        \State gradient\_step = 0
        \If {gradient\_step < $N_M$}
            \State Optimize $\theta$ to minimize $loss_M = \sum_{t=1}^{T-1}||f_\theta(\vect{x}^*_t)-\vect{x}^*_{t+1}||_2$ 
            \State gradient\_step = gradient\_step + 1
        \EndIf
    \EndWhile
    
    \end{algorithmic}
\end{algorithm}

\subsection{Variational Ordinary Differential
Equations Networks (VODEN)}
\label{sec:VODEN}    

\begin{figure}
  \centering
  \includegraphics[width=60mm]{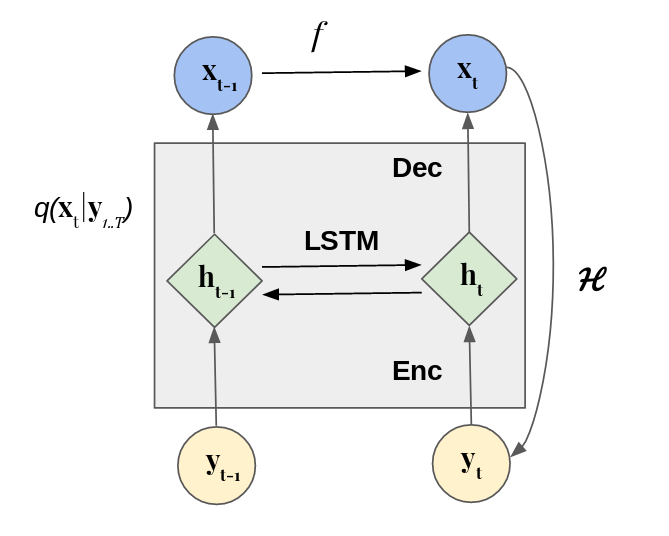}%
  \caption {VODEN architecture.}
 \label{fig:VBRNN}
\end{figure}

The second EM scheme for learning non-linear dynamical models is the Variational Ordinary Differential Equations Networks (VODEN), where the approximate posterior $q = q_\phi(\vect{x}_t|\vect{y}_{1:T})$ is parametrized by neural networks. This approach is inspired by the applausive successes of the neural network-based implementation of the variational methods \cite{maddison_filtering_2017, chung_recurrent_2015, fraccaro_sequential_2016}. The inference network, whose backbone is a LSTM, takes $\vect{y}_{1:T}$ as input and provides $\vect{x}_{1:T}$ at the output. Usually, an encoder and an decoder are also added to improve the capacity of $q$, as shown in Fig. \ref{fig:VBRNN}. 

The parameters of the whole inference network are denoted as $\phi$. Using the same simplifications in the M-step, in the E-step, we calibrate $q_\phi$ to minimize the following loss
\begin{align}
    loss_E = \sum_{t=1}^{T-1}\big(\lambda\Phi_t(||\mathcal{H}(f(\vect{x}^*_t)) - \vect{y}_t||_2) + ||f(\vect{x}^*_t) - \vect{x}^*_{t+1}||_2\big)
    \label{eq:lossE}
\end{align}
with $\vect{x}_t$ is the output of the inference network.

The first term of $loss_E$ is analogous to the innovation (the measurement of pre-fit residual) in data assimilation schemes, while the second term ensures that if $f$ is the true state-transition function, the minimization of $loos_E$ amounts to retrieving $q_\phi(\vect{x}_t|\vect{y}_{1:T})$ that best approximates the posterior $p_\theta(\vect{x}_t|\vect{y}_{1:T})$. $\lambda$ is a multiplication factor. The details of the training procedure is presented in Alg. \ref{alg:VODEN}. It may be noted that we set a maximum number of gradient steps for both the E and M step and do not let the gradient descent converge within each step. This was proven numerically more efficient in our experiments.

\begin{algorithm}[t]
\caption{VODEN procedure for learning dynamical systems}
\label{alg:VODEN}
    \begin{algorithmic}[1]
    \State {\bfseries\scshape VODEN}$(\vect{y}_{1:T}, \theta, \phi, N_E, N_M)$:
    \State Initialize $\theta$
    
    \While {not converge}
        \State \% E-step
        \State gradient\_step = 0
        \If {gradient\_step < $N_E$}
            \State Compute $\vect{x}^*_{1:T} = \mathbb{E}\left[q_{\phi}(\vect{x}_{1:T}|\vect{y}_{1:T})\right]$
            \State $loss_E = \sum_{t=1}^{T-1}\big(\lambda\Phi_t(||\mathcal{H}(f(\vect{x}^*_t)) - \vect{y}_t||_2) + ||f(\vect{x}^*_t) - \vect{x}^*_{t+1}||_2\big)$
            \State Optimize $\phi$ to minimize $loss_E$
            \State gradient\_step = gradient\_step + 1
        \EndIf
        
        \State \% M step
        \State gradient\_step = 0
        \If {gradient\_step < $N_M$}
            \State Optimize $\theta$ to minimize $loss_M = \sum_{t=1}^{T-1}||f(\vect{x}^*_t)-\vect{x}^*_{t+1}||_2$ 
            \State gradient\_step = gradient\_step + 1
        \EndIf
    \EndWhile
    
    \end{algorithmic}
\end{algorithm}

\section{Experiments and results}
\label{sec:experiments}

We tested the proposed methodology on two classic systems for learning non-linear dynamics: the Lorenz--63 system \cite{lorenz_deterministic_1963}.
We examined the learning under significant noise level with partial and irregular sampling of the observations. The performance of the proposed methodology is compared with state-of-the-art methods, namely Analog forecasting (AF) \cite{lguensat_analog_2017}, a Sparse regression model (SR) \cite{brunton_discovering_2016} and a bilinear residual Neural Network architecture (BiNN) \cite{fablet_learning_2017}.

\subsection{Lorenz--63 chaotic system}

\begin{wrapfigure}[12]{r}{0.45\textwidth}
    \centering
    \includegraphics[width=0.4\textwidth,clip, trim=100mm 40mm 80mm 40mm]{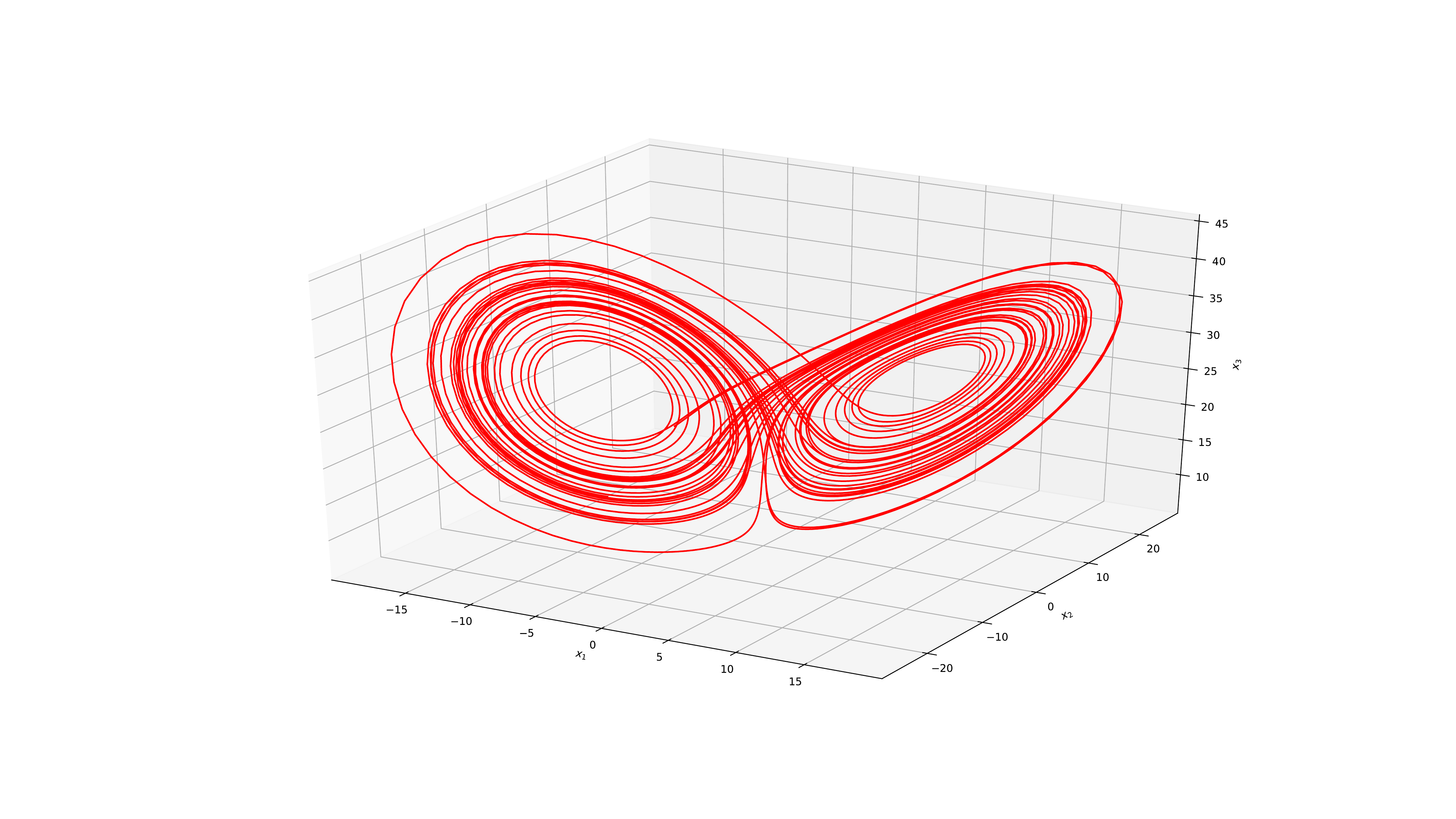} 
    \caption{Lorenz attractor of the Lorenz--63 system When $\sigma =10$, $\rho=28$ and  $\beta=8/3$.}
    \label{fig:true_lorenz}
\end{wrapfigure}

The Lorenz--63 dynamical system is a 3-dimensional model governed by the following ODE: 
\begin{equation}
    \label{eq:lorenz-63}
    \left \{\begin{array}{ccl}
    \frac{d\mathrm{x}_{t,1}}{dt} &=&\sigma \left (\mathrm{x}_{t,2}-\mathrm{x}_{t,2} \right ) \\
    \frac{d\mathrm{x}_{t,2}}{dt}&=&\rho \mathrm{x}_{t,1}-\mathrm{x}_{t,2}-\mathrm{x}_{t,1}\mathrm{x}_{t,3} \\
    \frac{d\mathrm{x}_{t,3}}{dt} &=&\mathrm{x}_{t,1}\mathrm{x}_{t,2}-\beta \mathrm{x}_{t,3}
    \end{array}\right.
\end{equation}
When $\sigma =11$, $\rho=28$ and  $\beta=8/3$, this system has a chaotic behavior, with the Lorenz attractor shown in Fig. \ref{fig:true_lorenz}.

We simulate Lorenz-63 state sequences using the LOSDA (Livermore Solver for Ordinary Differential Equations) ODE solver \cite{hindmarsh_odepack_1983} with an integration step of 0.01. We then add Gaussian noise with several variance levels $\sigma^2$ and evaluate the learned models given the noisy training data. This means, $\mathcal{H}$ is now an identity operator, $\epsilon_t$ is zero-mean Gaussian white noise.

For this task, we used the following settings: for AF, SR and BiNN, we used the setting mentioned in the original paper of each method;
for EnKS-EM and VODEN: the integration scheme is the neural network implementation of the Runge-Kutta 4 integration scheme as in \cite{fablet_bilinear_2017}; $\mathcal{F}$ is parametrized by a bi-linear residual network with the same setting as in \cite{fablet_bilinear_2017}.
We may note that this parametrization embeds the true parametrization of the Lorenz-63 model. For EnKS-EM, we chose an ensemble of 50 members. For VODEN, the approximate posterior $q_\phi(\vect{x}_t|\vect{y}_{1:T})$ was modeled by a 2-layer bi-directional LSTM with a size of 9. Both the encoder and the decoder were modelled by a fully connected network, with one hidden layer of size 7. We used RMSprop as the optimizer, with a learning rate of $3e-4$. $\lambda$ was set to $0.1$.  

\subsection{Identification with noisy observations}
\label{sec:lorenz63_noisy}

We first assess the identification performance with noisy observations only, which means that masking operator $\Phi_t$ is the identity at all time steps. We evaluated both short-term prediction performance and the capacity of maintaining the very-long-term topology through the first Lyapunov exponent $\lambda_1$ \cite{sprott_chaos_2003} calculated in a forecasting sequence of length 10 lorenz-time (10000 time steps). The first Lyapunov exponent of the considered Lorenz-63 system is 0.91 \cite{sprott_chaos_2003}. 

As shown in Table \ref{tab:lorenz63_noisy} both the EnKS-EM and the VODEN outperform existing methods. The EnKS-EM gives the best forecasting when the noise level is small, however, when the noise level increases, the forecasting gradually becomes worse. This is because the increase of noise level leads to the increase of uncertainty (error covariance), we may need a bigger ensemble to maintain the same performance level. On the other hand, the VODEN performs extremely well on very noisy data. We believe that this relates to the ability of LSTM-based models to capture longer-term time patterns in the data. 

\begin{figure}
    \centering
    \includegraphics[width=0.8\textwidth,clip, trim=40mm 40mm 20mm 40mm]{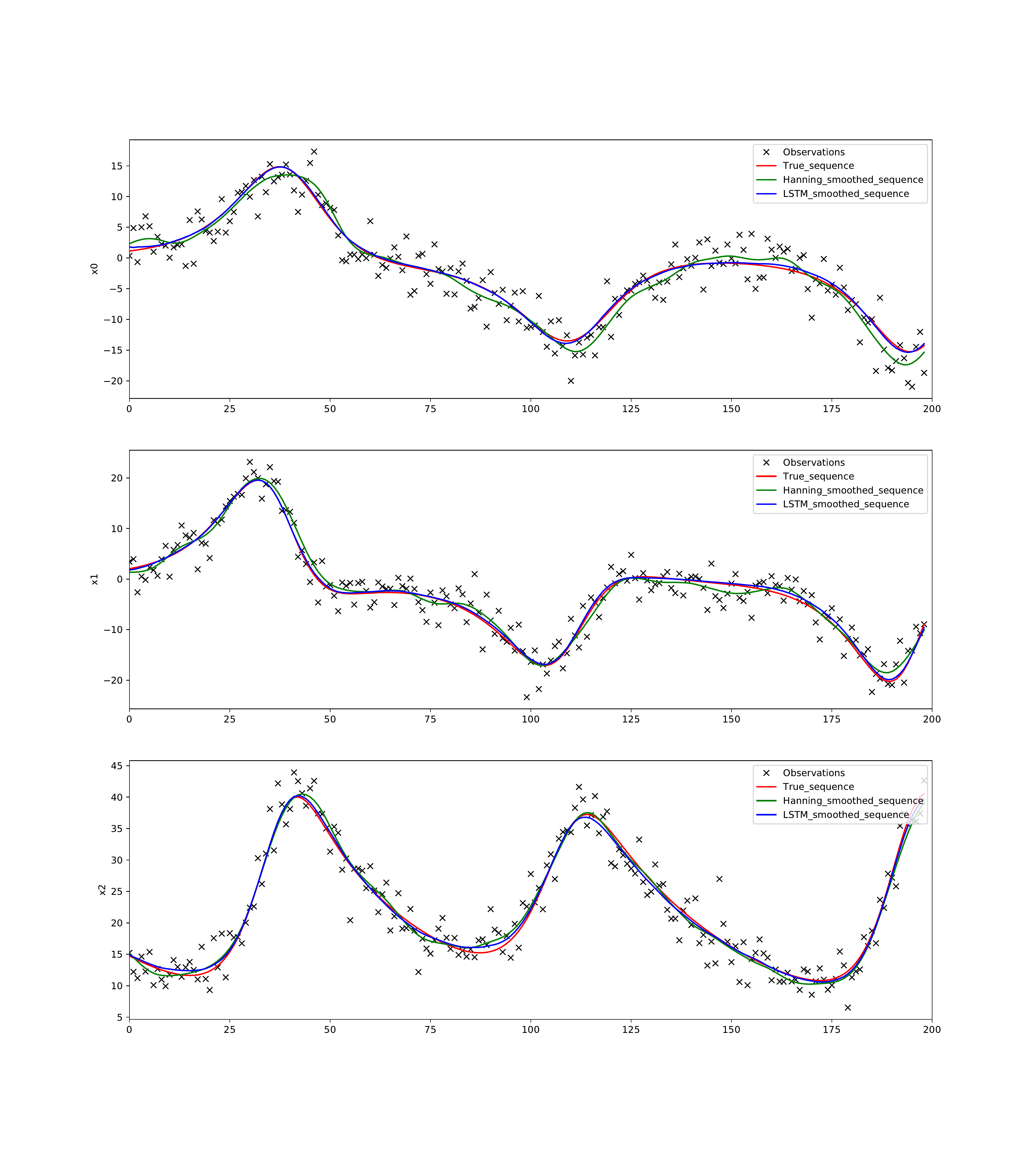}
    \caption{LSTM acts as a denoiser.}
    \label{fig:denoiser}
\end{figure}

It should be noted that the BiNN is the EnKS-EM/VODEN without the inference schemes. When the training data are clean, many data-driven methods can successfully identify the underlying dynamics. For example, the BiNN and the VODEN have similar result when $\sigma^2=0.5$. However, when the data is very noisy, all the three model without inference schemes (AF, SR, BiNN) fail. We also did an additional experiment by adding a preprocessing step, in which a Hanning window of size 20 was applied to reduce the effect of noise, before applying SR. This setting is referred as SR\_Hann in Table \ref{tab:lorenz63_noisy}. The Hanning denoiser does significantly improve the performance of SR. Fig. \ref{fig:denoiser} shows the sequences smoothed by the Hanning window and the LSTM. However, overall, our proposed methods are more robust to noise. For example, at a noise level of 64, the EnKS-EM gives a better prediction error than the SR\_Hann with a factor of 2, the VODEN performs even better, with a factor of 5. Topology-wise, we can see that the attractors generated by the EnKS-EM and the VODEN are visually better.          

\begin{table}[tb]
    \caption {Short-term forecasting error and very-long-term forecasting topology of data-driven models learned on noisy Lorenz-63 data.}
    \label{tab:lorenz63_noisy}

    \centering
    \begin{tabular}{ll*{7}c}
    \toprule
    \multicolumn{2}{c}{\multirow{2}{*}{Model}} & \multicolumn{7}{c}{$\sigma^2$} \\
    & & 0.5 & 2 & 4 & 8 & 16  & 32 & 64\\
    \midrule \midrule 
    \multirow{3}{*}{AF}
    & $t_0+h$       & 0.239  & 0.596  & 0.629   &  0.969 & 2.845  & 3.309  & 3.735\\
    & $t_0+4h$      & 0.245  & 0.698  & 0.795   &  2.213 & 3.540  & 3.887  & 7.944\\ 
    &  $\lambda_1$  & -1.356 & -2.496 & -16.339 & -1.900 & 12.064 & 32.432 & 71.302\\
    \midrule
    \multirow{3}{*}{SR}
    & $t_0+h$       & 0.012 & 0.034 & 0.057 & 0.106  & 0.187 & 0.305 & 0.440\\
    & $t_0+4h$      & 0.037 & 0.104 & 0.177 & 0.326  & 0.577 & 0.933 & 1.330\\ 
    &  $\lambda_1$  & 0.890 & 0.876 & 0.833 & -0.367 & nan & nan & -0.043\\
    \midrule
    \multirow{3}{*}{SR\_Hann}
    & $t_0+h$       & 0.030 & 0.031 & 0.033 & 0.041 & 0.056 & 0.077 & 0.108\\
    & $t_0+4h$      & 0.085 & 0.088 & 0.095 & 0.123 & 0.173 & 0.236 & 0.327\\ 
    &  $\lambda_1$  & 0.902 & 0.858 & 0.845 & 0.772 & 0.824 & 0.802 & 0.777\\
    \midrule
    \multirow{3}{*}{BiNN}
    & $t_0+h$       & 0.013 & 0.022 & 0.060 & 0.096 & 0.150 & 0.252 & 0.321\\
    & $t_0+4h$      & 0.043 & 0.061 & 0.177 & 0.296 & 0.466 & 0.773 & 0.972\\ 
    &  $\lambda_1$  & 0.912 & 0.833 & 0.844 & nan & nan & -0.014 & nan\\
    \midrule
    \multirow{3}{*}{EnKS-EM}
    & $t_0+h$       & 0.004 & 0.008 & 0.012 & 0.017 & 0.020 & 0.050 & 0.060\\
    & $t_0+4h$      & 0.013 & 0.027 & 0.040 & 0.055 & 0.060 & 0.156 & 0.197\\ 
    &  $\lambda_1$  & 0.859 & 0.842 & 0.888 & 0.878 & 0.901 & 0.892 & 0.803\\
    \midrule
    \multirow{3}{*}{VODEN}
    & $t_0+h$           & 0.013 & 0.023 & 0.022 & 0.026 & 0.021 & 0.028 & 0.024\\
    & $t_0+4h$          & 0.038 & 0.062 & 0.059 & 0.067 & 0.061 & 0.081 & 0.070\\ 
    &  $\lambda_1$      & 0.896 & 0.909 & 0.859 & 0.919 & 0.898 & 0.904 & 0.934\\
    \bottomrule
    \end{tabular}

\end{table}

\begin{figure}
    \centering
    \includegraphics[width=1.0\textwidth,clip, trim=40mm 20mm 30mm 20mm]{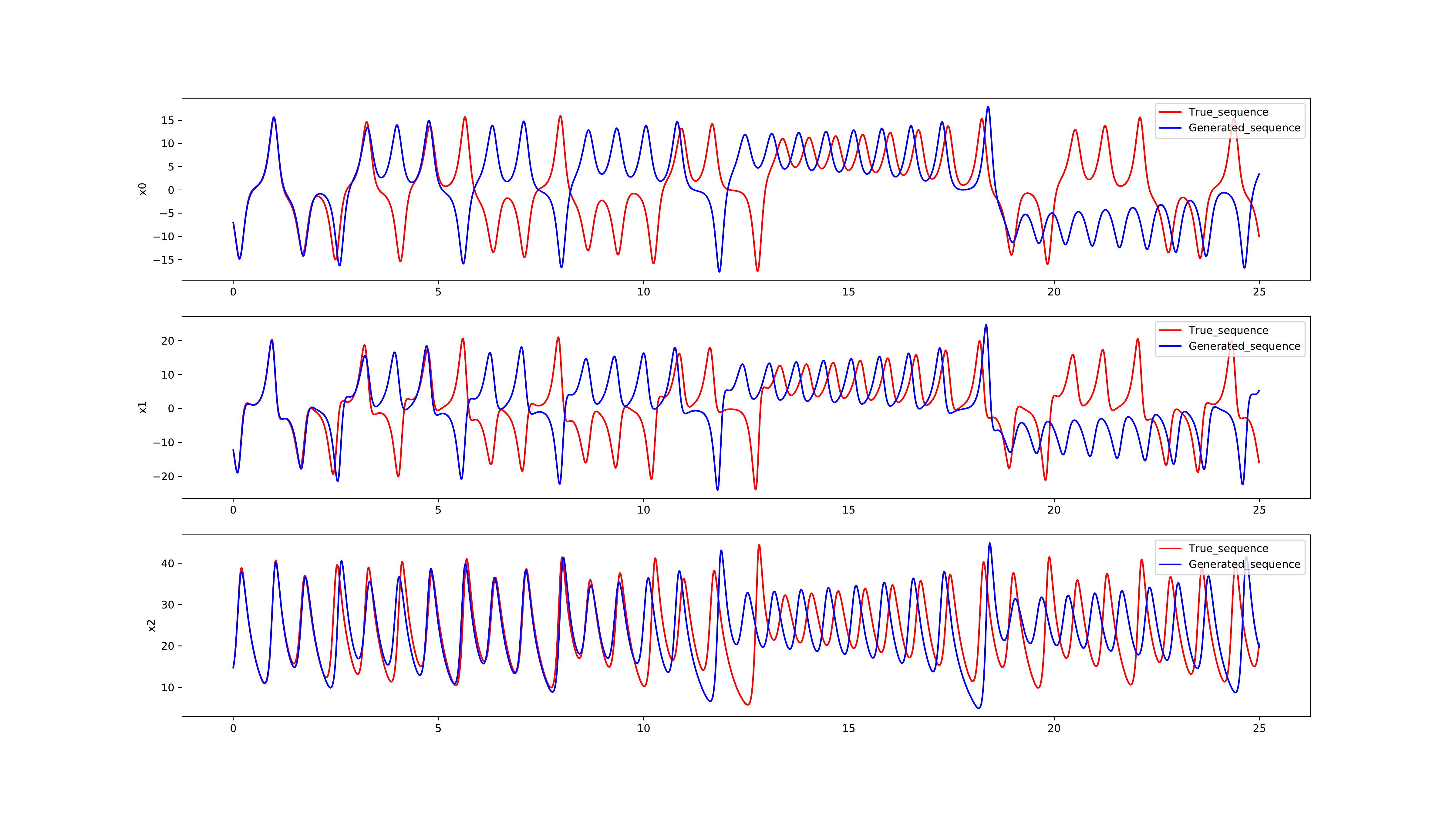}
    \caption{An example of sequences forecast by a VODEN learned on Lorenz-63 data, $\sigma^2=4$.}
    \label{fig:seq_comparision_voden_lorenz63_noisy_4}
\end{figure}

We plot in Fig. \ref{fig:seq_comparision_voden_lorenz63_noisy_4} the sequences forecast by the a VODEN learned on Lorenz-63 data, $\sigma^2=4$. The Lorenz-63 system has a positive Lyapunov exponent, any small difference between the true and learned models grows exponentially. However, the "patterns" of the sequences, or the topology, remains intact. As discussed in \cite{brunton_discovering_2016} \cite{pathak_using_2017}, for dynamical models identification, the most important criterion is the ability to maintain this topology in very-long forecasting. Fig. \ref{fig:attactors_lorenz63_noisy} show the attractor of the sequences generated by the models in Table \ref{tab:lorenz63_noisy}.

\newcommand{\ltrim}{100mm}%
\newcommand{\btrim}{40mm}%
\newcommand{\rtrim}{100mm}%
\newcommand{\ttrim}{40mm}%

\newcommand{\nfwidth}{0.8\linewidth}%
\newcommand{\swidth}{0.057\linewidth}%
\newcommand{\bwidth}{0.16\linewidth}%
\begin{figure}
    \centering
	\begin{subfigure}[t]{0.04\linewidth}
		\hfill
	    \vspace{-2.5mm}
		\caption*{}
	\end{subfigure}%
	\begin{subfigure}[t]{0.96\linewidth}
		\hspace{4mm} $\sigma^2=0.5$ \hspace{\swidth} $\sigma^2=2.0$ \hspace{\swidth} $\sigma^2=4.0$ \hspace{\swidth} $\sigma^2=8.0$ \hspace{\swidth} $\sigma^2=16.0$ \hspace{\swidth} $\sigma^2=32.0$ \hfill
	\end{subfigure}%
	
	\begin{subfigure}[b]{0.04\linewidth}
	    \rotatebox[origin=t]{90}{\scriptsize AF}\vspace{0.79\linewidth}
	\end{subfigure}%
	\begin{subfigure}[t]{0.96\linewidth}
		\centering
		\includegraphics[width=\bwidth,clip, trim=100mm 40mm 80mm 40mm]{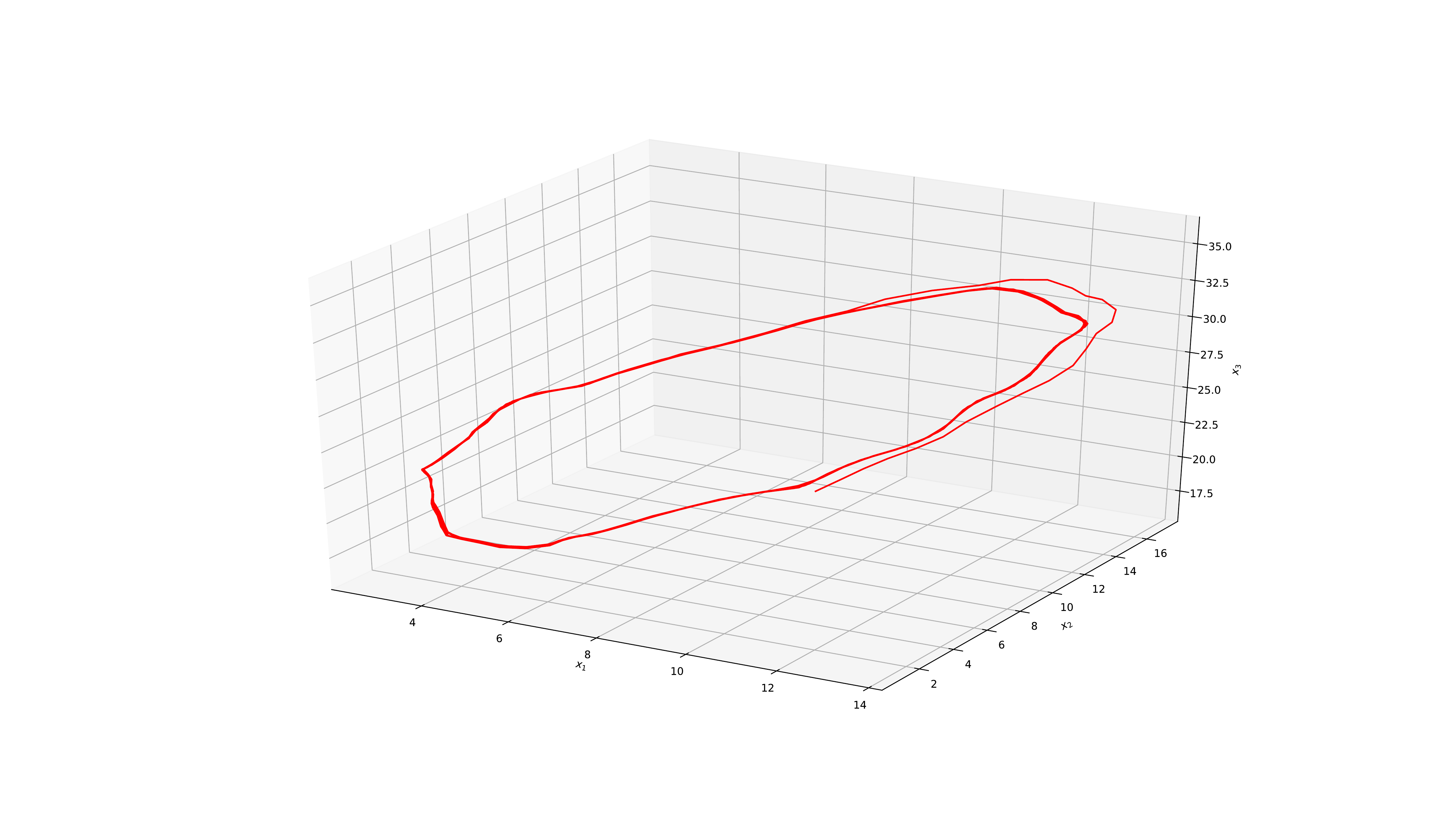}
		\includegraphics[width=\bwidth,clip, trim=100mm 40mm 80mm 40mm]{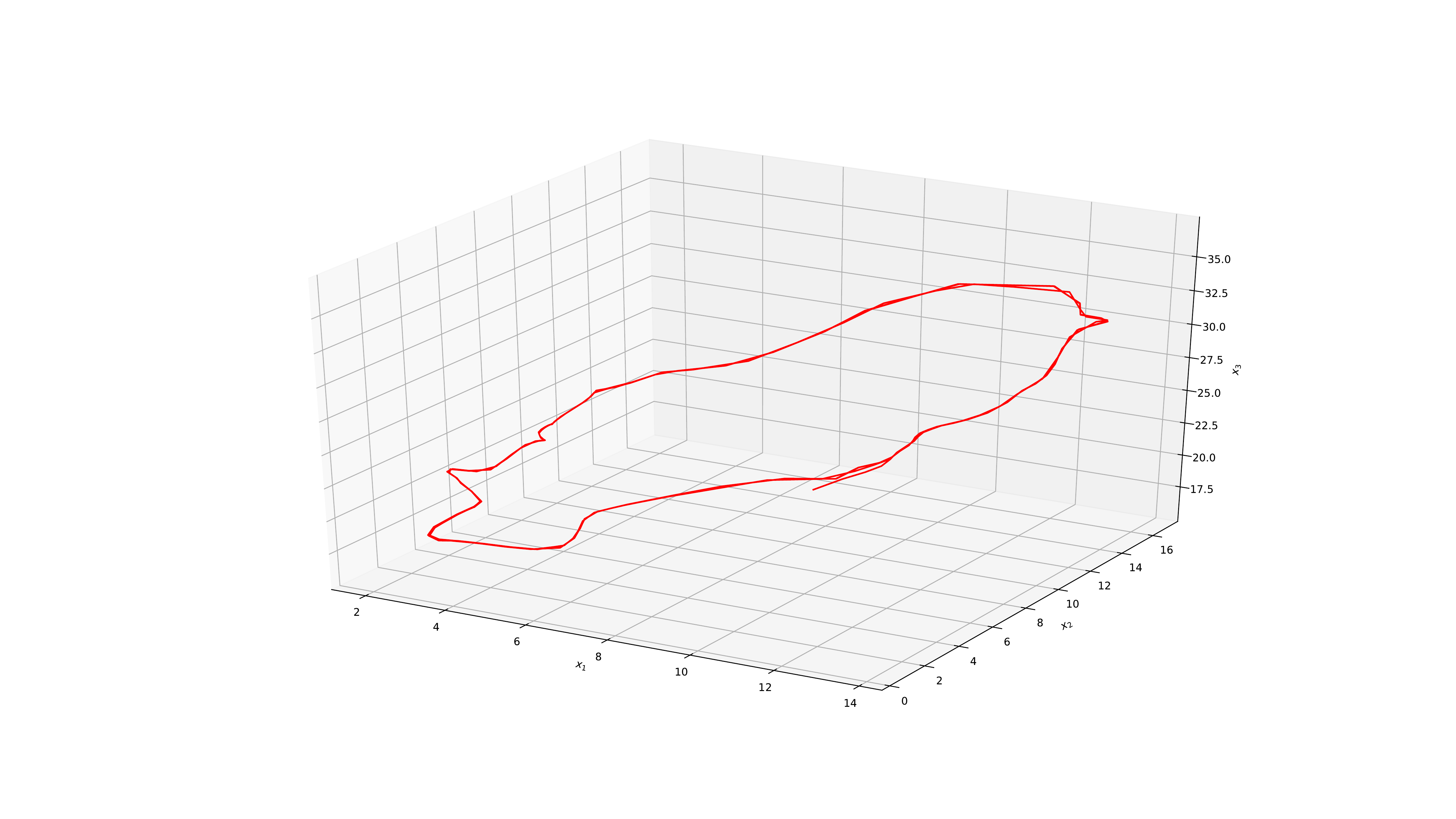}
		\includegraphics[width=\bwidth,clip, trim=100mm 40mm 80mm 40mm]{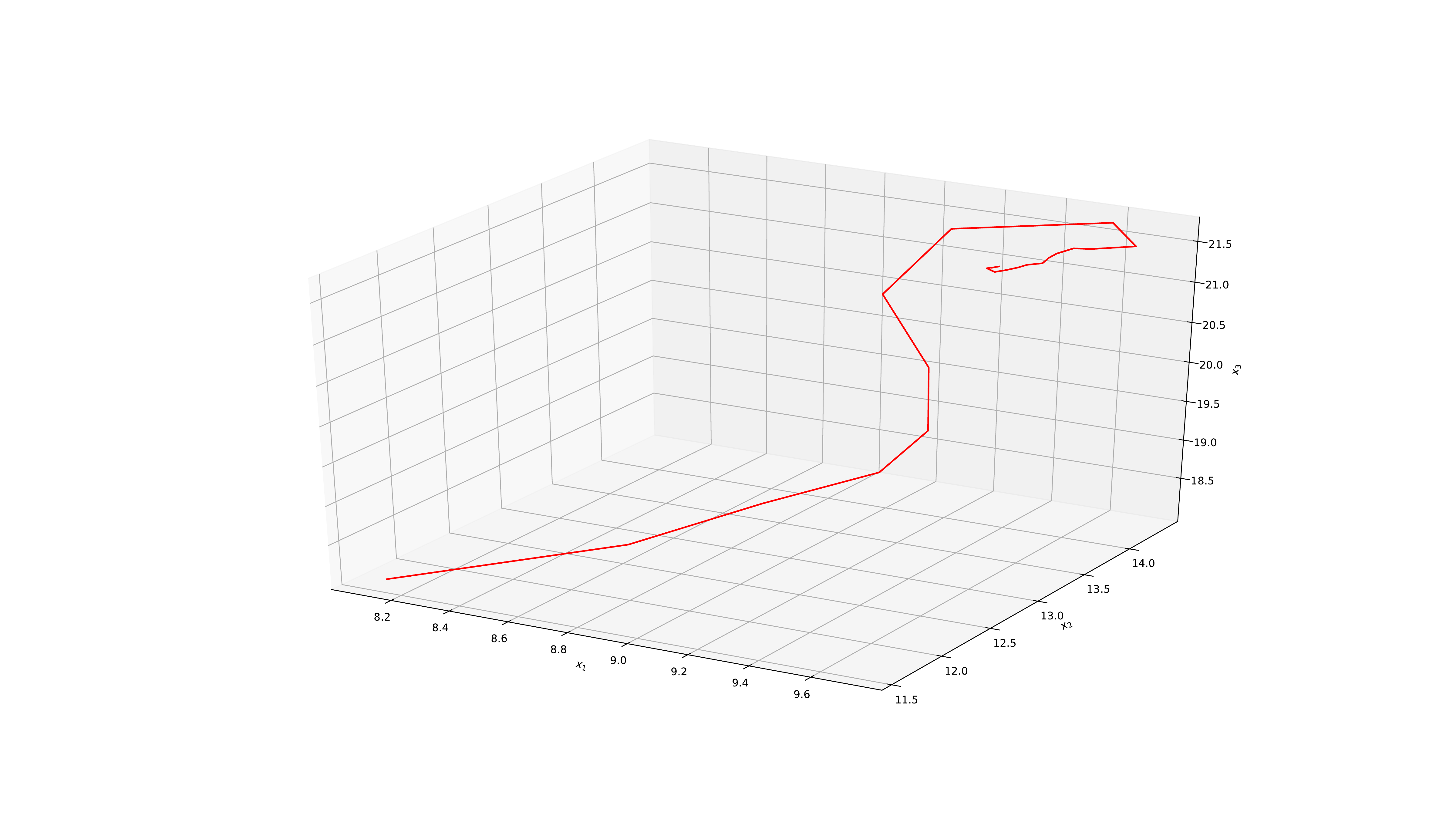}
		\includegraphics[width=\bwidth,clip, trim=100mm 40mm 80mm 40mm]{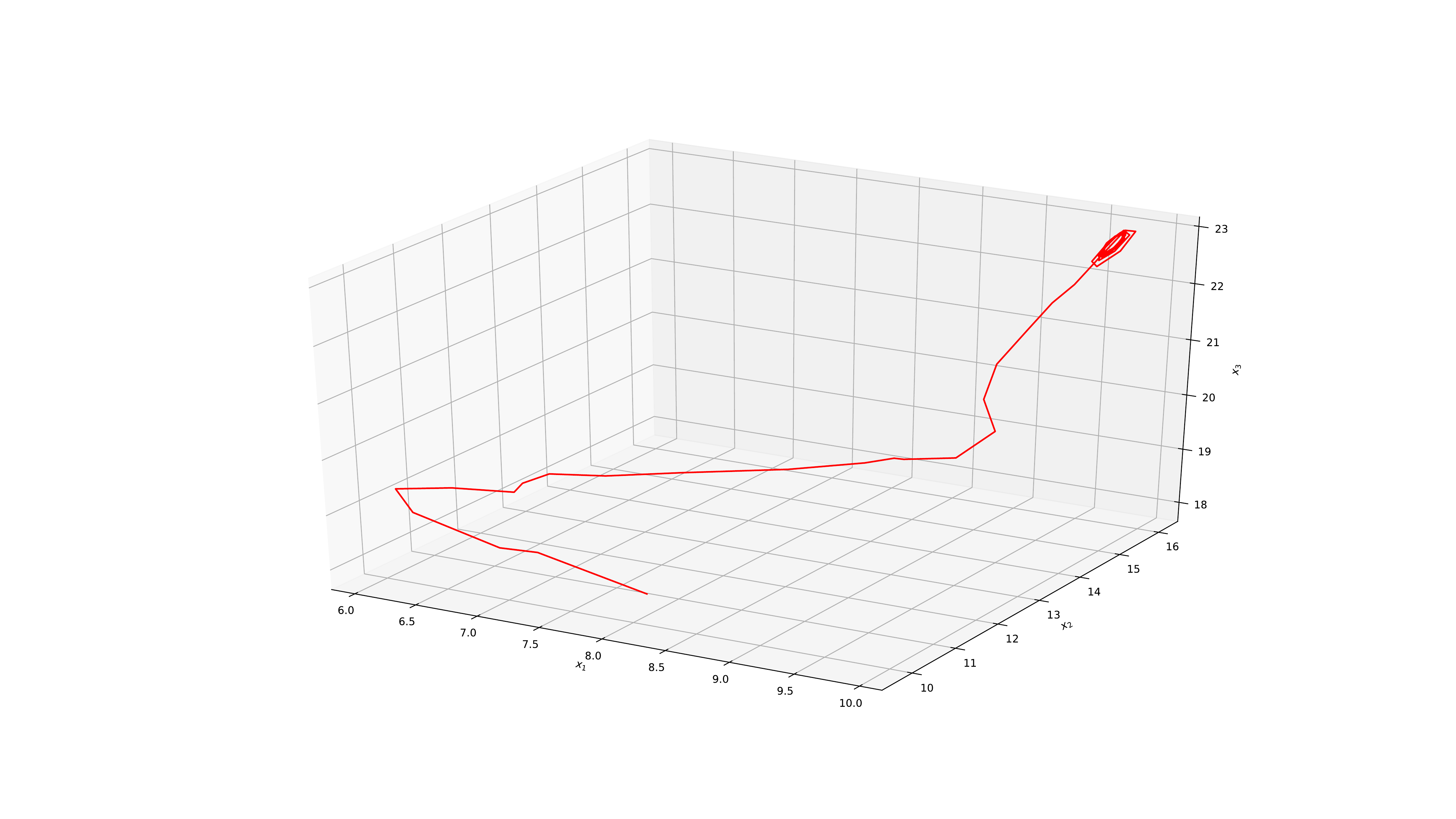}
		\includegraphics[width=\bwidth,clip, trim=100mm 40mm 80mm 40mm]{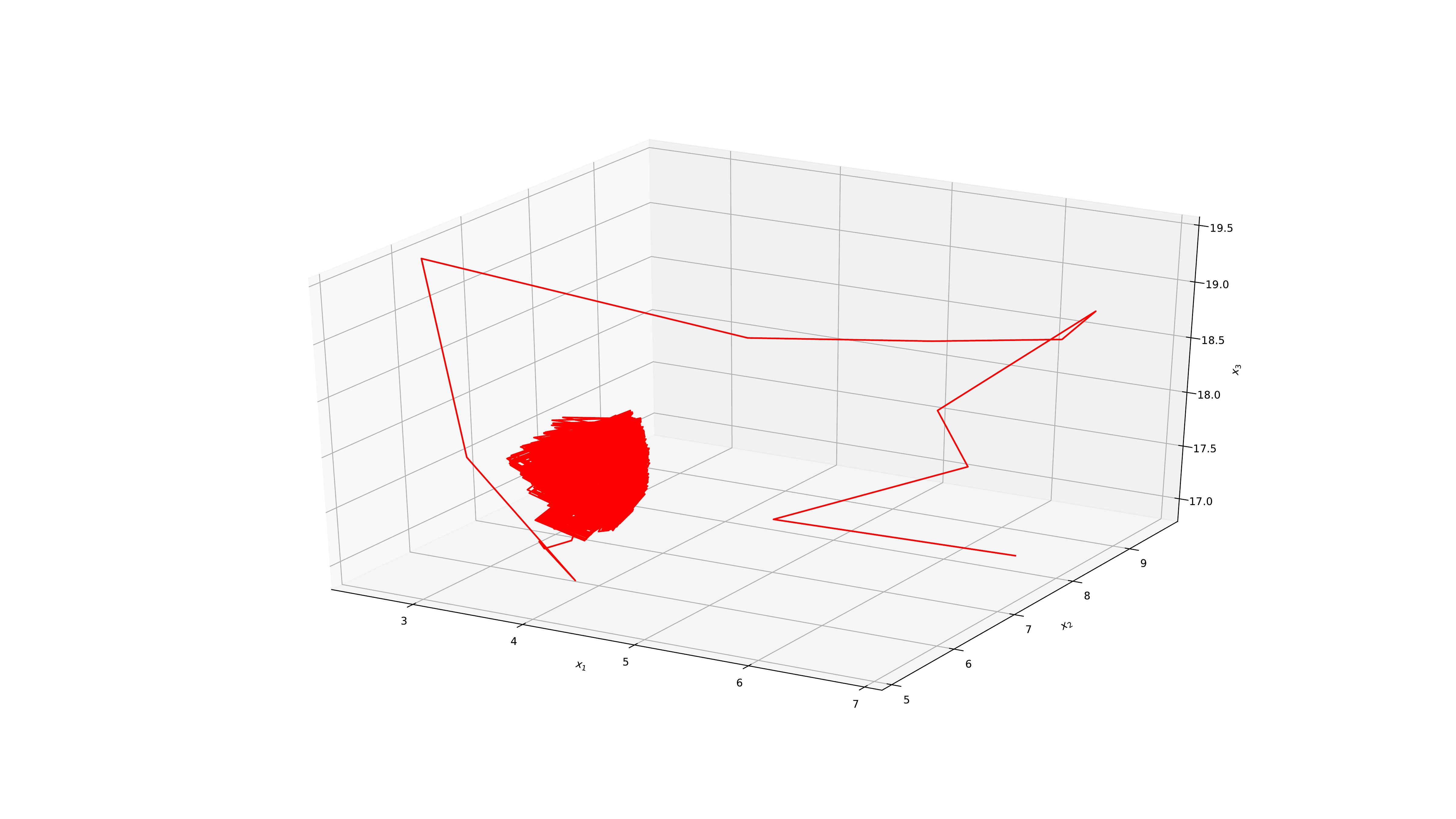}
		\includegraphics[width=\bwidth,clip, trim=100mm 40mm 80mm 40mm]{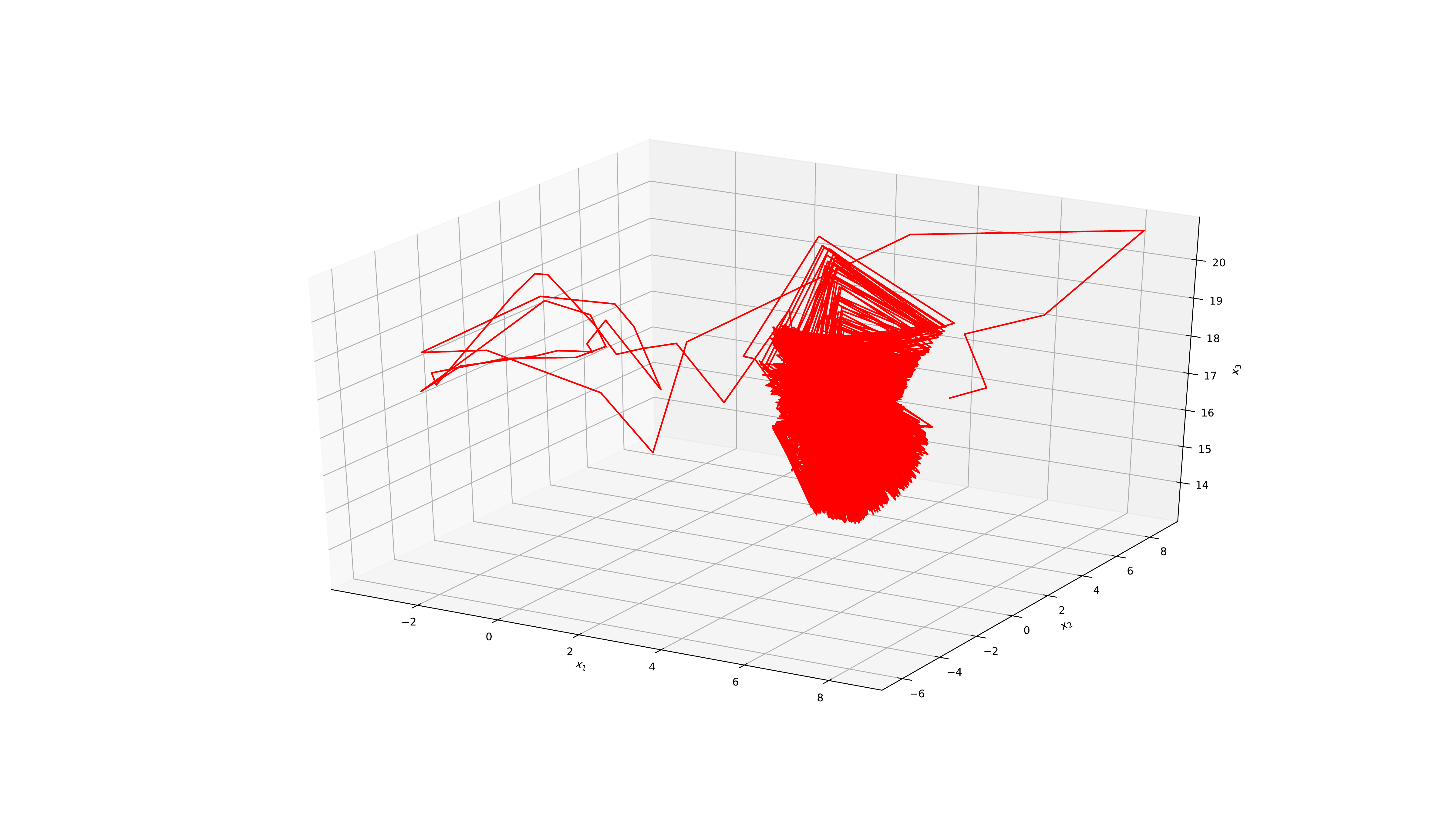}
	\end{subfigure}%
	
	\begin{subfigure}[b]{0.04\linewidth}
	    \rotatebox[origin=t]{90}{\scriptsize SR}\vspace{0.9\linewidth}
	\end{subfigure}%
	\begin{subfigure}[t]{0.96\linewidth}
		\centering
		\includegraphics[width=\bwidth,clip, trim=100mm 40mm 80mm 40mm]{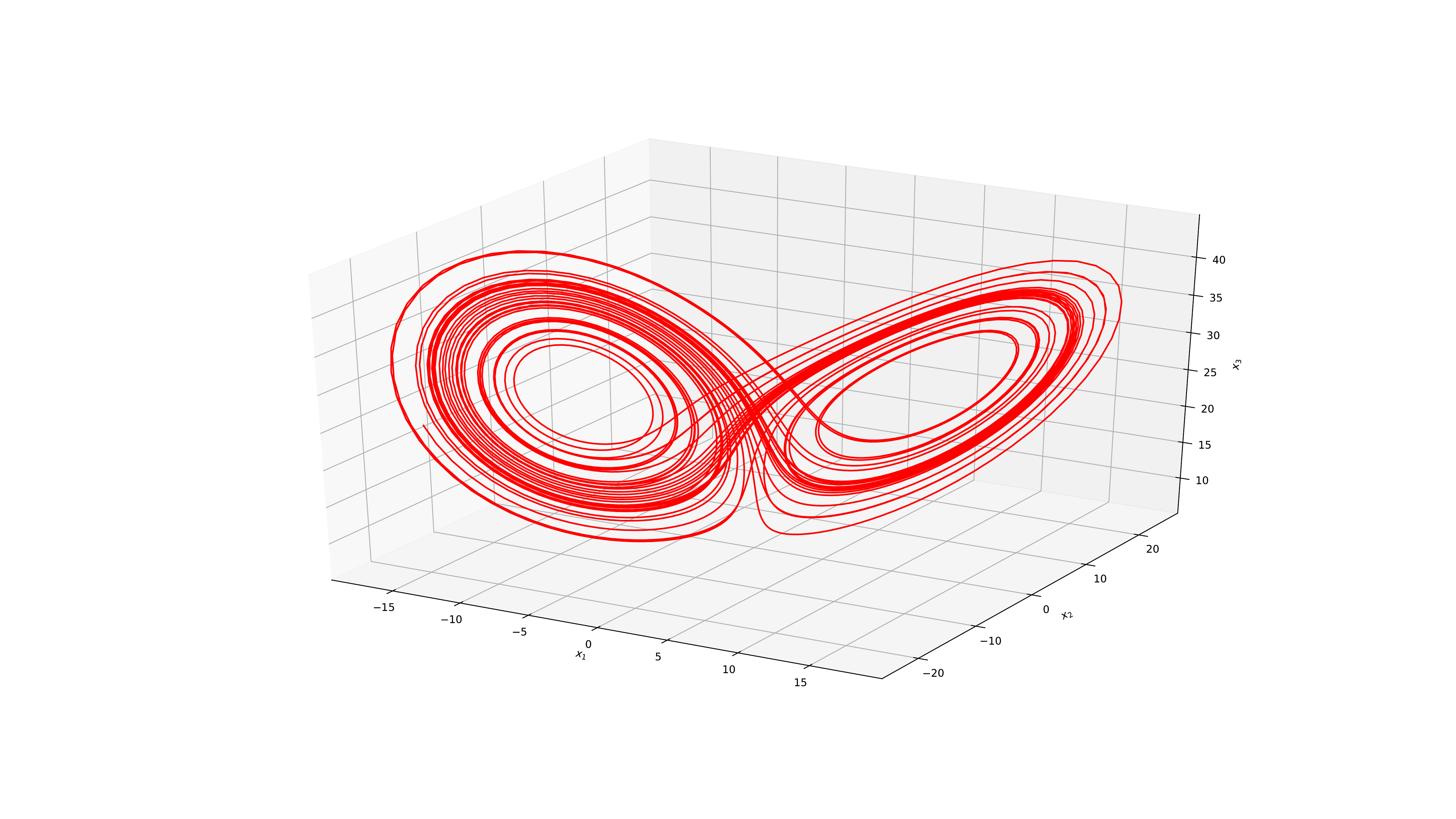}
		\includegraphics[width=\bwidth,clip, trim=100mm 40mm 80mm 40mm]{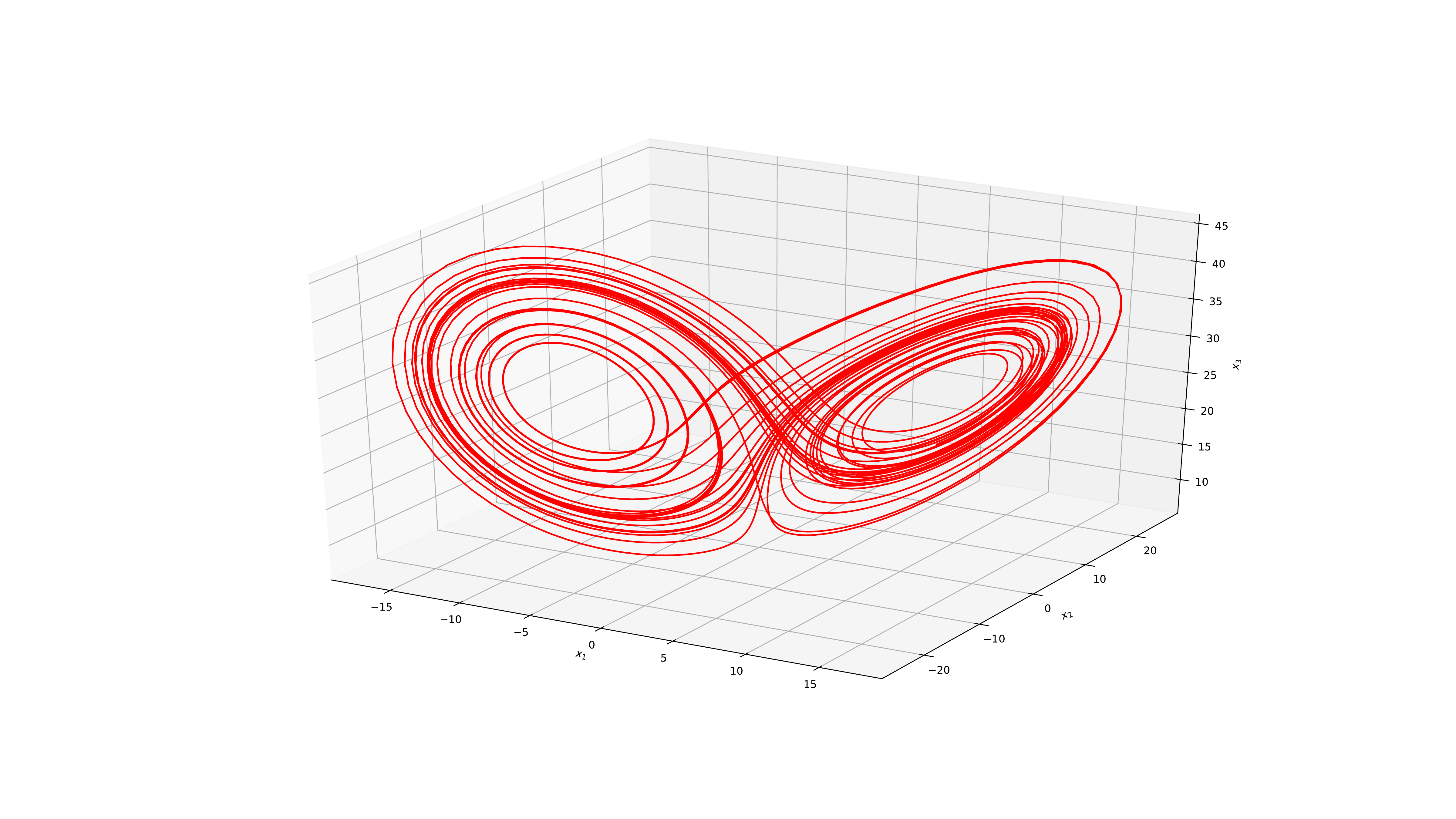}
		\includegraphics[width=\bwidth,clip, trim=100mm 40mm 80mm 40mm]{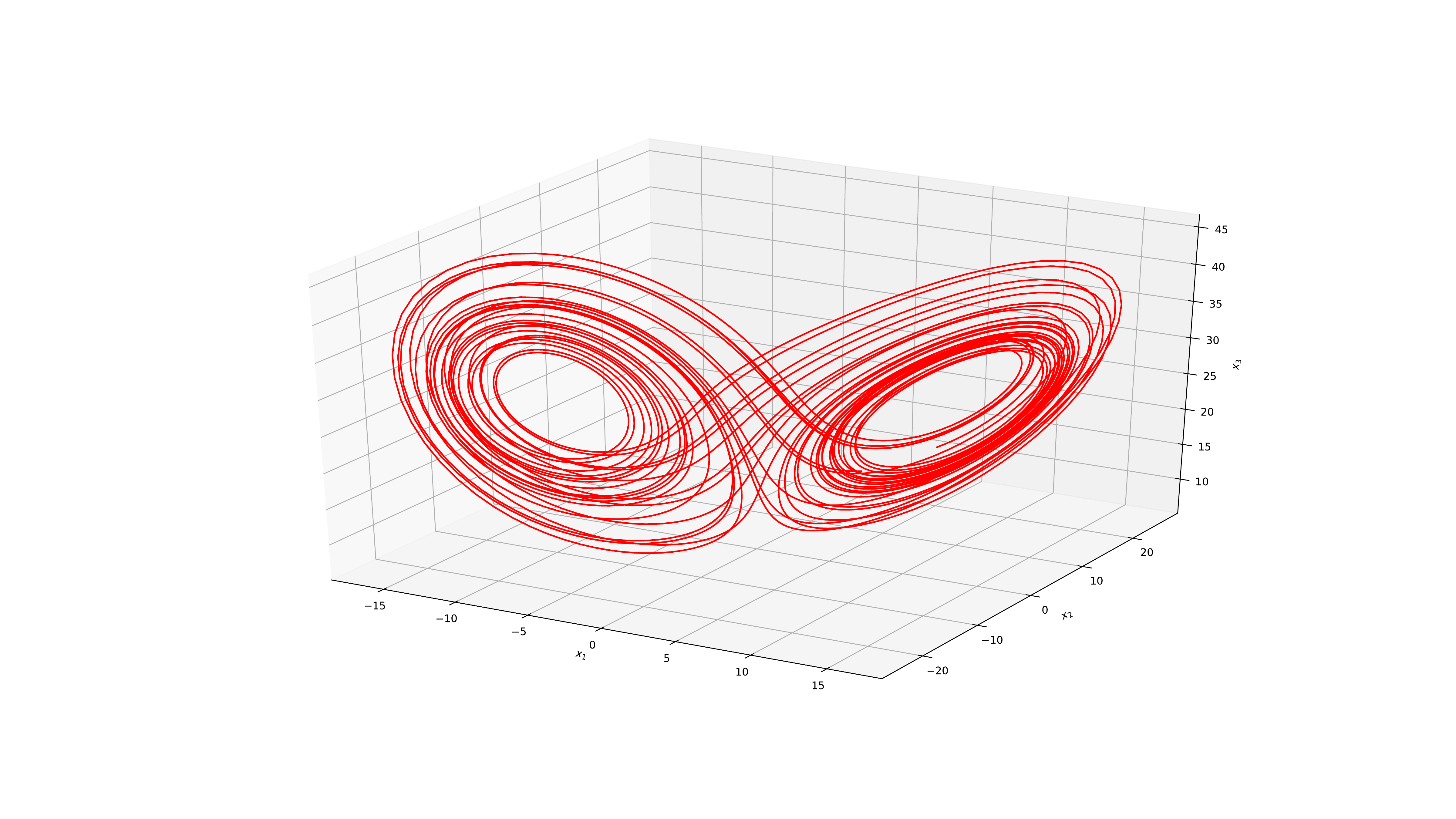}
		\includegraphics[width=\bwidth,clip, trim=100mm 40mm 80mm 40mm]{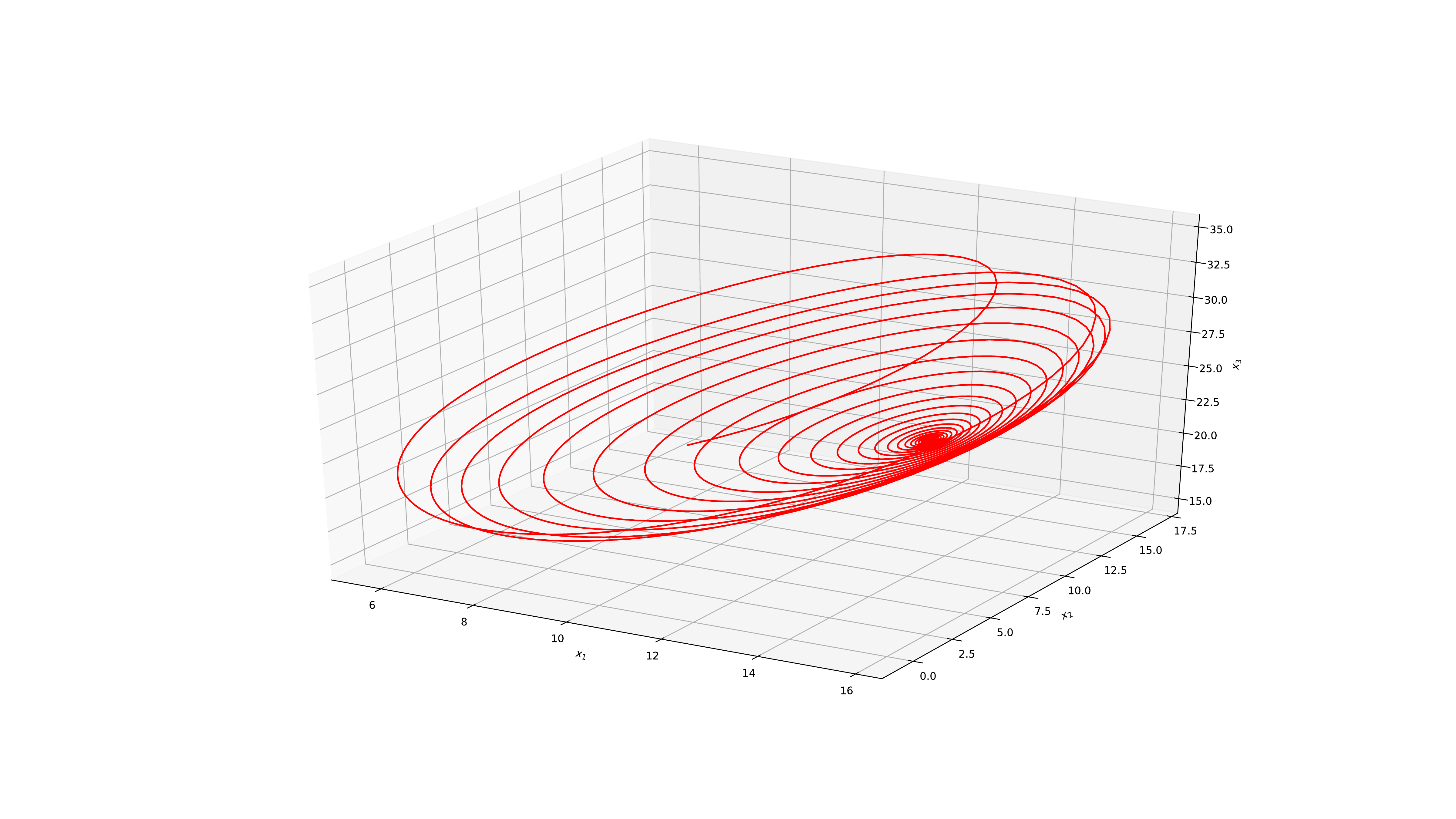}
		\includegraphics[width=\bwidth,clip, trim=100mm 40mm 80mm 40mm]{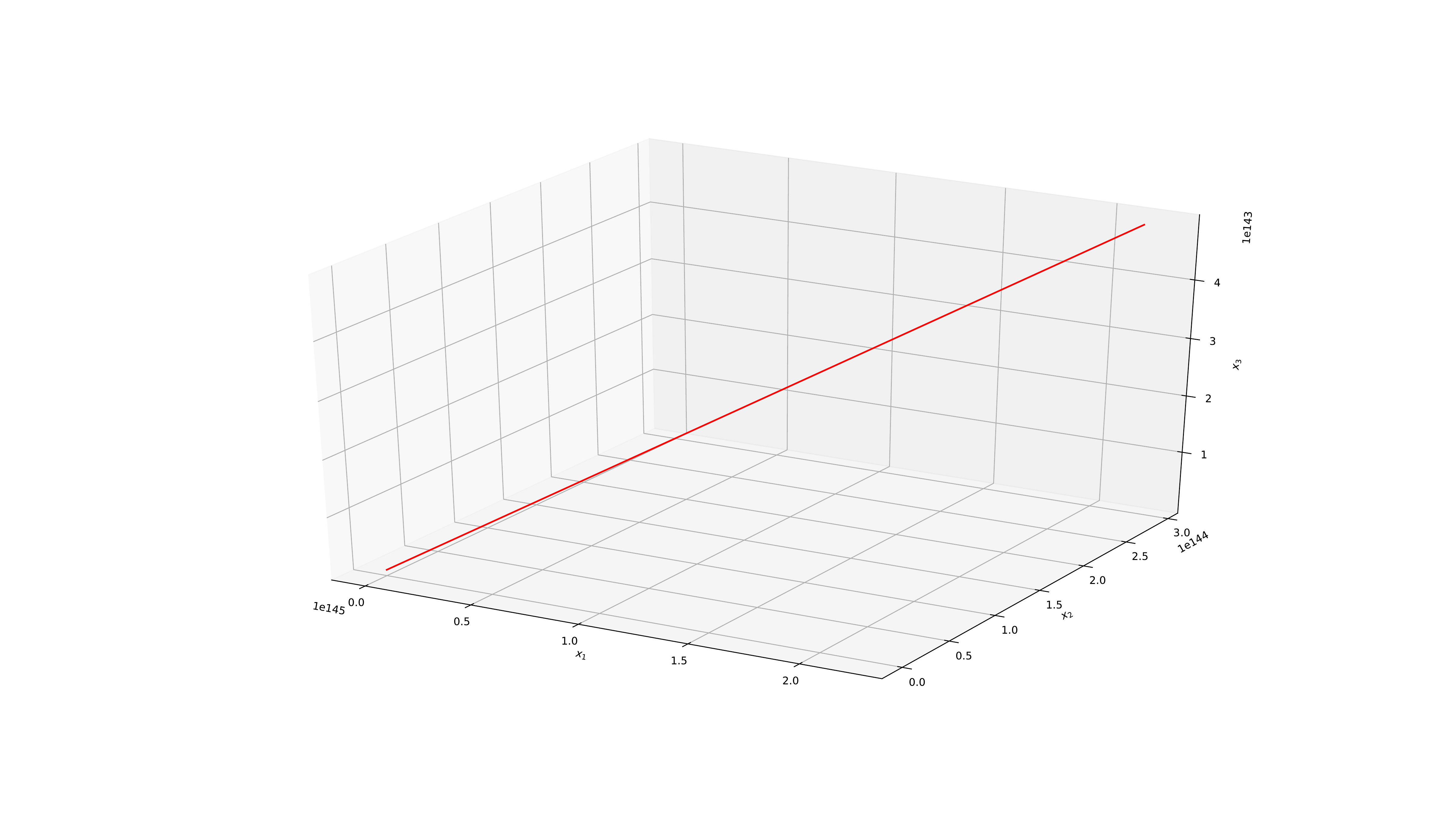}
		\includegraphics[width=\bwidth,clip, trim=100mm 40mm 80mm 40mm]{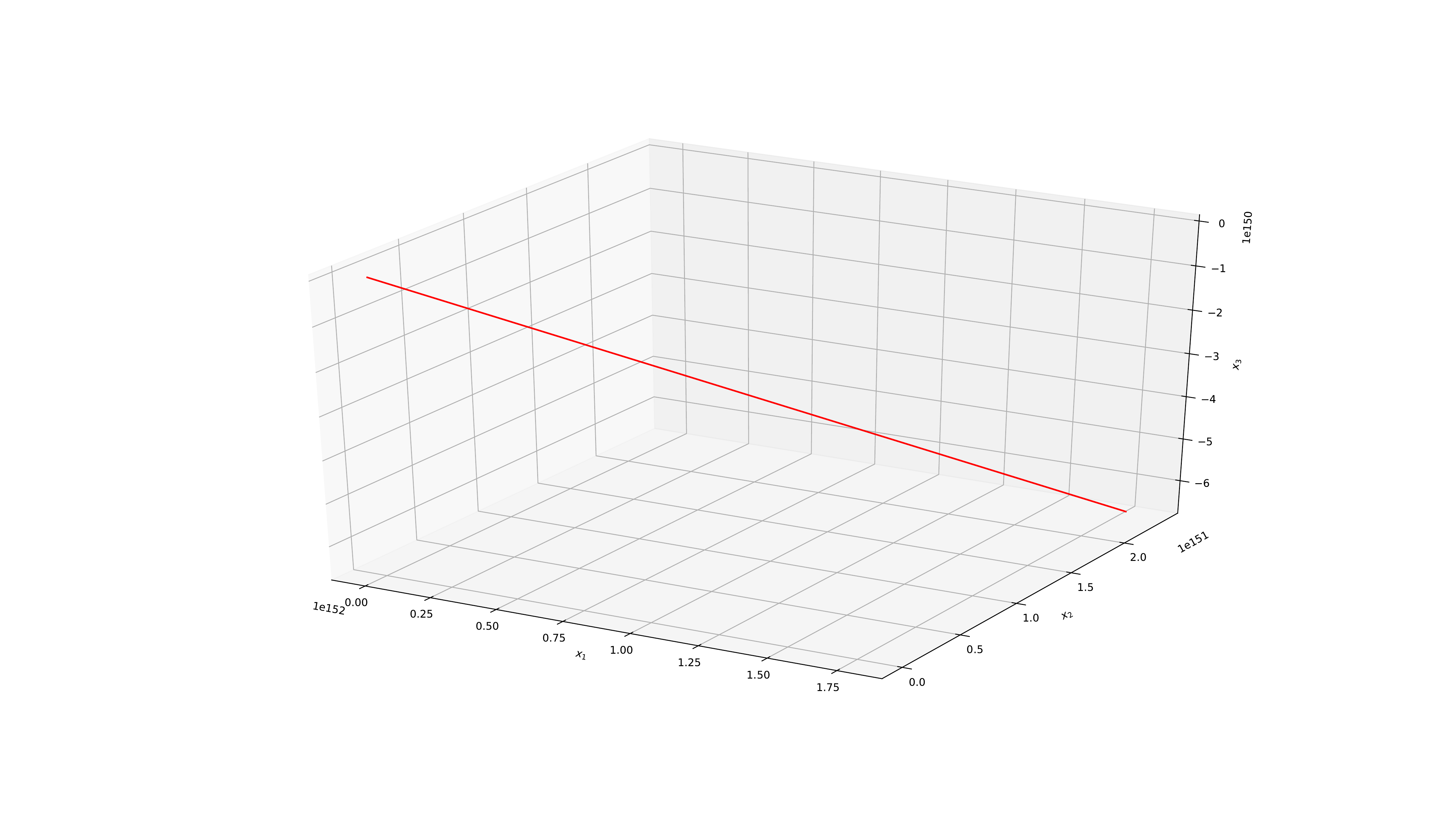}
	\end{subfigure}%
	
	\begin{subfigure}[b]{0.04\linewidth}
	    \rotatebox[origin=t]{90}{\scriptsize SR--Hann}\vspace{0.3\linewidth}
	\end{subfigure}%
	\begin{subfigure}[t]{0.96\linewidth}
		\centering
		\includegraphics[width=\bwidth,clip, trim=100mm 40mm 80mm 40mm]{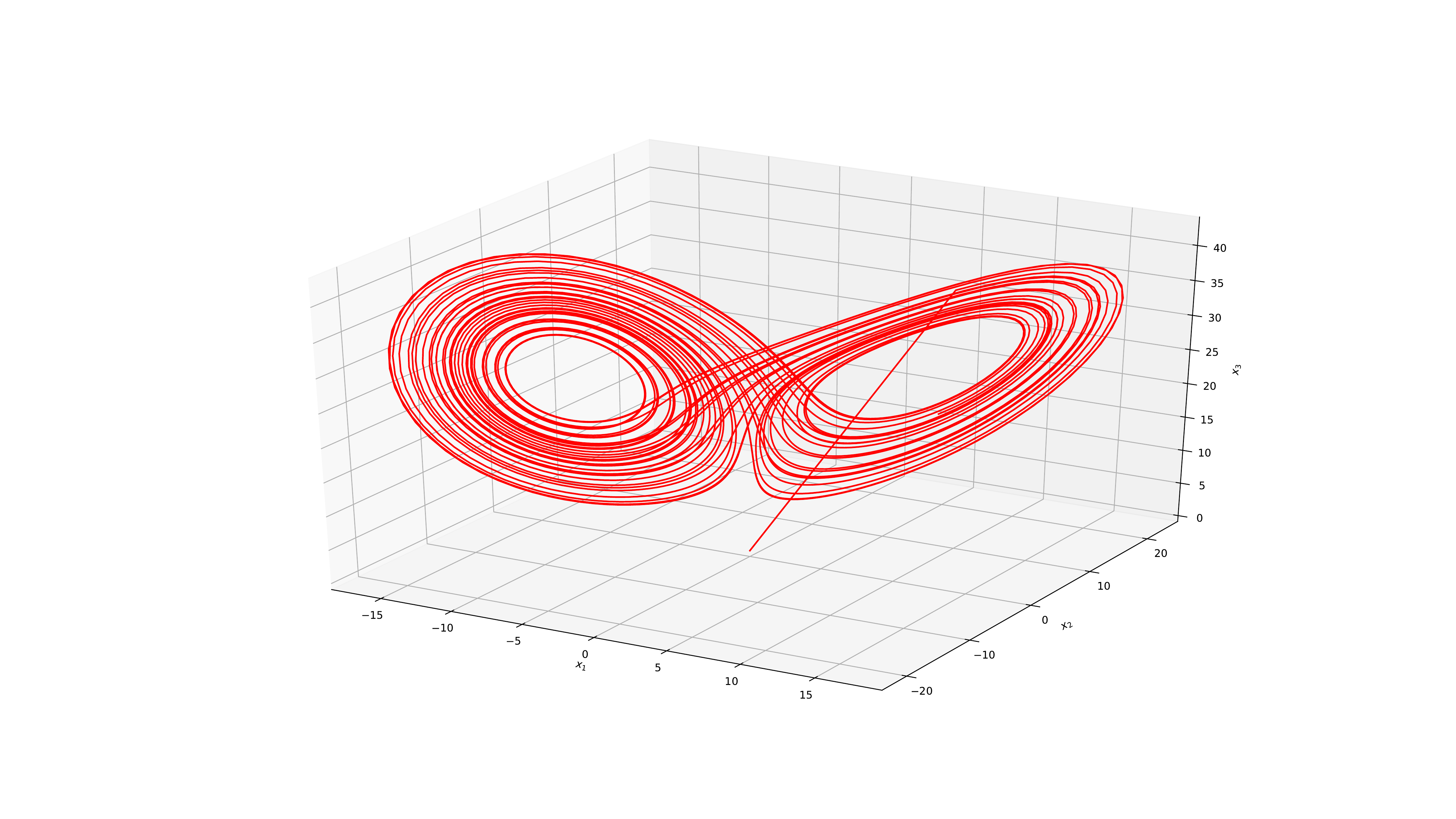}
		\includegraphics[width=\bwidth,clip, trim=100mm 40mm 80mm 40mm]{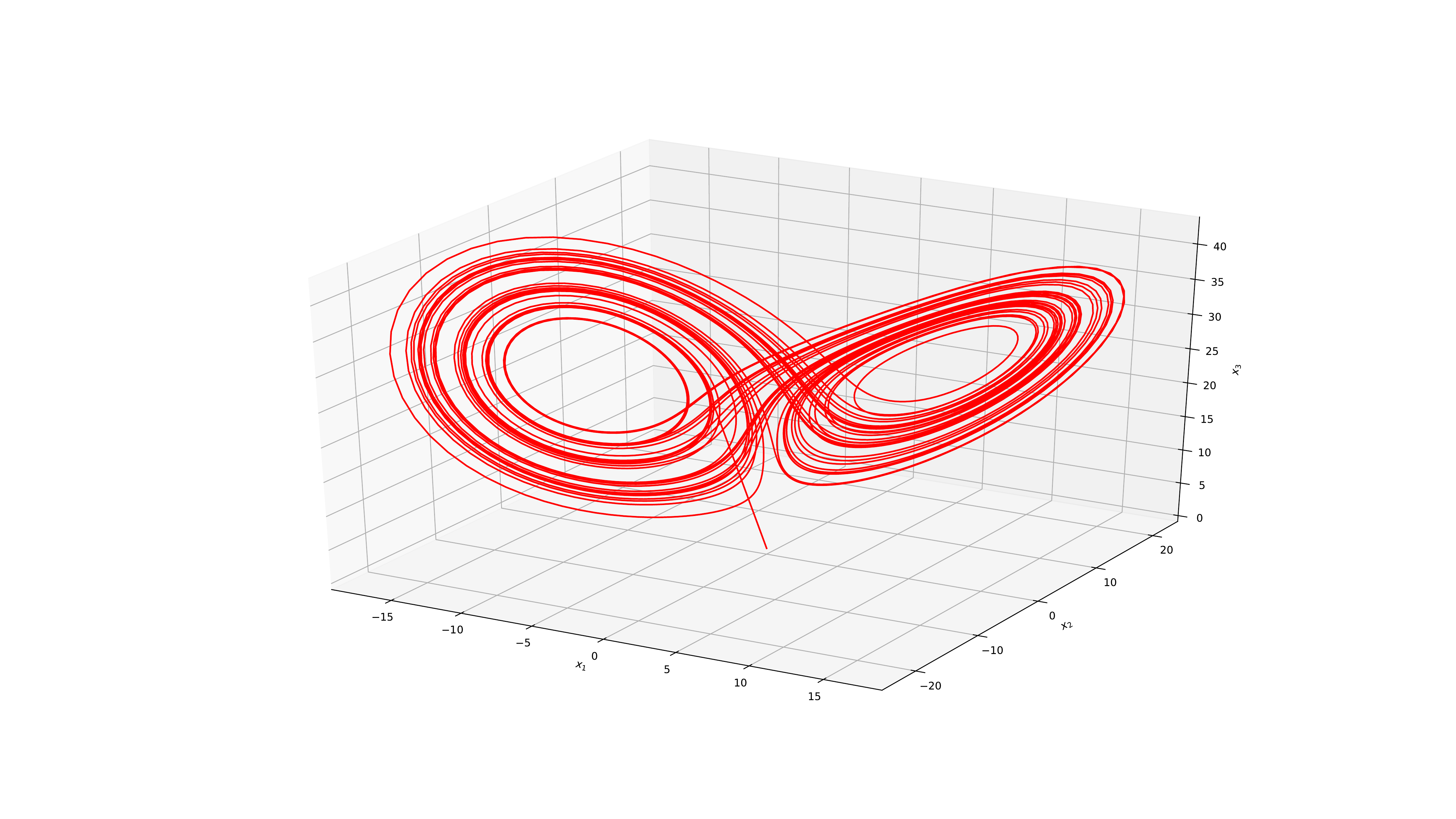}
		\includegraphics[width=\bwidth,clip, trim=100mm 40mm 80mm 40mm]{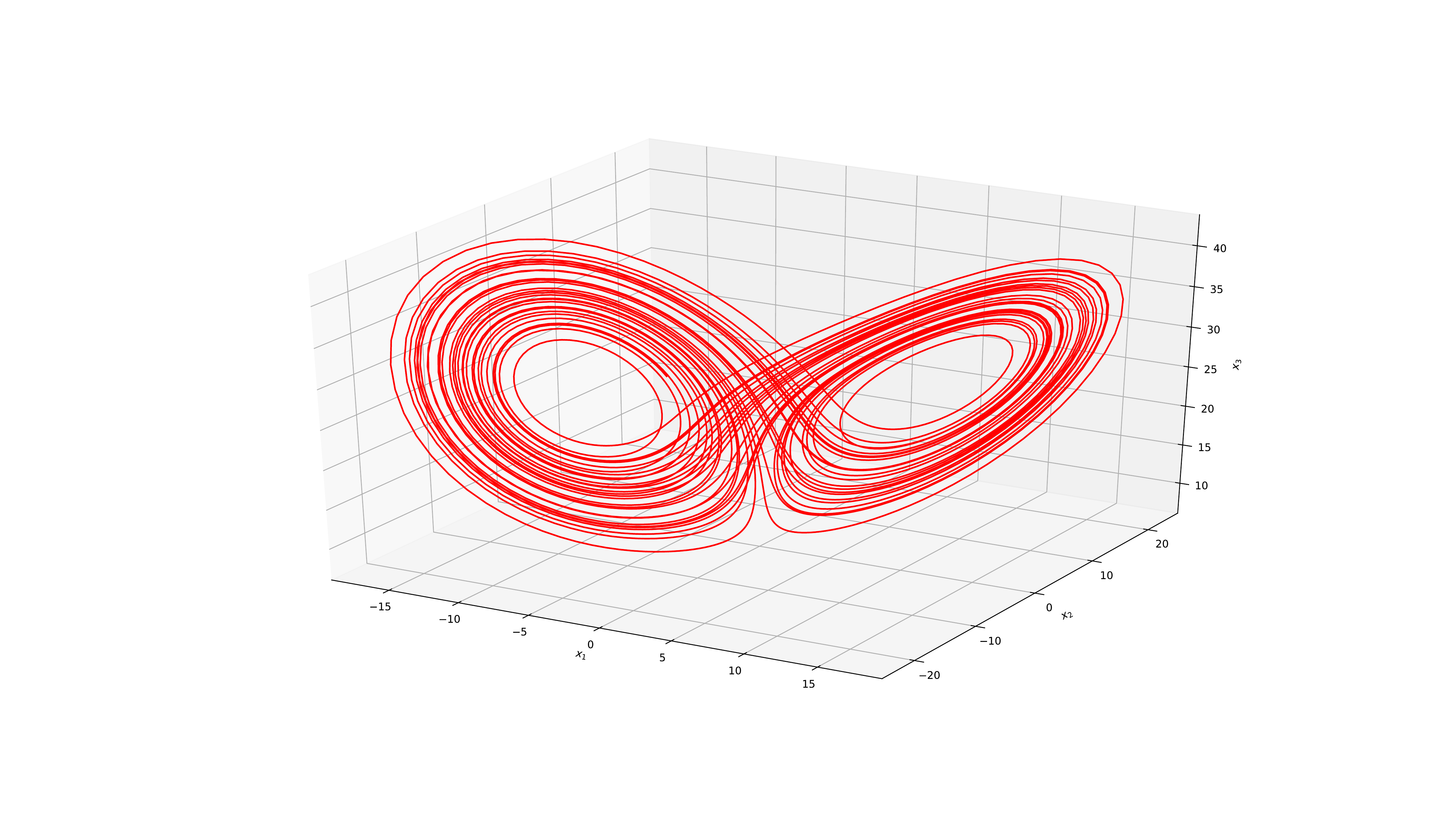}
		\includegraphics[width=\bwidth,clip, trim=100mm 40mm 80mm 40mm]{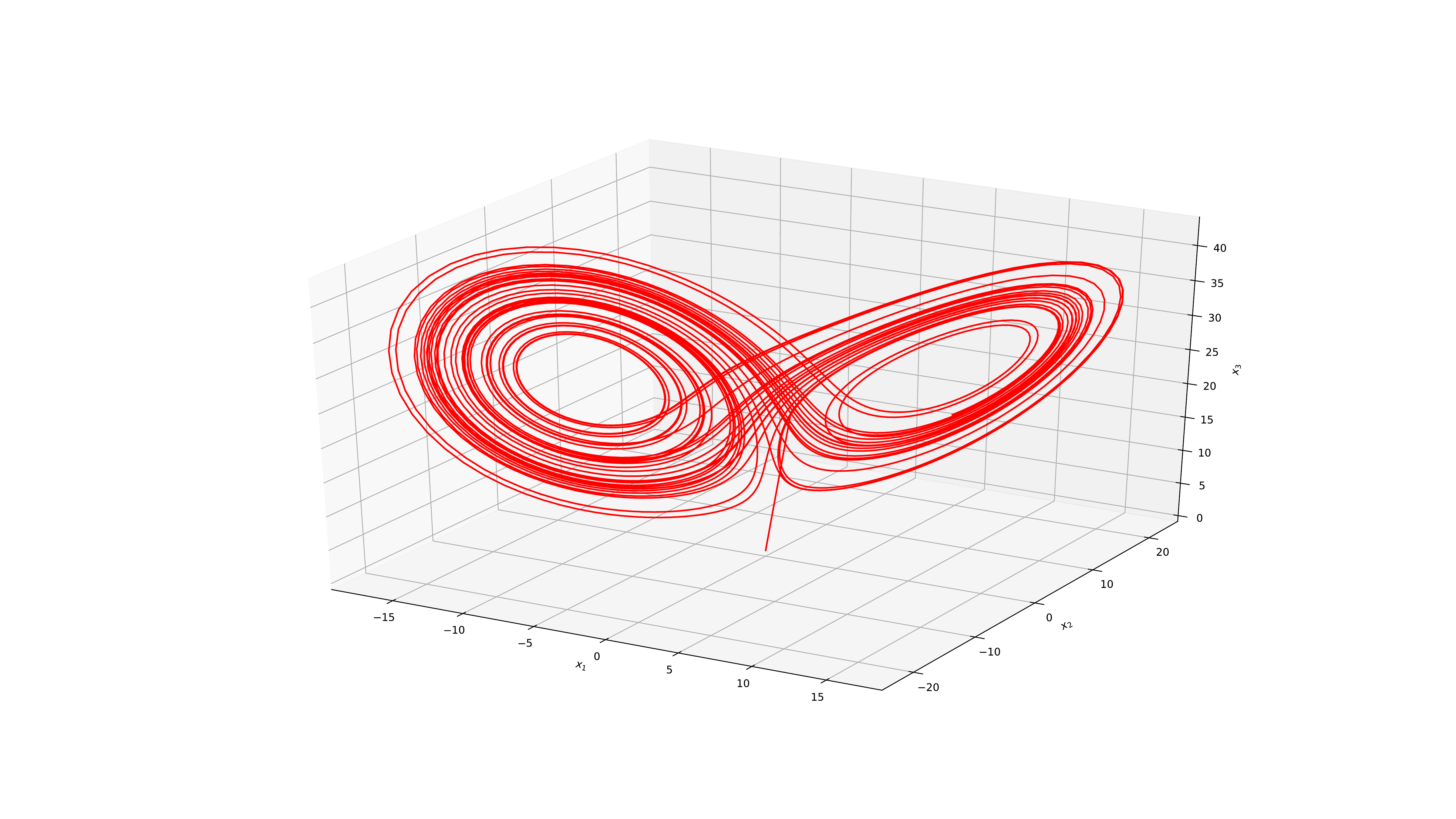}
		\includegraphics[width=\bwidth,clip, trim=100mm 40mm 80mm 40mm]{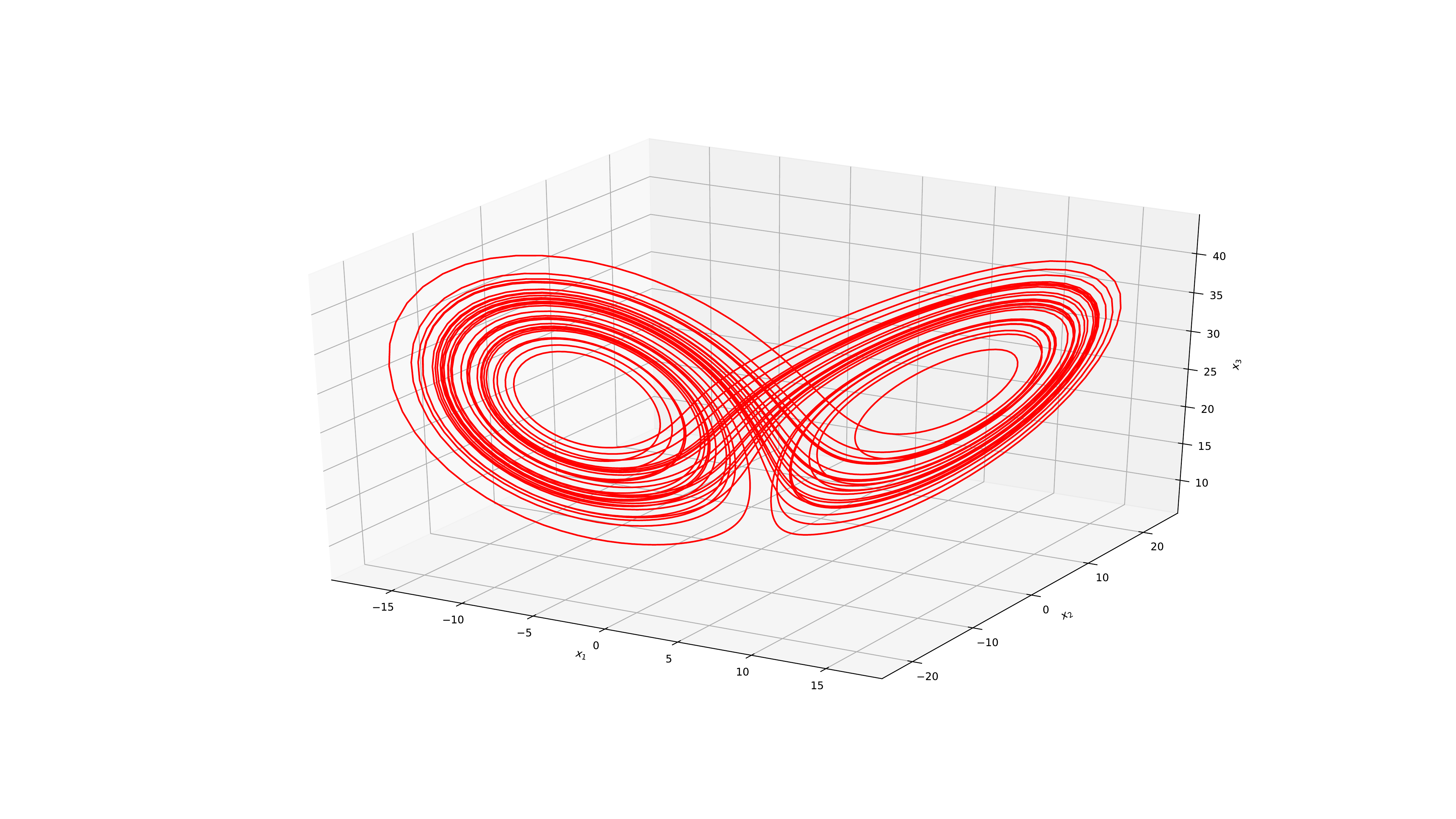}
		\includegraphics[width=\bwidth,clip, trim=100mm 40mm 80mm 40mm]{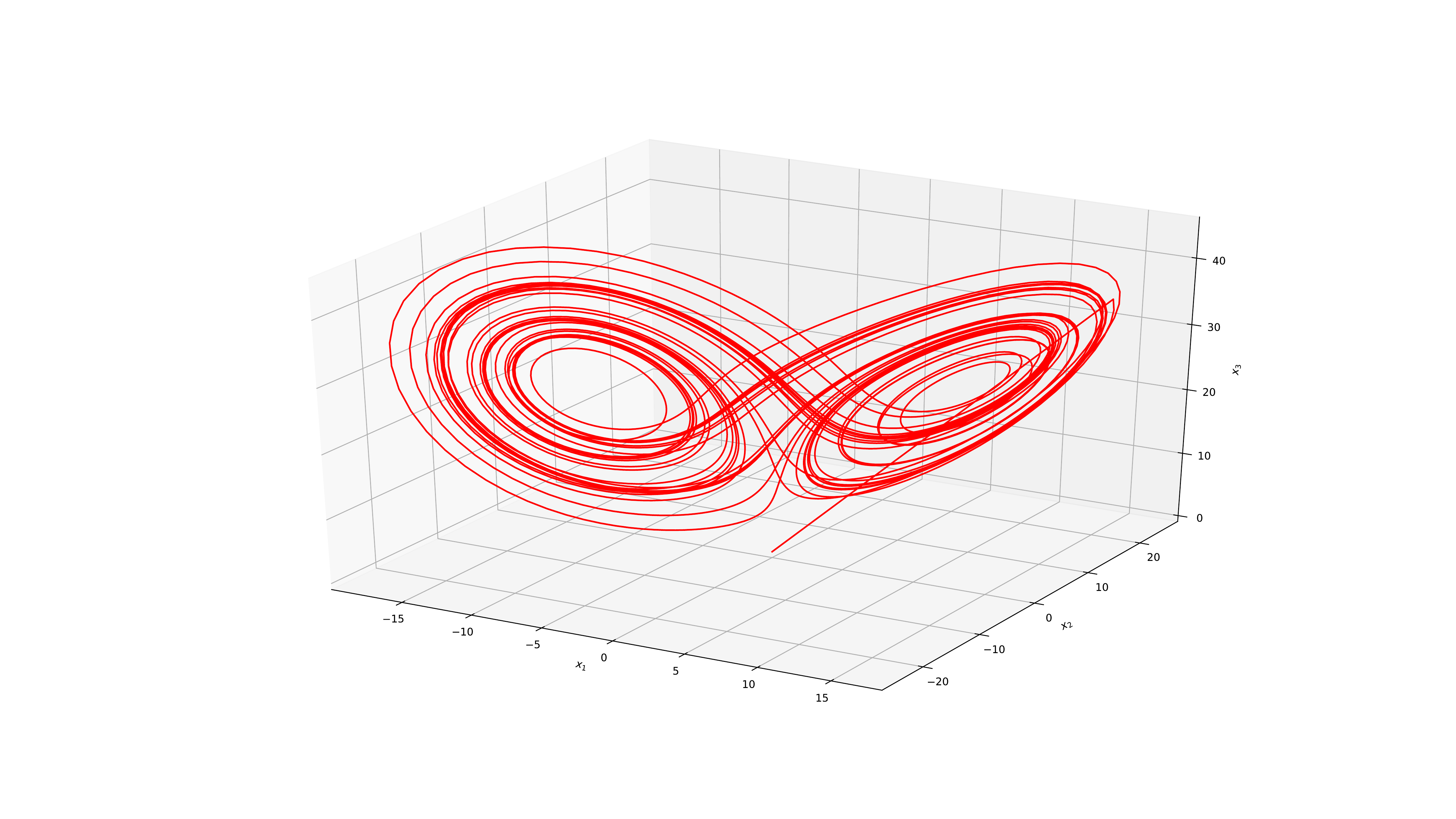}
	\end{subfigure}%

	\begin{subfigure}[b]{0.04\linewidth}
	    \rotatebox[origin=t]{90}{\scriptsize BiNN}\vspace{0.2\linewidth}
	\end{subfigure}%
	\begin{subfigure}[t]{0.96\linewidth}
		\centering
		\includegraphics[width=\bwidth,clip, trim=100mm 40mm 80mm 40mm]{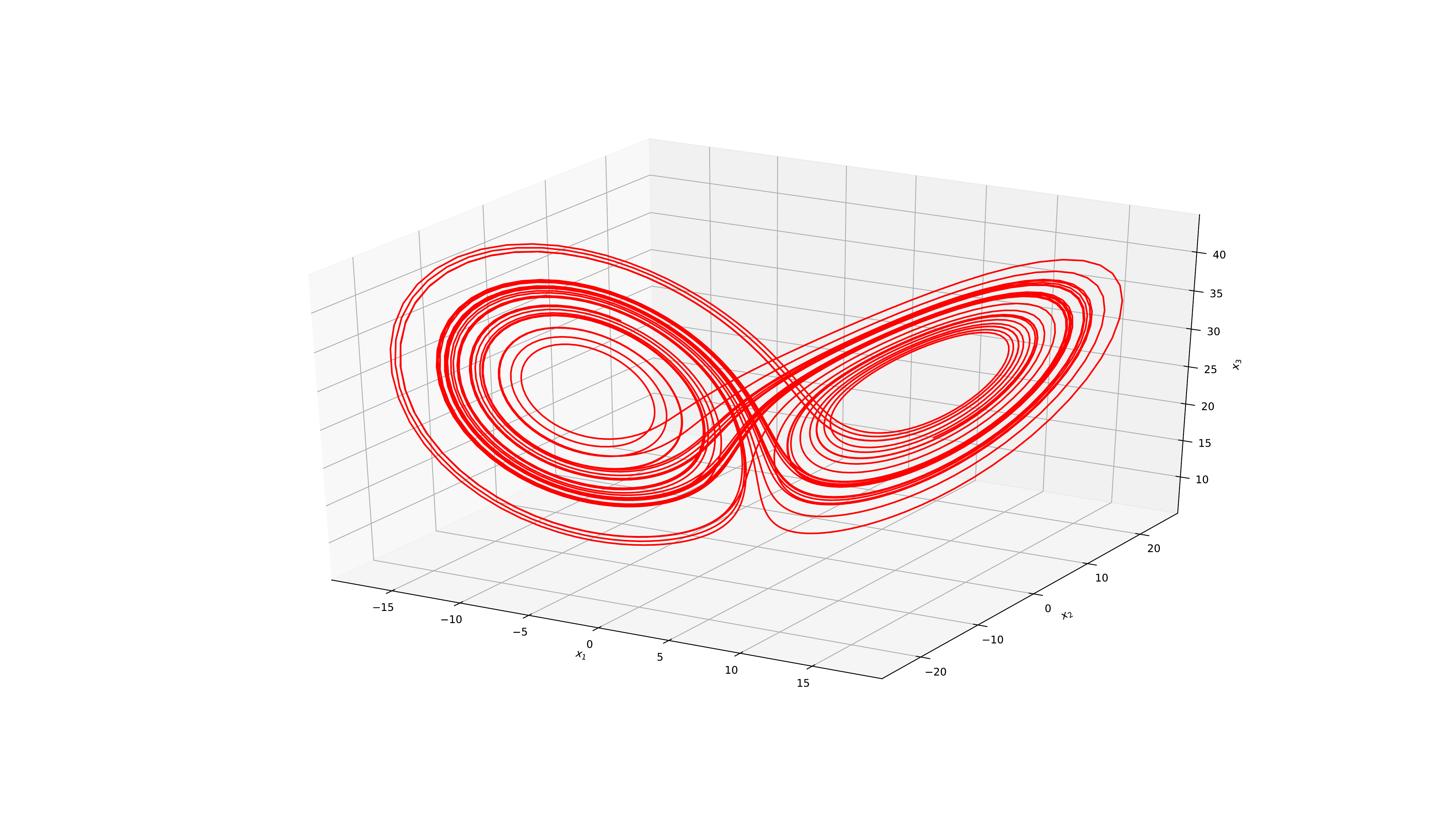}
		\includegraphics[width=\bwidth,clip, trim=100mm 40mm 80mm 40mm]{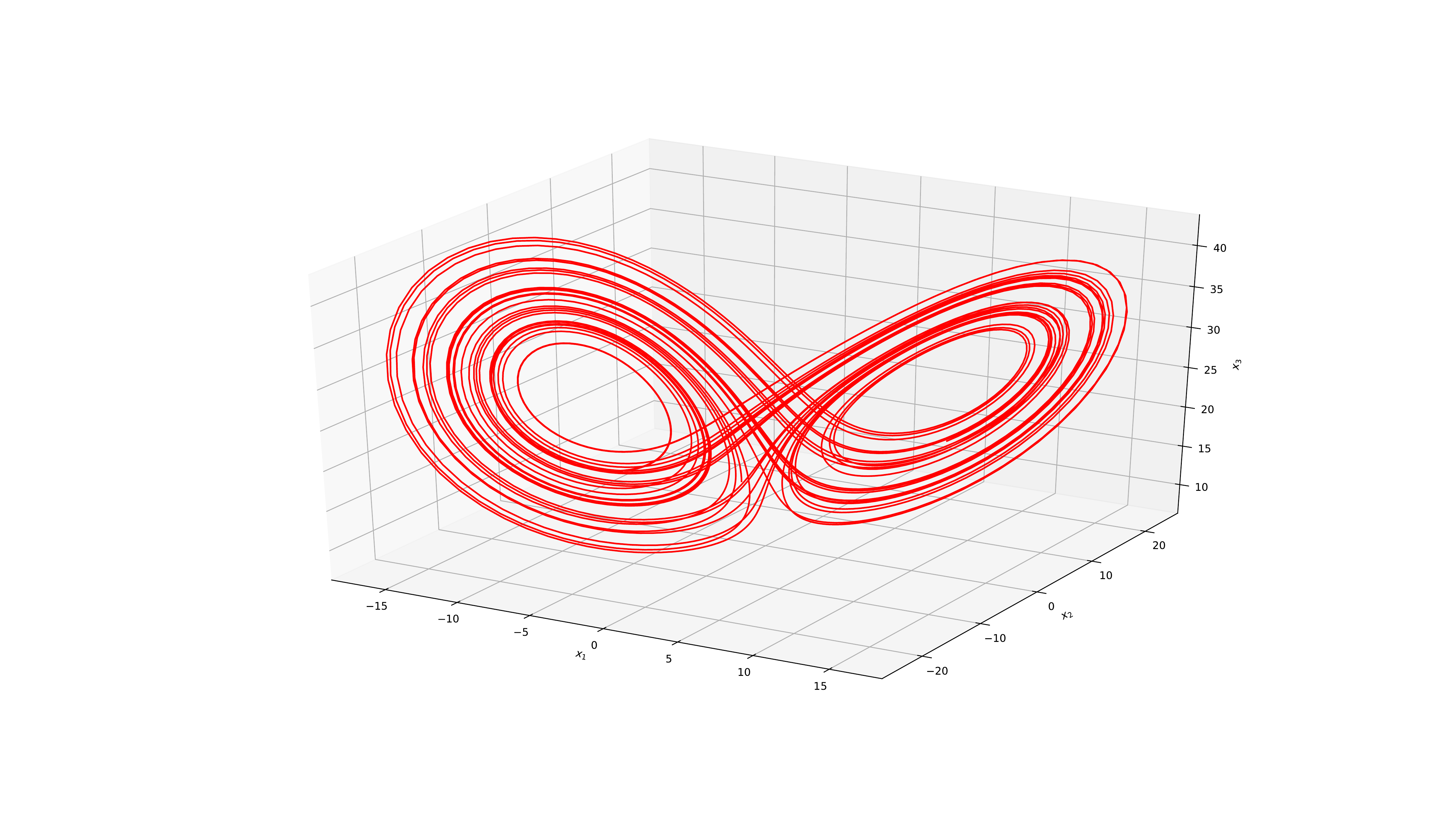}
		\includegraphics[width=\bwidth,clip, trim=100mm 40mm 80mm 40mm]{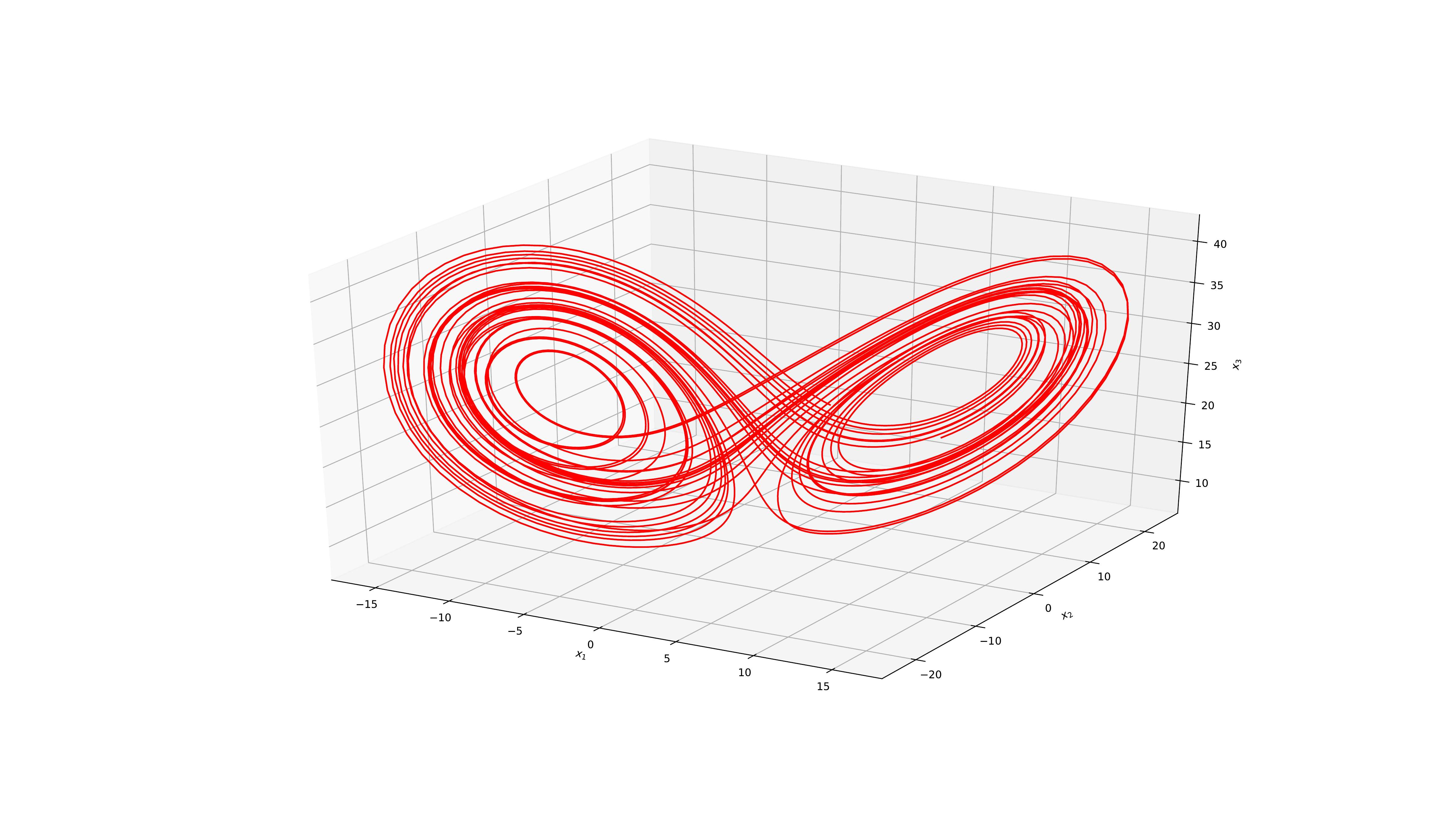}
		\includegraphics[width=\bwidth,clip, trim=100mm 40mm 80mm 40mm]{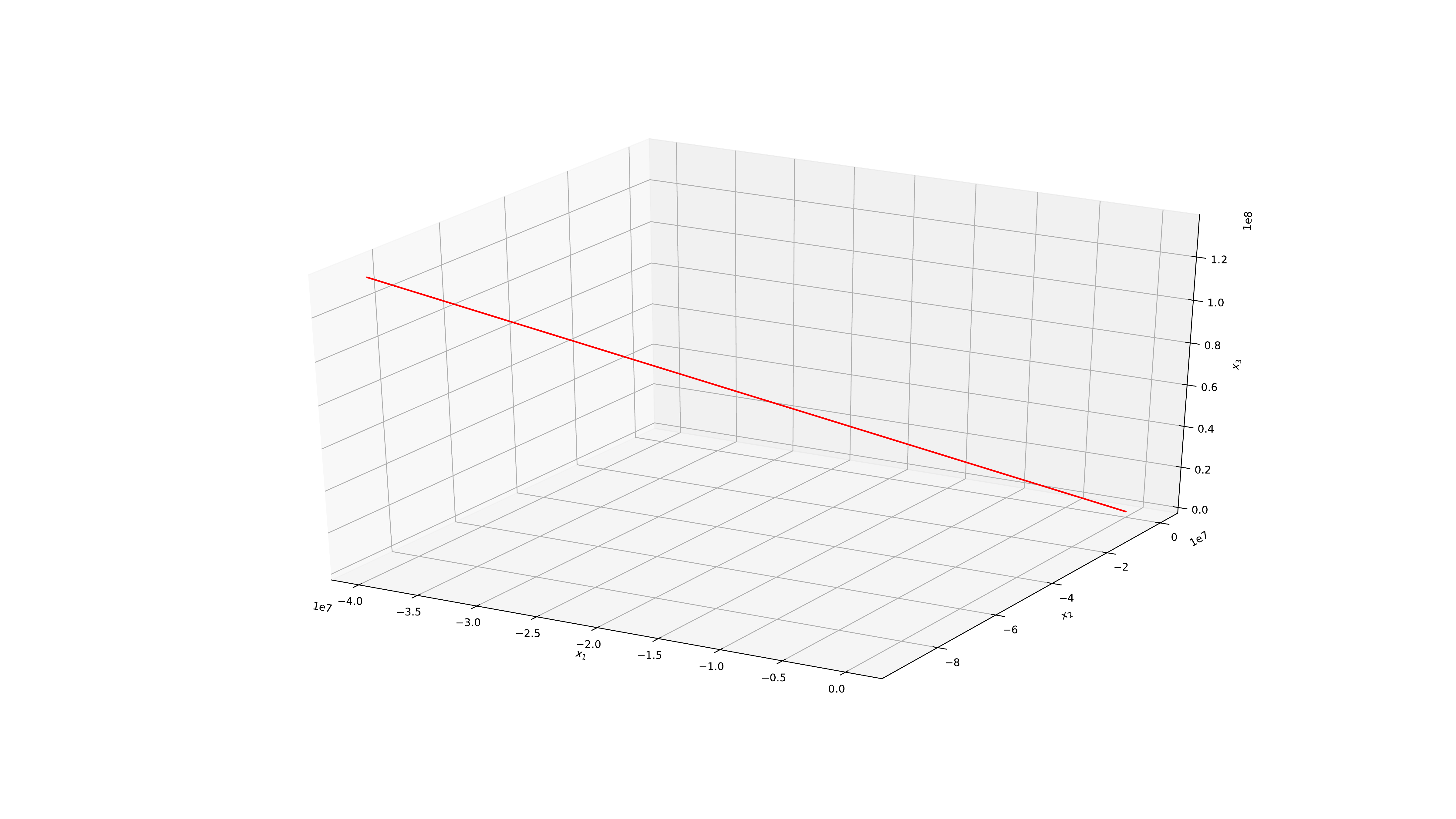}
		\includegraphics[width=\bwidth,clip, trim=100mm 40mm 80mm 40mm]{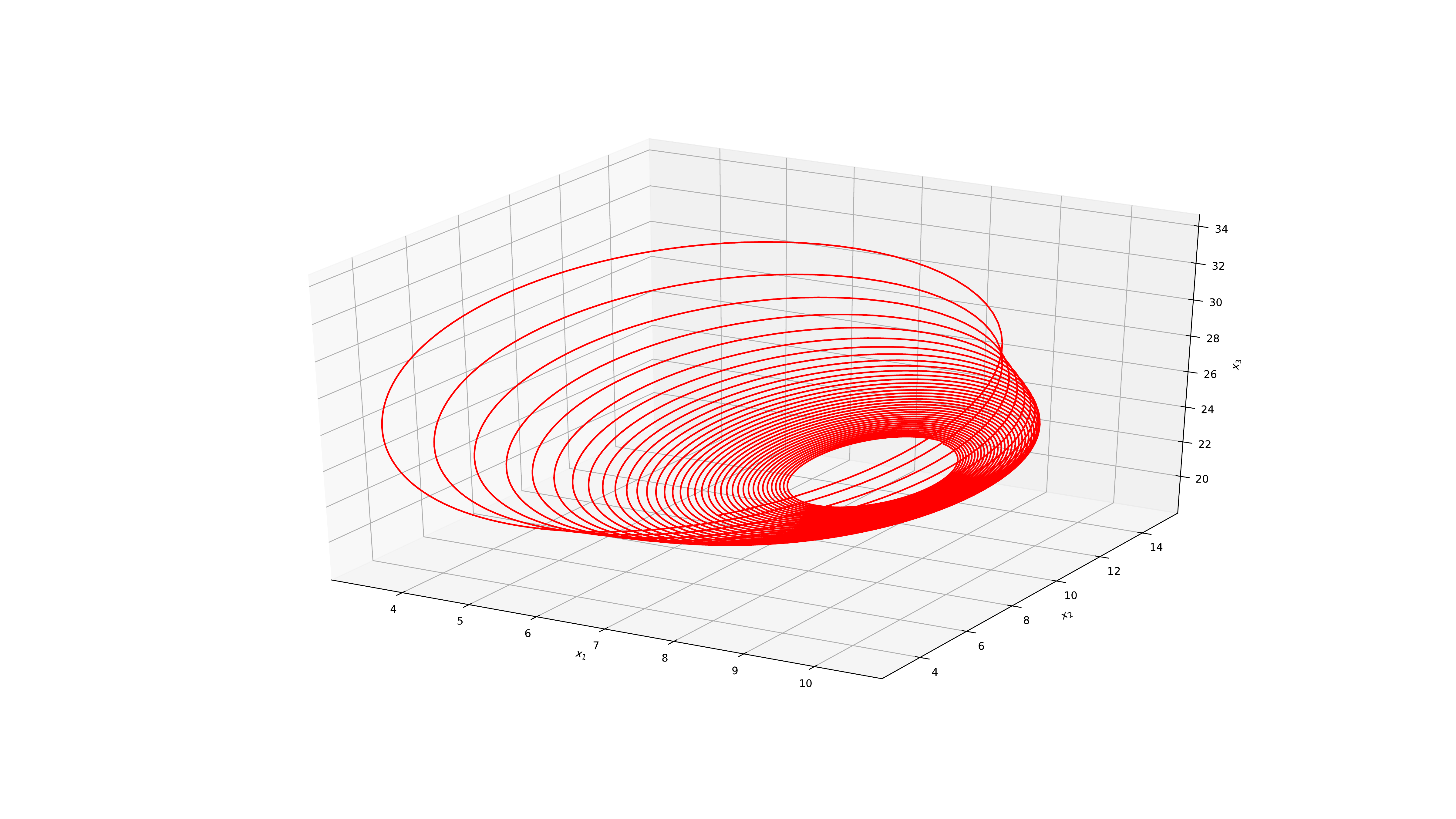}
		\includegraphics[width=\bwidth,clip, trim=100mm 40mm 80mm 40mm]{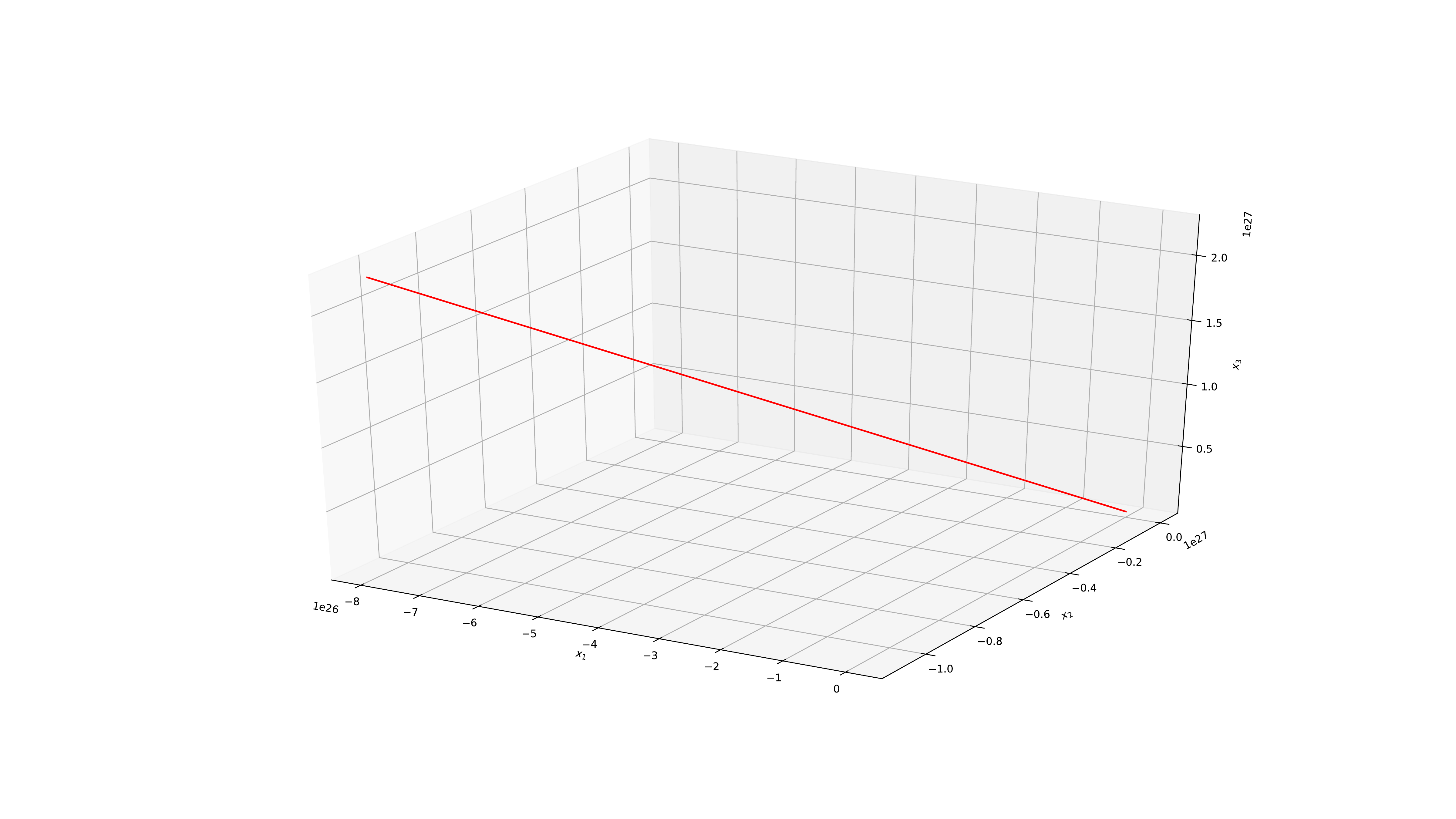}
	\end{subfigure}%

	\begin{subfigure}[b]{0.04\linewidth}
	    \rotatebox[origin=t]{90}{\scriptsize EnKS-EM}\vspace{0.2\linewidth}
	\end{subfigure}%
	\begin{subfigure}[t]{0.96\linewidth}
		\centering
		\includegraphics[width=\bwidth,clip, trim=100mm 40mm 80mm 40mm]{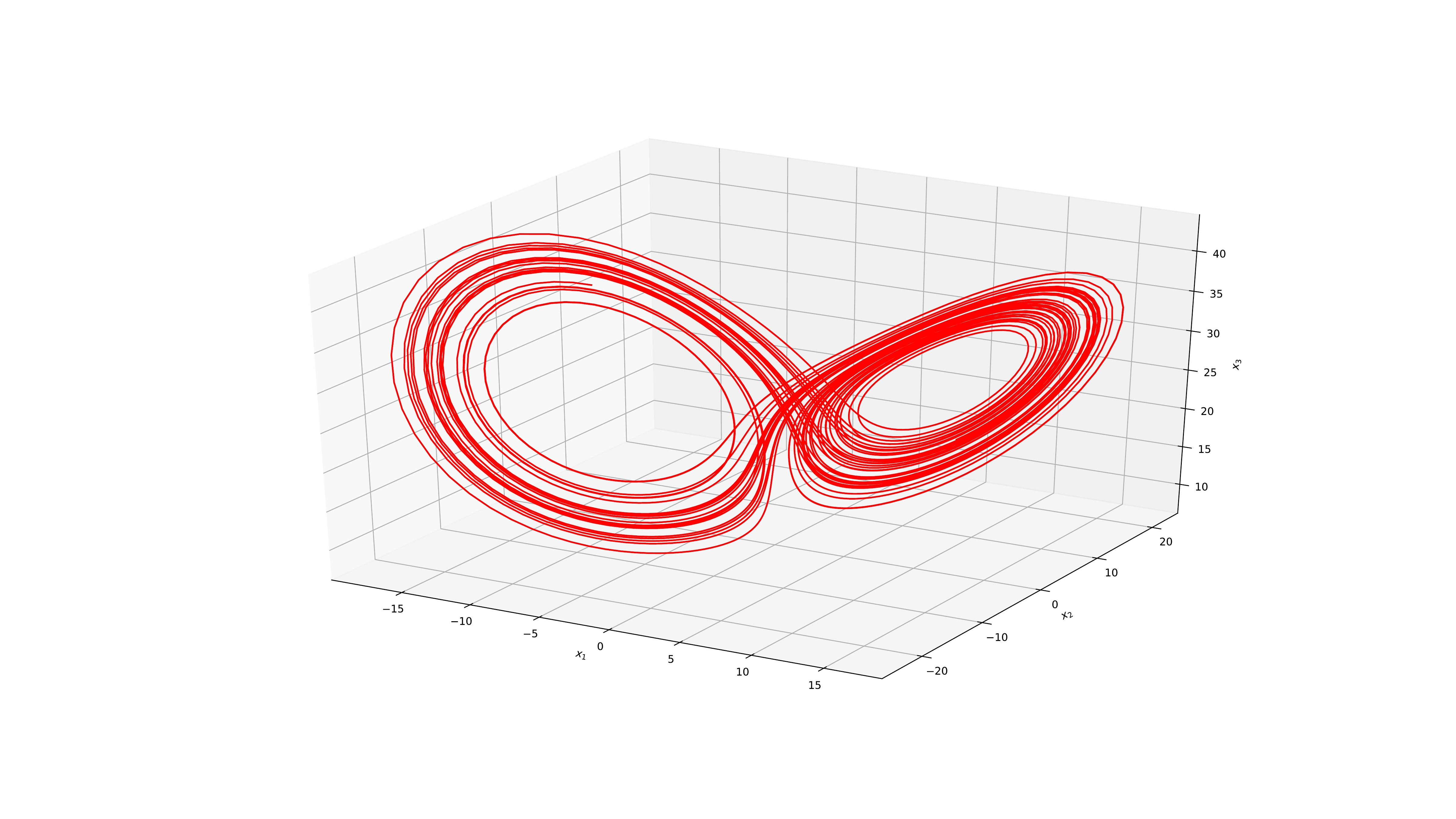}
		\includegraphics[width=\bwidth,clip, trim=100mm 40mm 80mm 40mm]{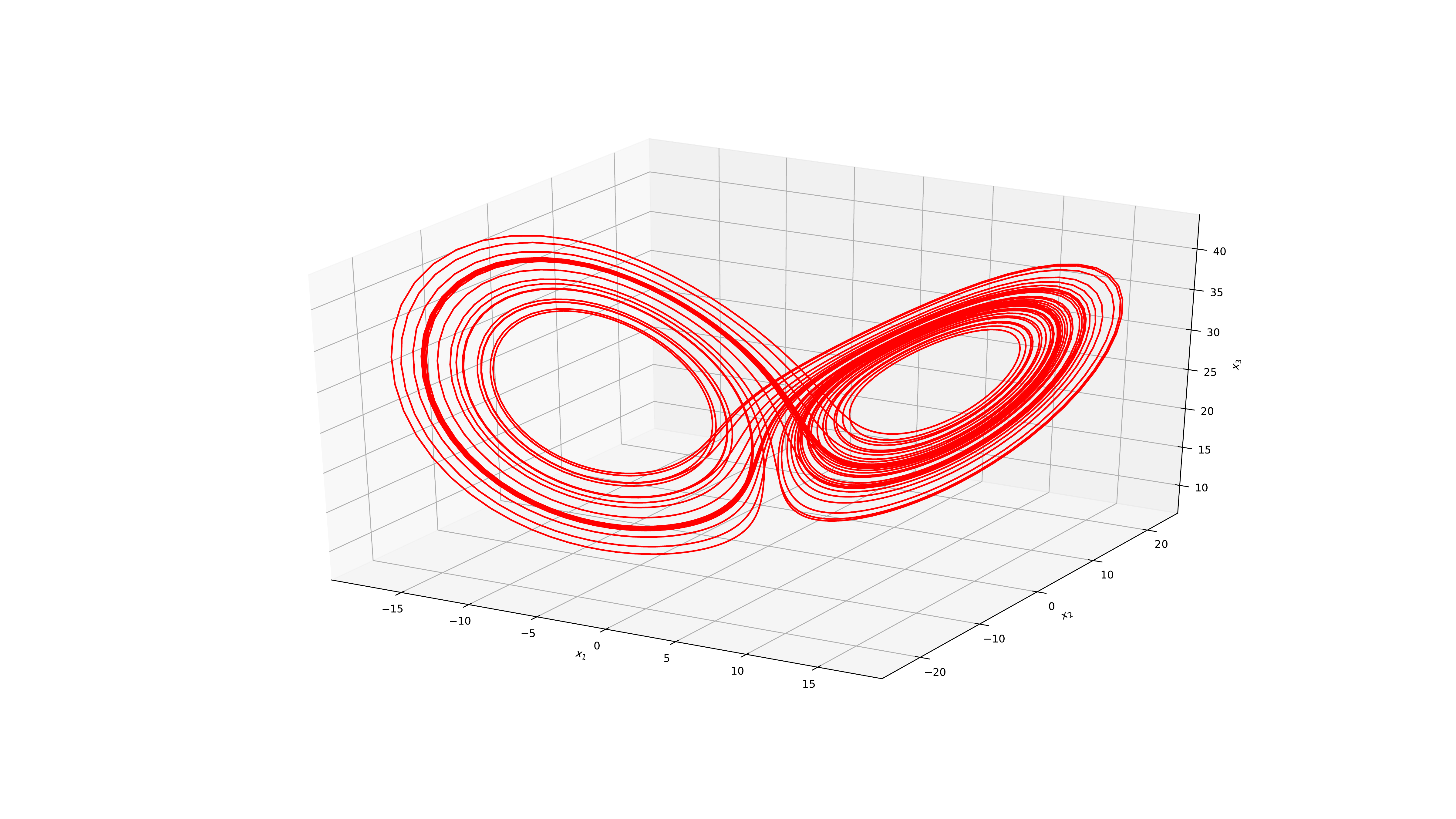}
		\includegraphics[width=\bwidth,clip, trim=100mm 40mm 80mm 40mm]{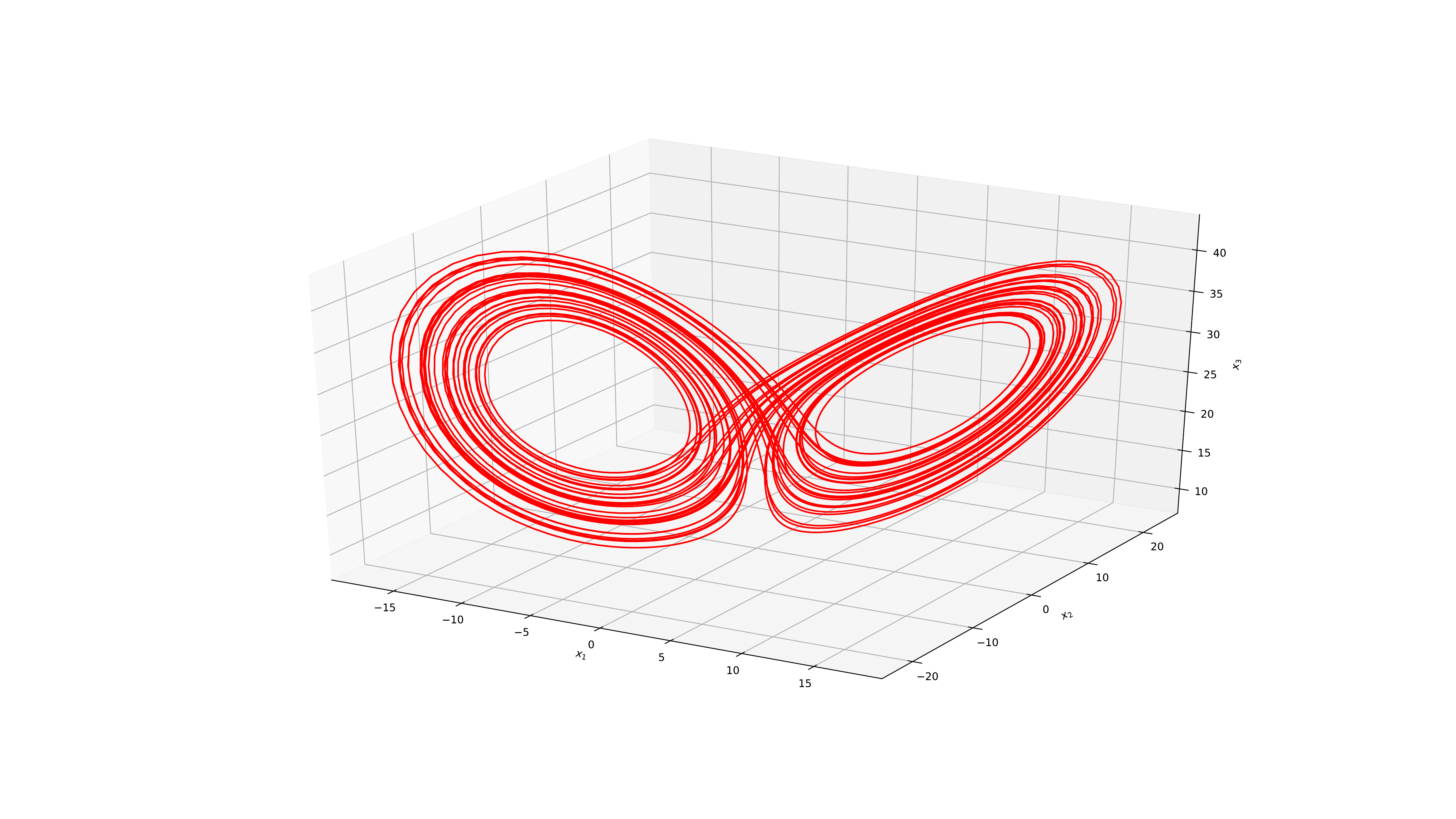}
		\includegraphics[width=\bwidth,clip, trim=100mm 40mm 80mm 40mm]{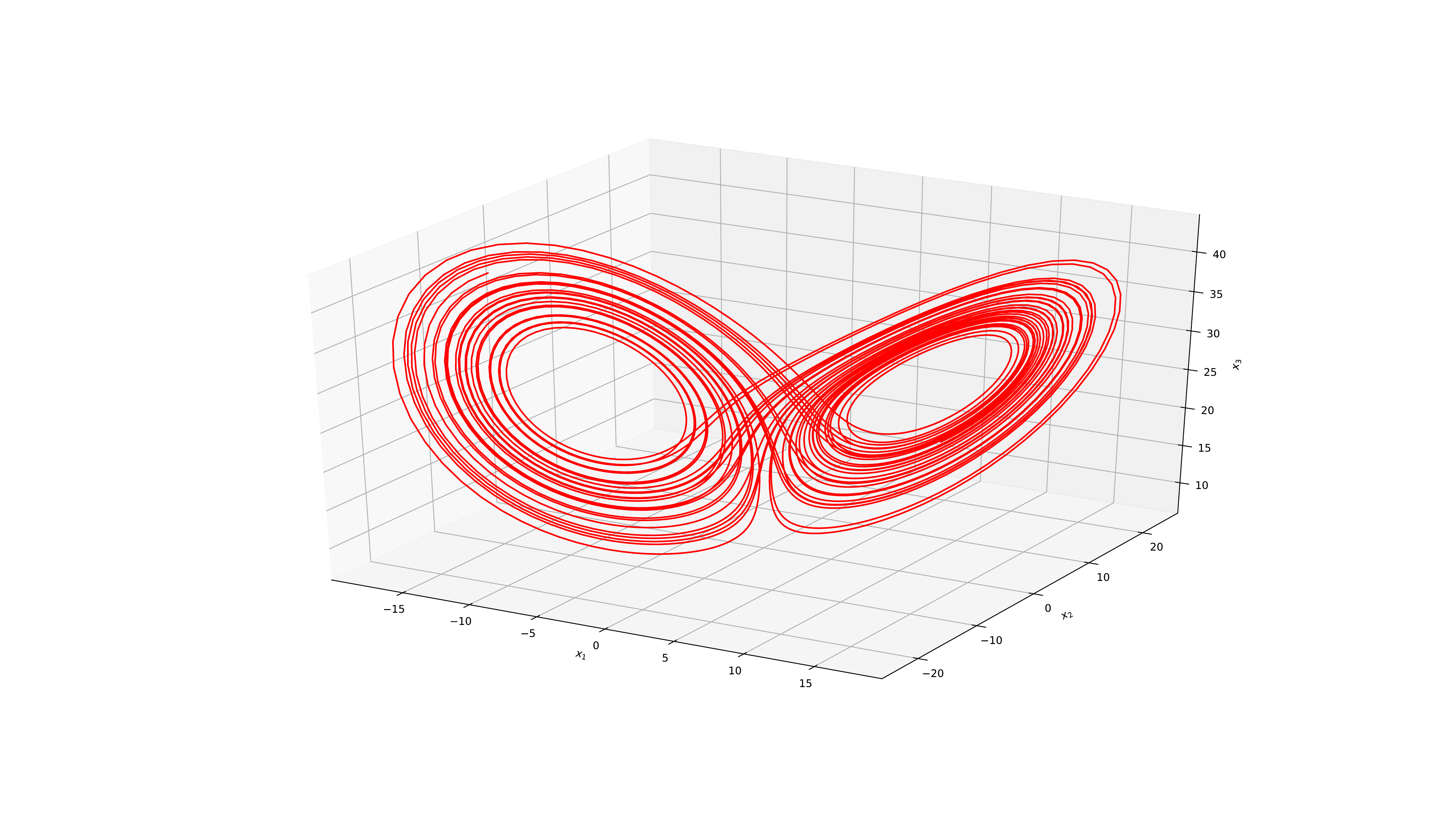}
		\includegraphics[width=\bwidth,clip, trim=100mm 40mm 80mm 40mm]{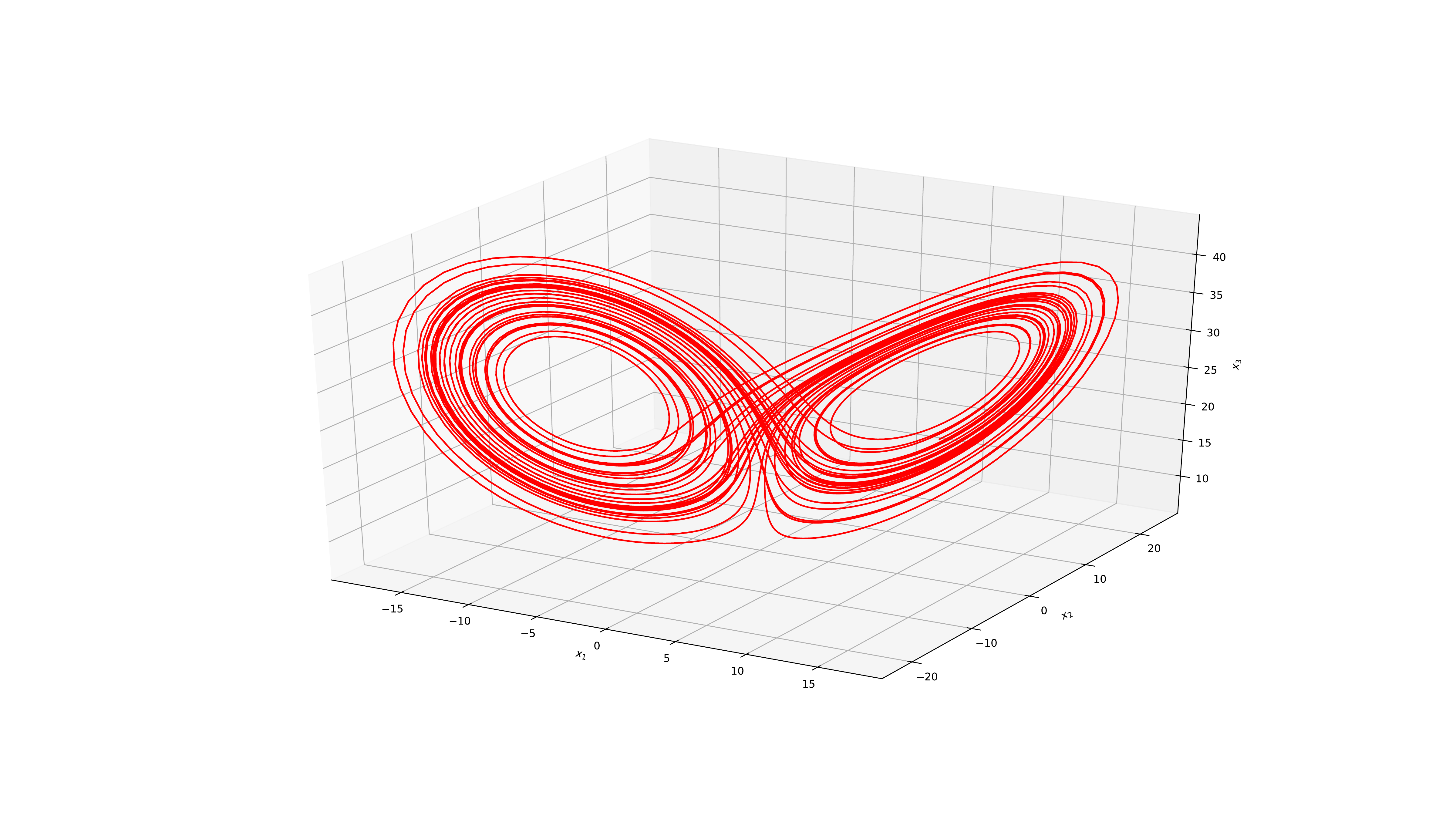}
		\includegraphics[width=\bwidth,clip, trim=100mm 40mm 80mm 40mm]{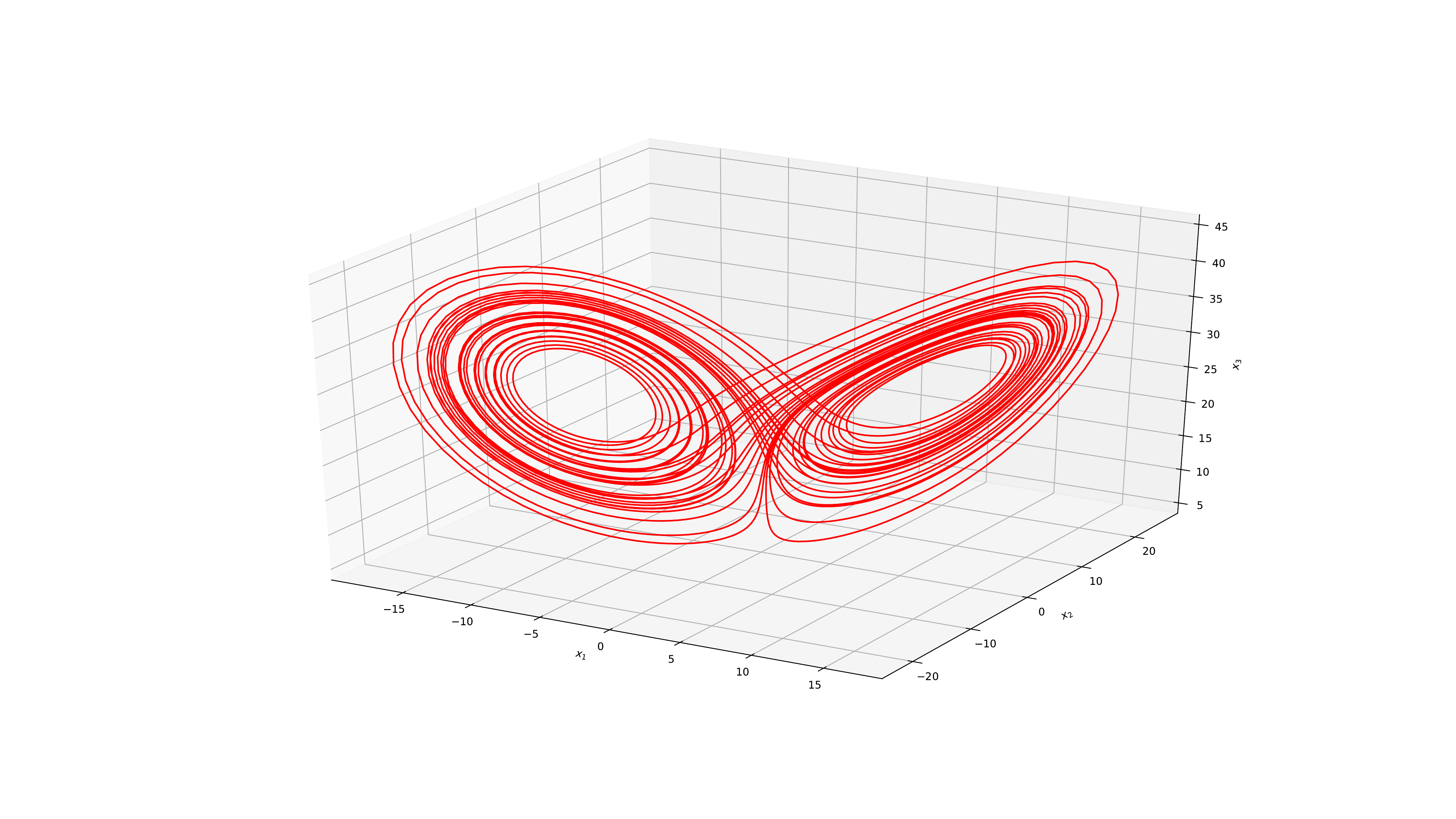}
	\end{subfigure}%
	
	\begin{subfigure}[b]{0.04\linewidth}
	    \rotatebox[origin=t]{90}{\scriptsize VODEN}\vspace{0.4\linewidth}
	\end{subfigure}%
	\begin{subfigure}[t]{0.96\linewidth}
		\centering
		\includegraphics[width=\bwidth,clip, trim=100mm 40mm 80mm 40mm]{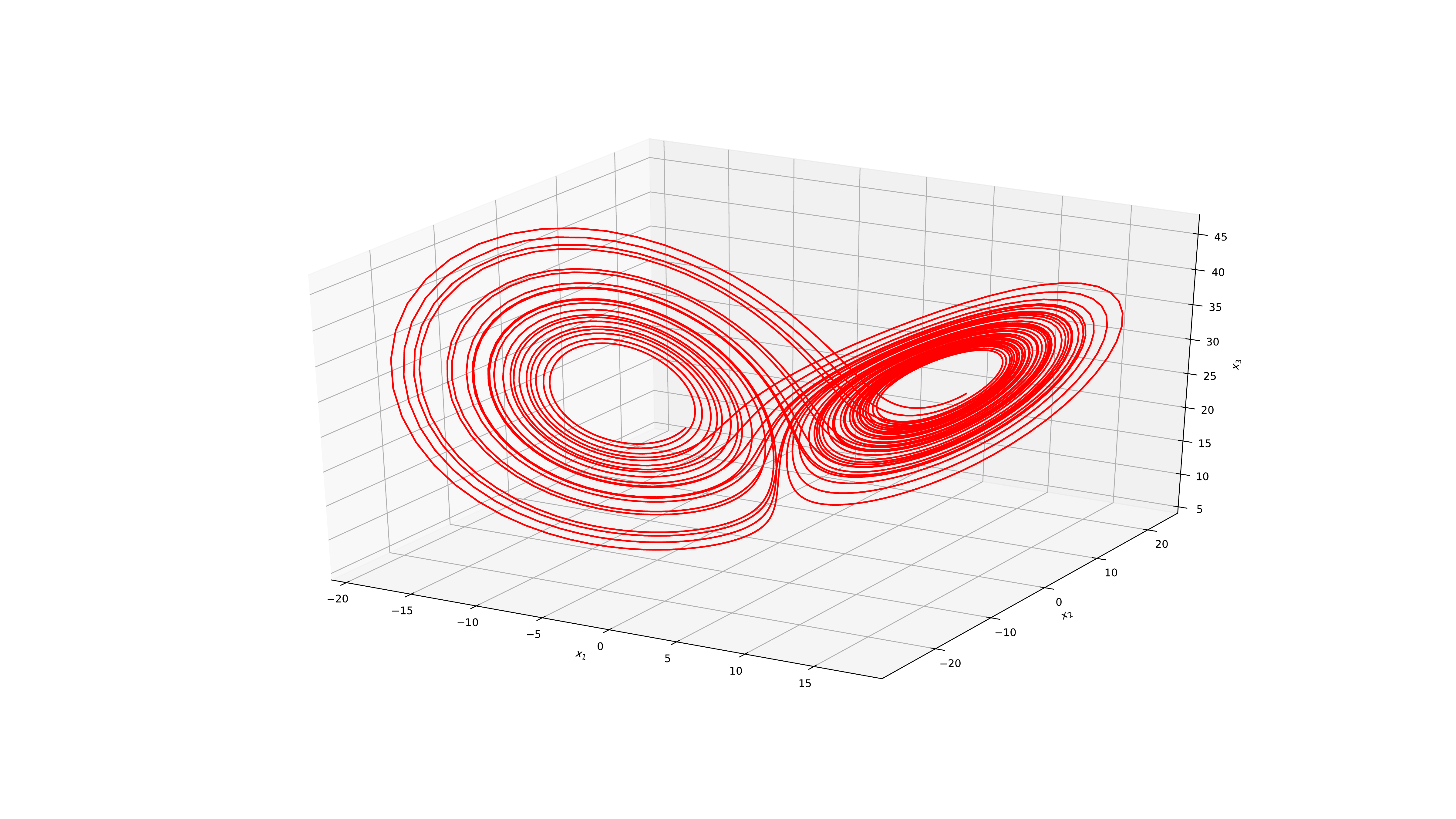}
		\includegraphics[width=\bwidth,clip, trim=100mm 40mm 80mm 40mm]{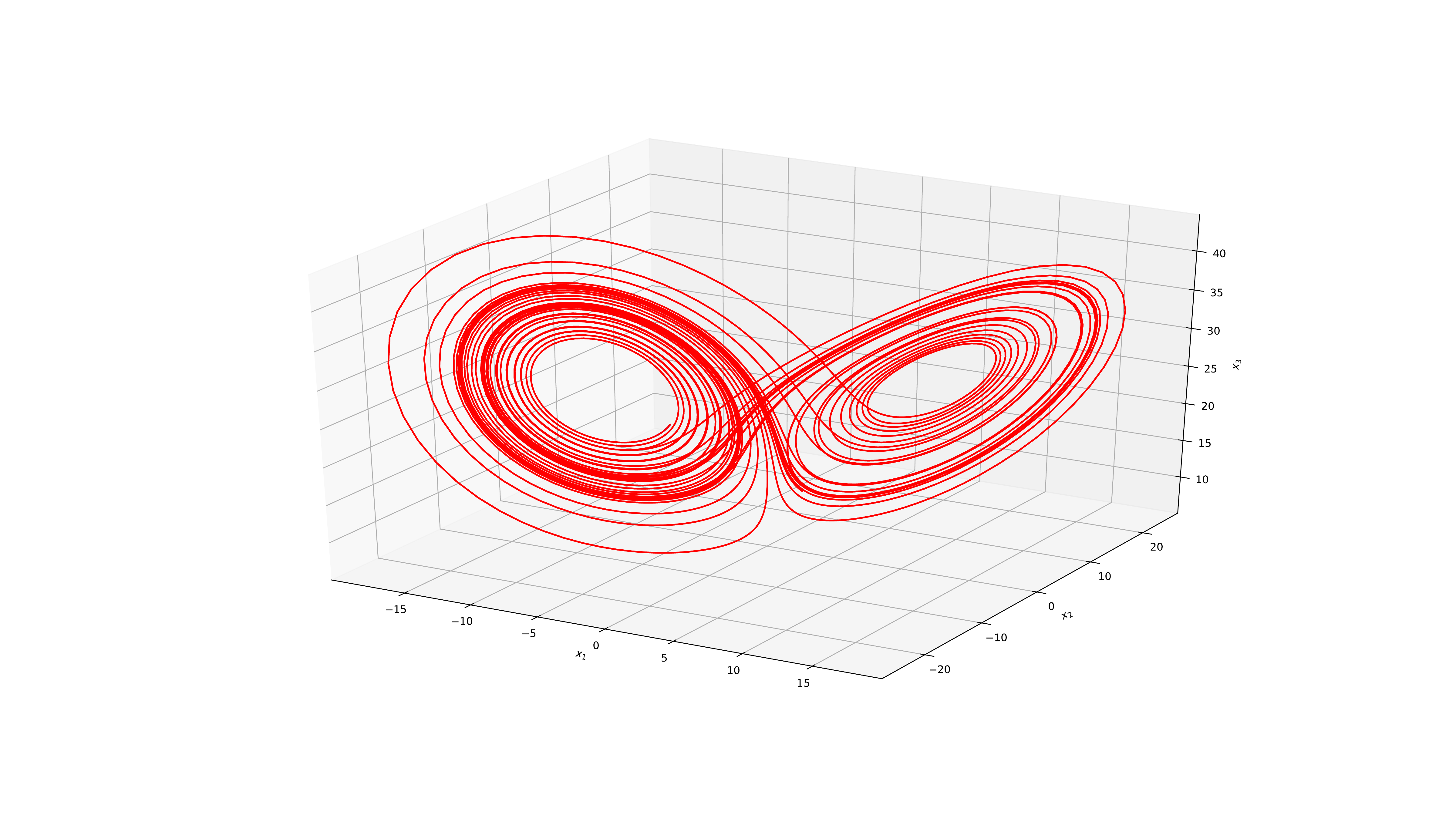}
		\includegraphics[width=\bwidth,clip, trim=100mm 40mm 80mm 40mm]{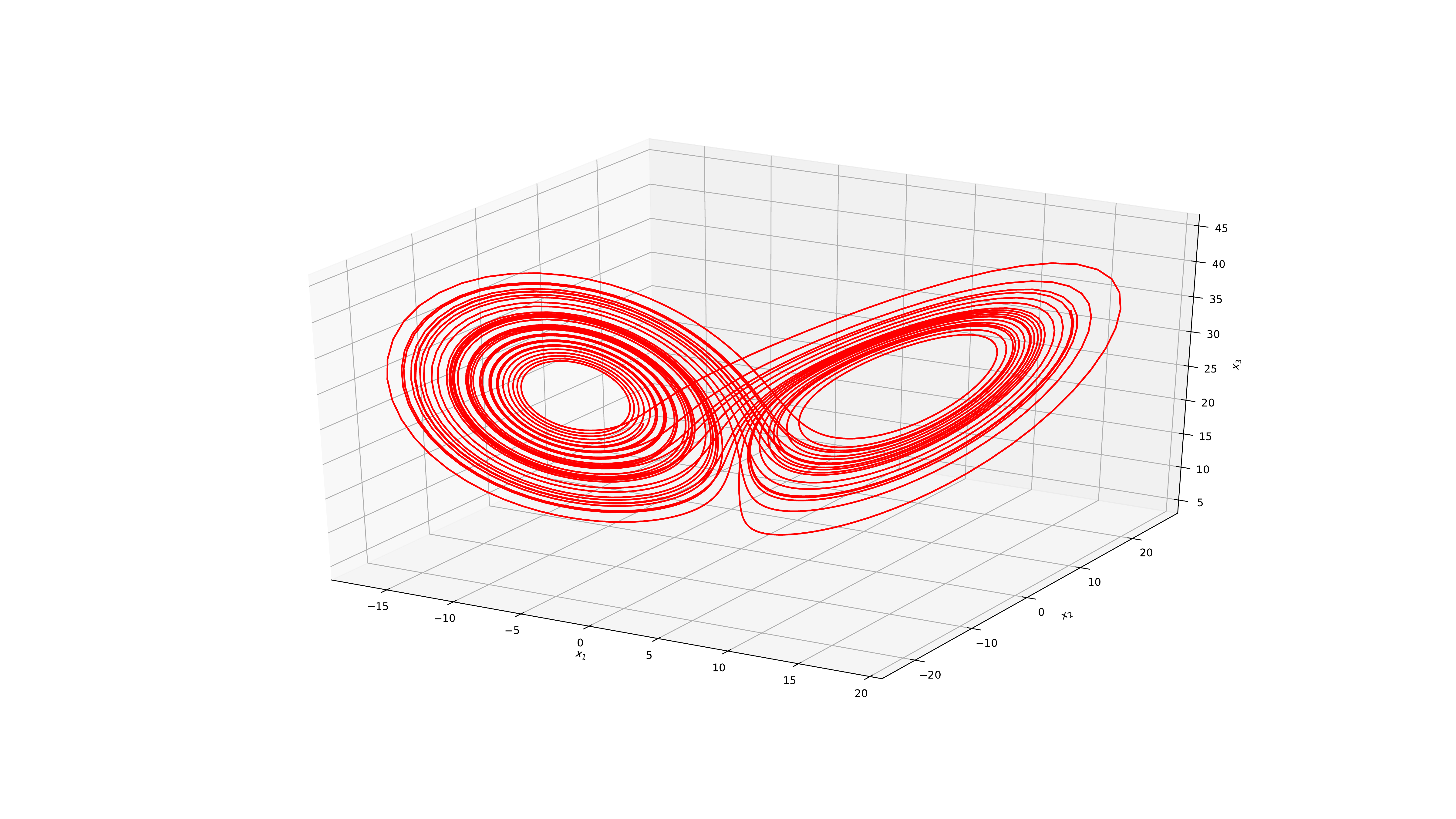}
		\includegraphics[width=\bwidth,clip, trim=100mm 40mm 80mm 40mm]{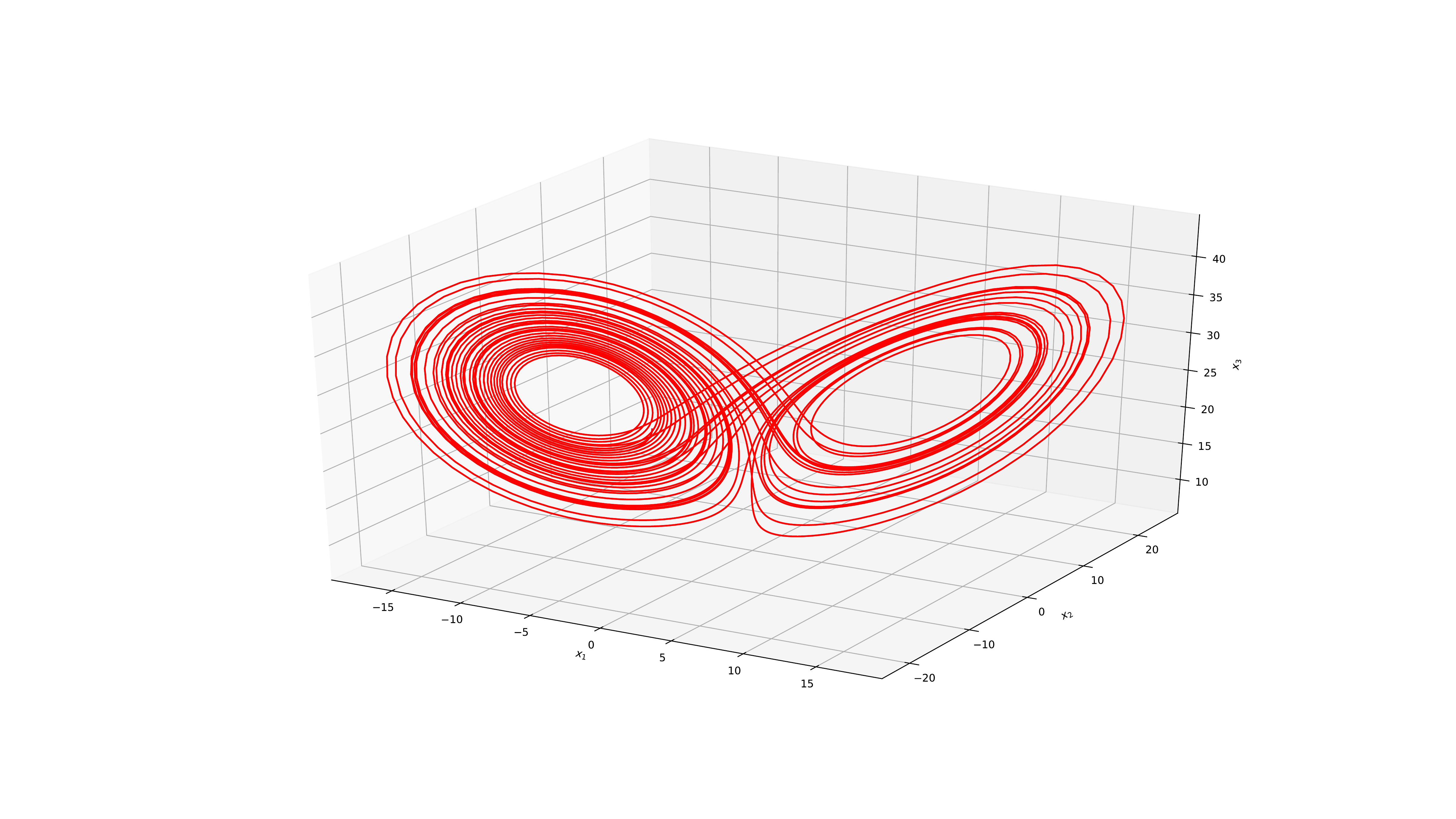}
		\includegraphics[width=\bwidth,clip, trim=100mm 40mm 80mm 40mm]{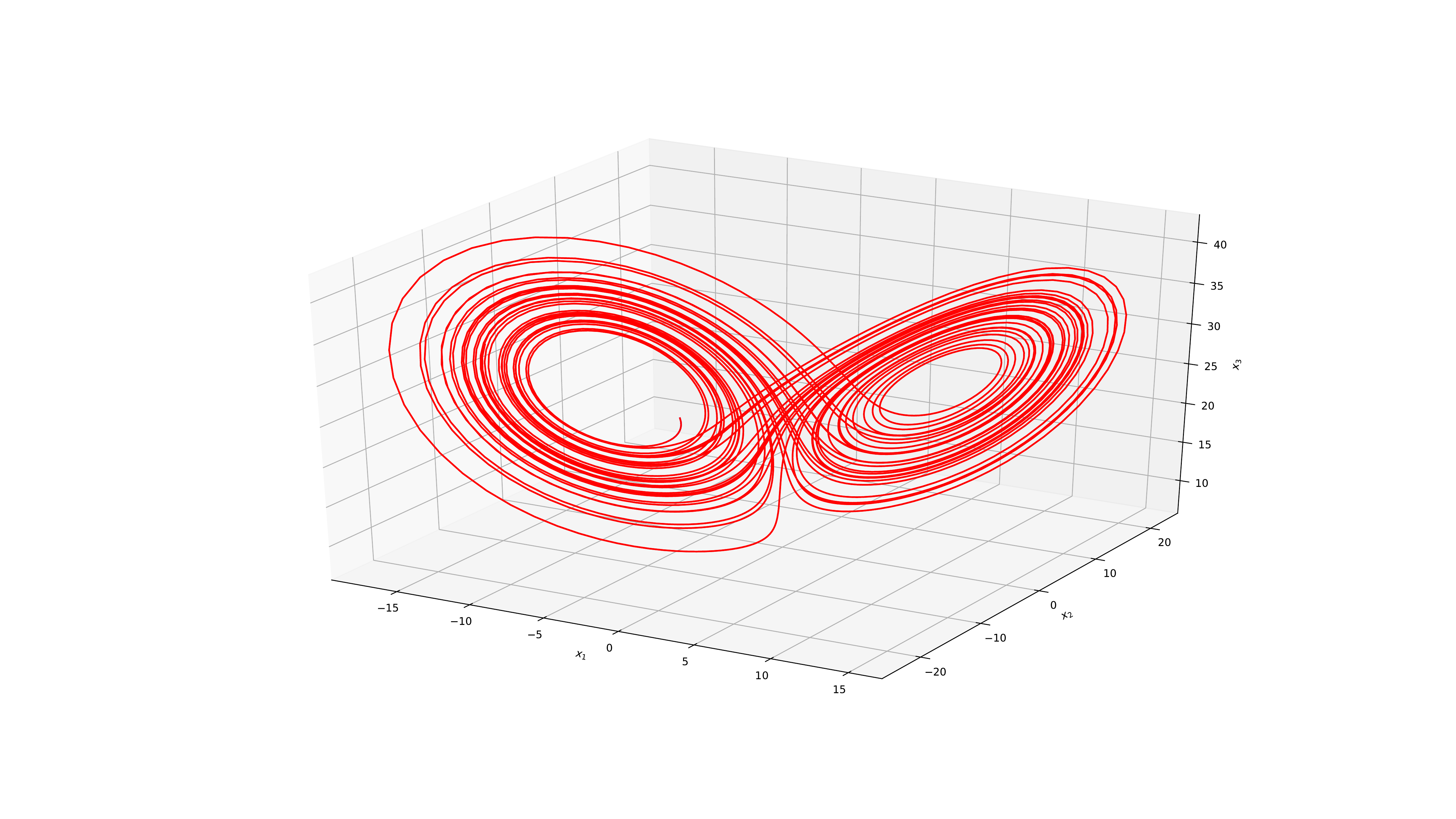}
		\includegraphics[width=\bwidth,clip, trim=100mm 40mm 80mm 40mm]{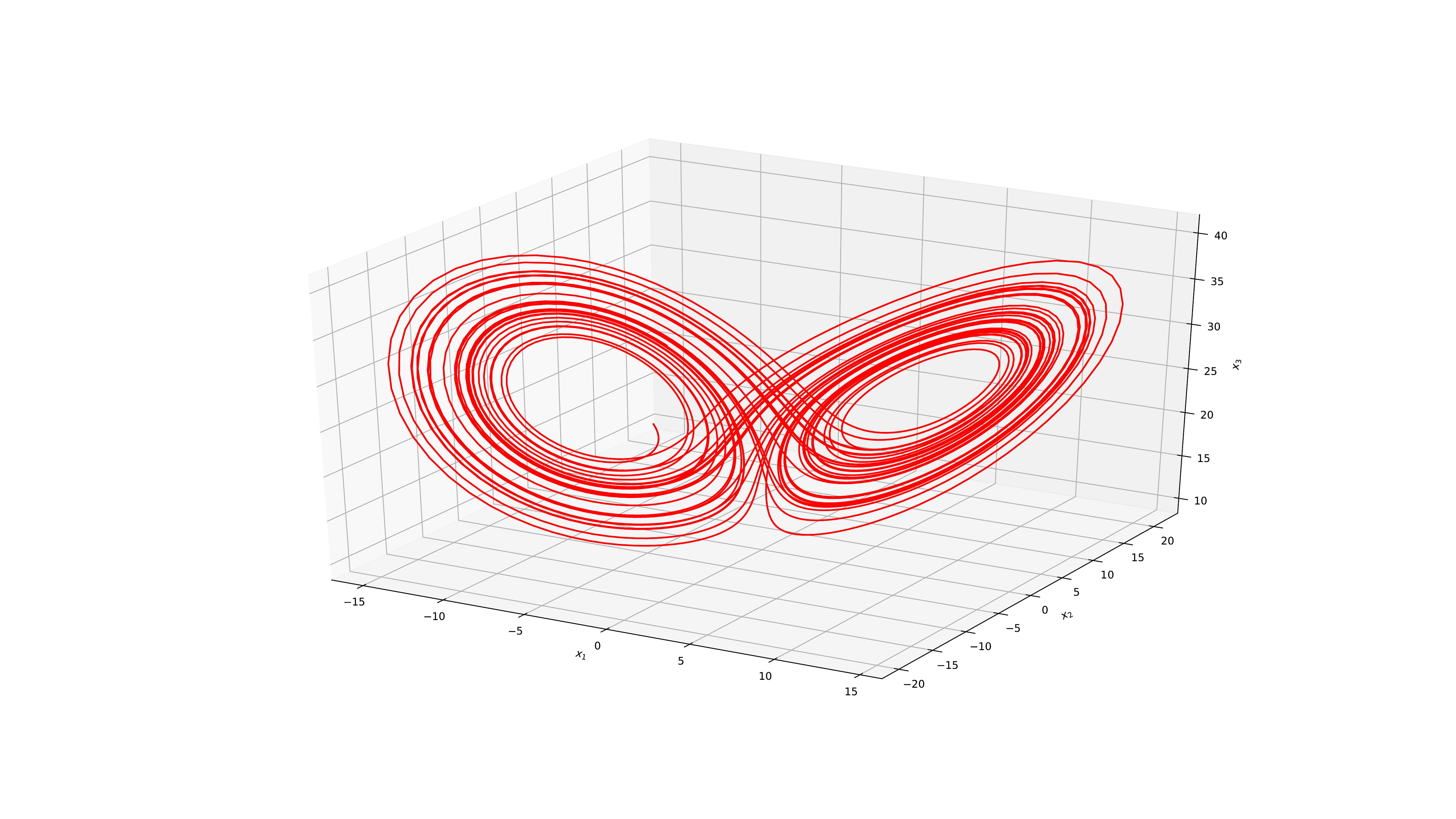}
	\end{subfigure}%
	
	\caption{Attactors generated by models trained on noisy data.}
    \label{fig:attactors_lorenz63_noisy}
\end{figure}

\subsection{Partial observation}

As mentioned above, one importance advantage of EM-like procedures for learning dynamical model is the ability to deal with partial observations. In this and the next sections, we show that the models trained by the proposed methodology on partial observations can obtain comparable performance with models in Section \ref{sec:lorenz63_noisy}.  

The first case study is when the observation frequency is small. We downsampled the data used in Section \ref{sec:lorenz63_noisy} 8 times (partial and regular observation). That means
\[
 \Phi_t = 
  \begin{cases} 
   [1, 1, 1]^T  & \text{if } t\%8 == 0 \\
   [0, 0, 0]^T  & \text{elsewhere}
  \end{cases}
\]
In the second case study we also reduced the number of observations 8 times, moreover, the observation is also irregular, both in time and in space. 

\begin{figure}
\centering
    \begin{subfigure}{0.49\linewidth}
        \centering
        \includegraphics[width=1.0\textwidth,clip, trim=40mm 40mm 30mm 40mm]{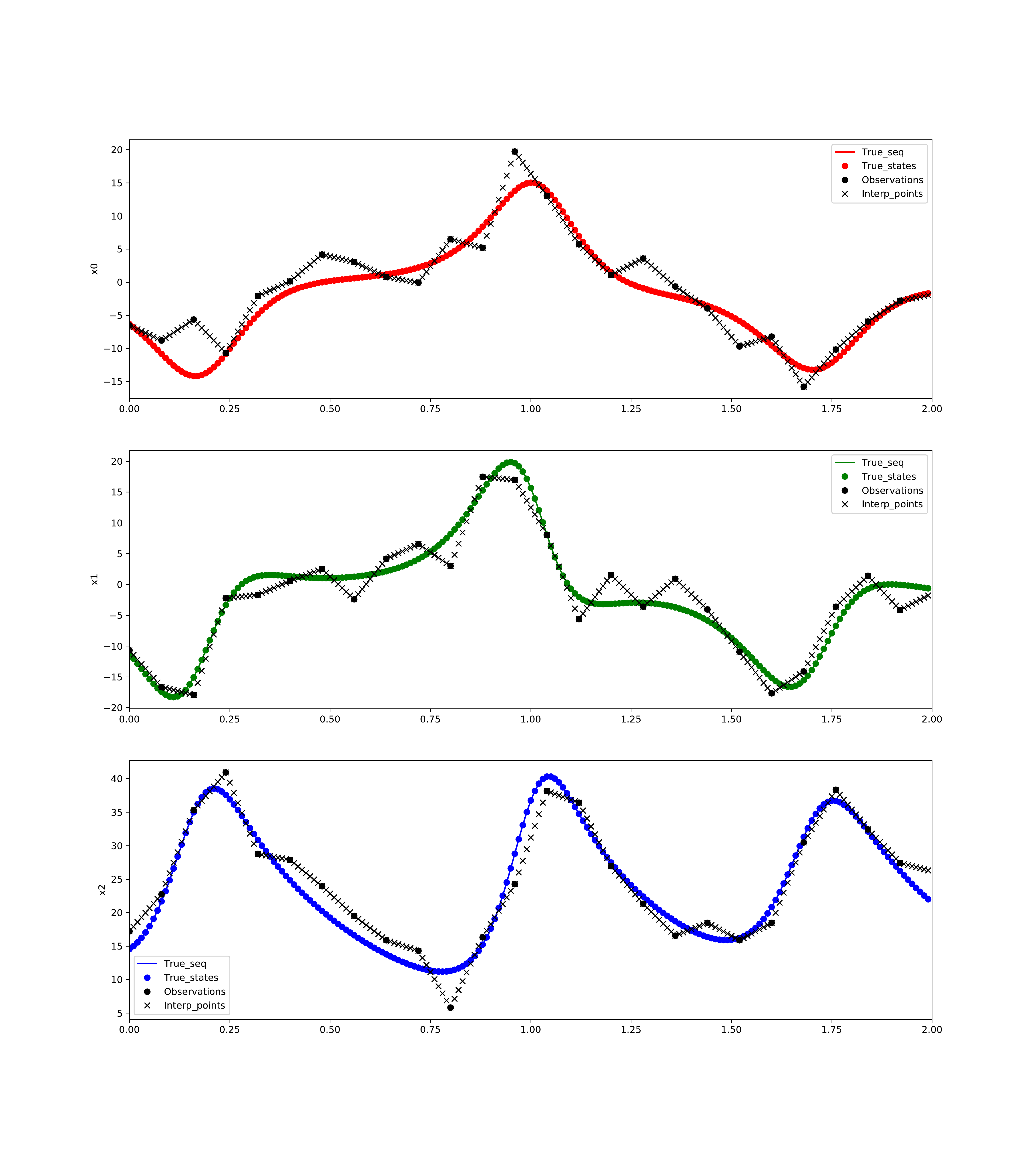}
        \caption{Scenario 1: Partial and regular.}
        \label{fig:scenario1}
    \end{subfigure}
    \begin{subfigure}{0.49\linewidth}
        \centering
        \includegraphics[width=1.0\textwidth,clip, trim=40mm 40mm 30mm 40mm]{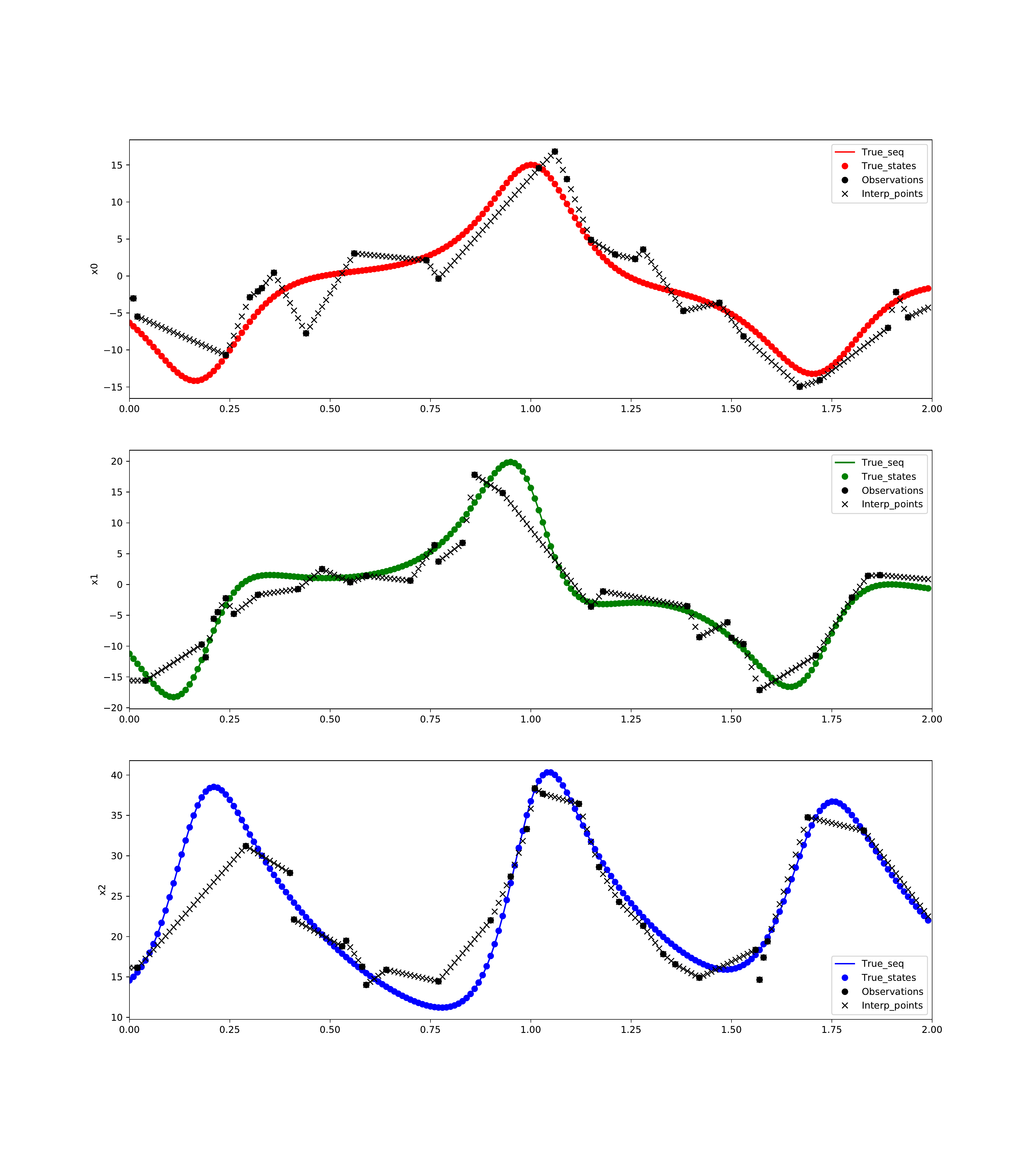}
        \caption{Scenario 2: Partial and irregular.}
        \label{fig:scenario2}
    \end{subfigure}
    \caption{Noisy and partial observation, noise level = 8.0. For VODEN, we use linear interpolation to interpolate the observation.}
    \label{fig:lorenz63_partial}
\end{figure}

While EnKS-EM can naturally deal with partial observations, VODEN needs an input at each time step. We use linear interpolation to interpolate where the observations is missing, as shown in Fig. \ref{fig:lorenz63_partial}.

\begin{figure}
    \centering
    \includegraphics[width=1.0\textwidth,clip, trim=40mm 20mm 30mm 20mm]{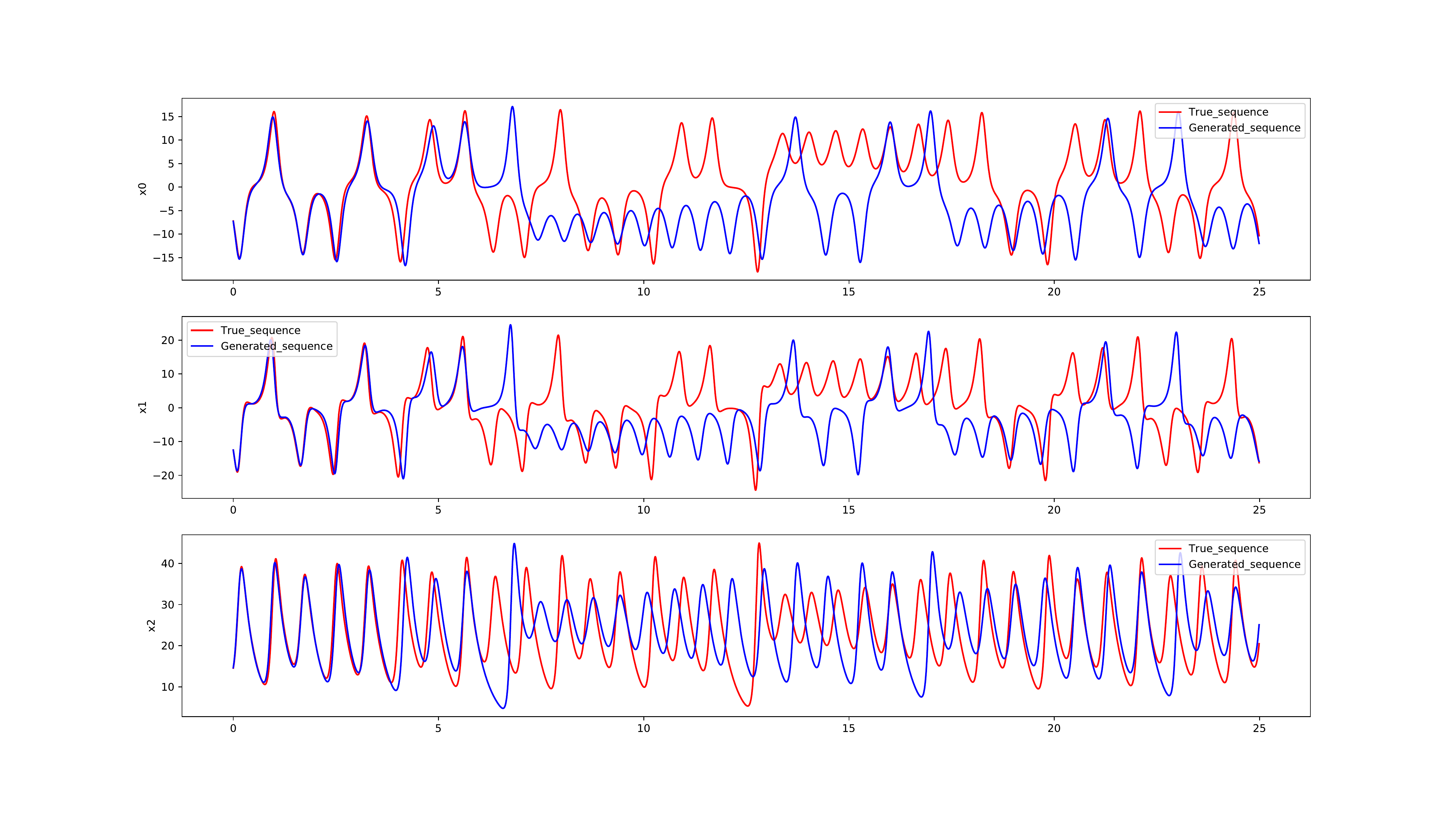}
    \caption{Sequences forecast by a VODEN learned on partial and irregular Lorenz-63 data (scenario 2), $\sigma^2=0.5$.}
    \label{fig:seq_comparision_voden_lorenz63_s2_0.5}
\end{figure}

\begin{figure}
    \centering
	\begin{subfigure}[t]{0.04\linewidth}
		\hfill
	    \vspace{-2.5mm}
		\caption*{}
	\end{subfigure}%
	\begin{subfigure}[t]{0.96\linewidth}
		\hspace{4mm} $\sigma^2=0.5$ \hspace{\swidth} $\sigma^2=2.0$ \hspace{\swidth} $\sigma^2=4.0$ \hspace{\swidth} $\sigma^2=8.0$ \hspace{\swidth} $\sigma^2=16.0$ \hspace{\swidth} $\sigma^2=32.0$ \hfill
	\end{subfigure}%
	
	\begin{subfigure}[b]{0.04\linewidth}
        \rotatebox[origin=t]{90}{\scriptsize EnKS-EM\_S1}\vspace{0.1\linewidth}
	\end{subfigure}%
	\begin{subfigure}[t]{0.96\linewidth}
		\centering
		\includegraphics[width=\bwidth,clip, trim=100mm 40mm 80mm 40mm]{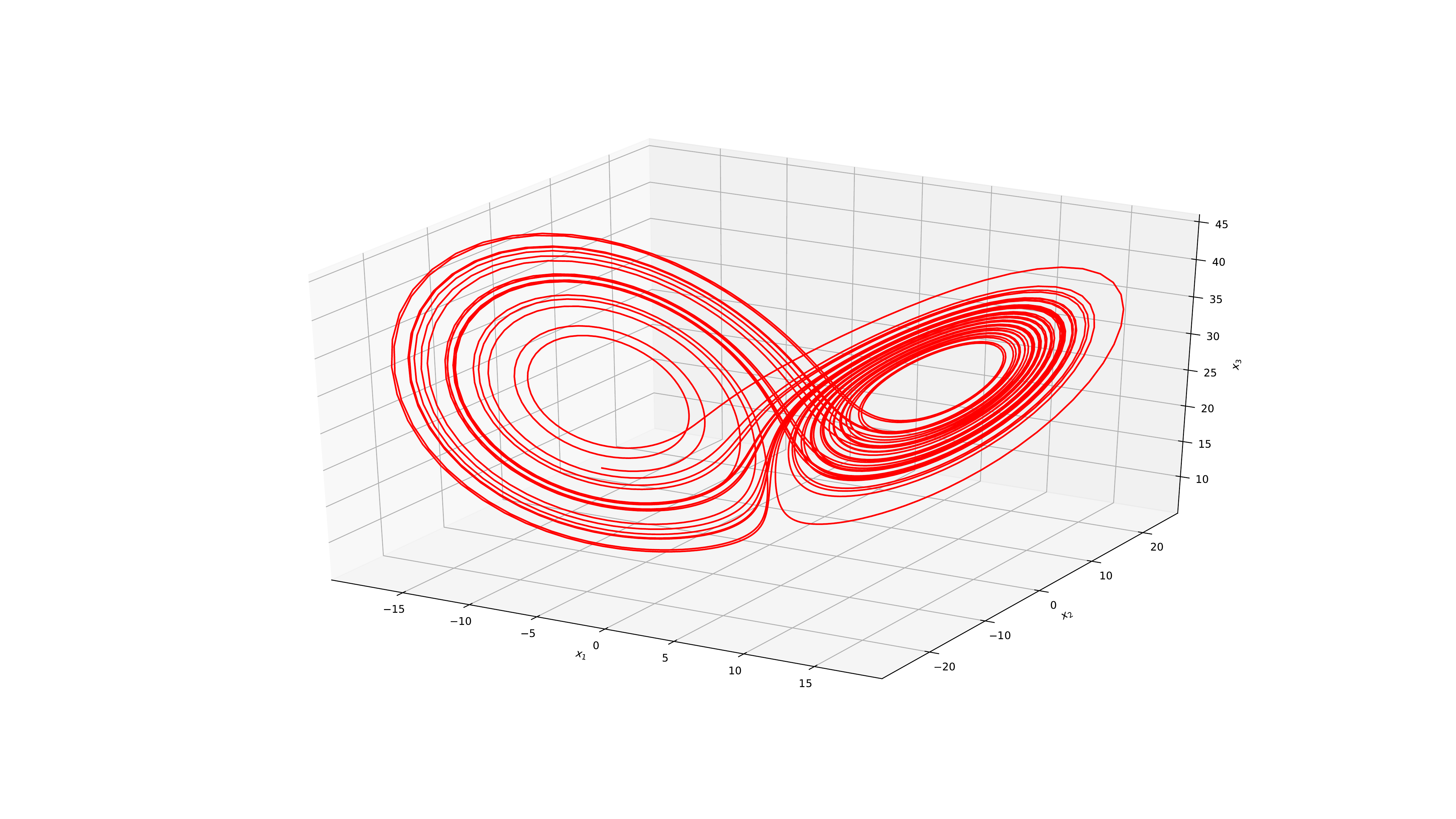}
		\includegraphics[width=\bwidth,clip, trim=100mm 40mm 80mm 40mm]{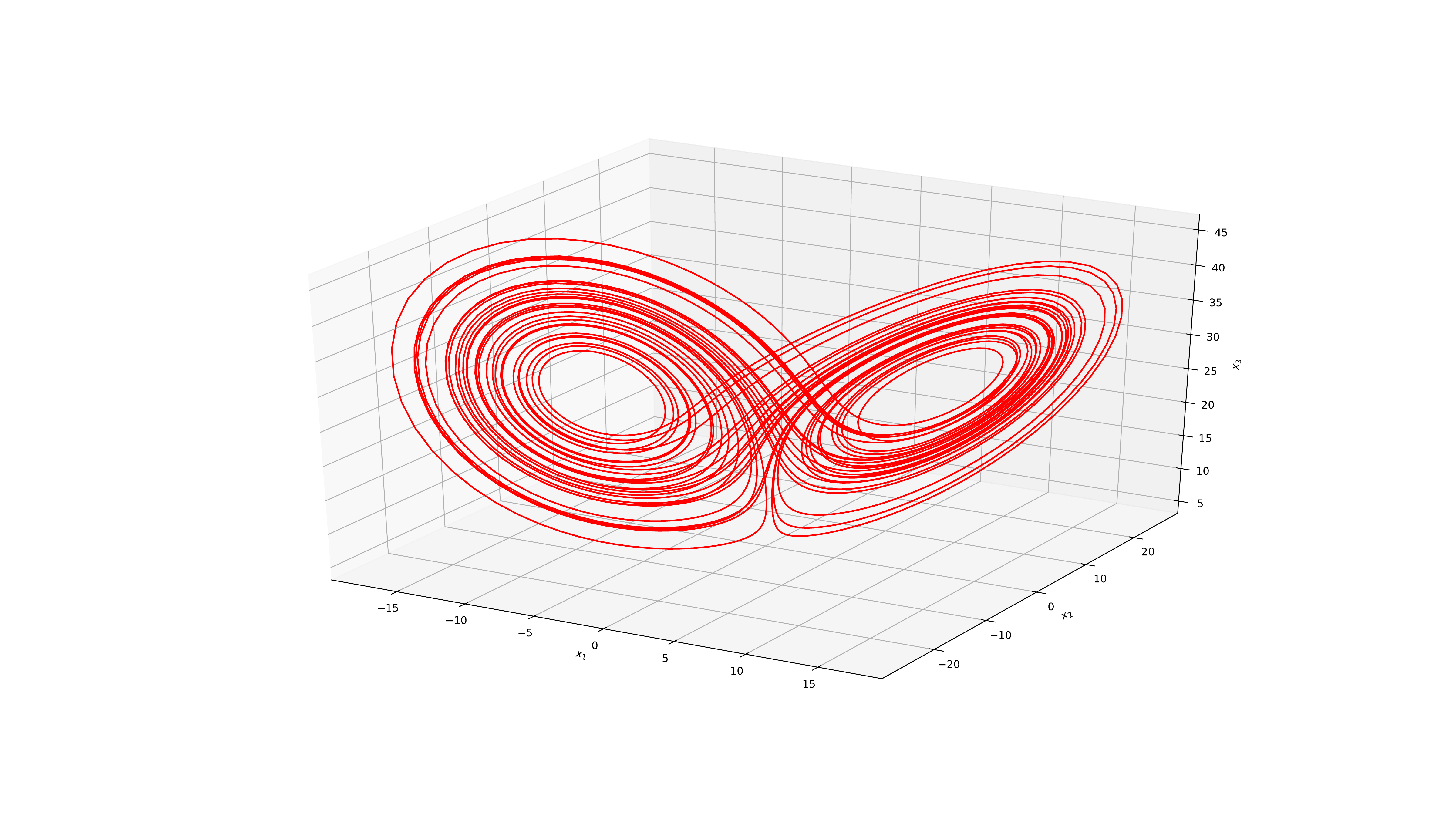}
		\includegraphics[width=\bwidth,clip, trim=100mm 40mm 80mm 40mm]{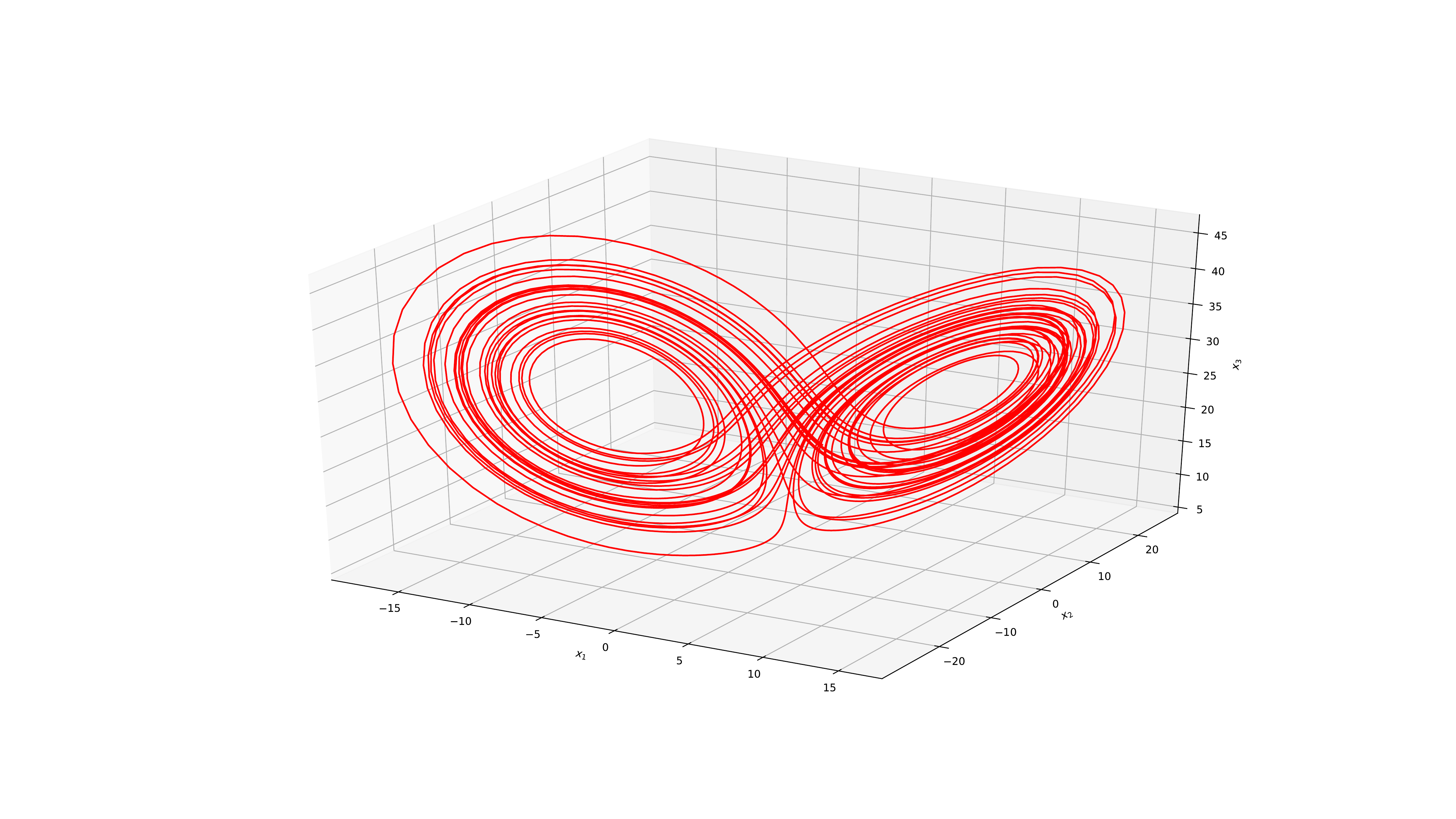}
		\includegraphics[width=\bwidth,clip, trim=100mm 40mm 80mm 40mm]{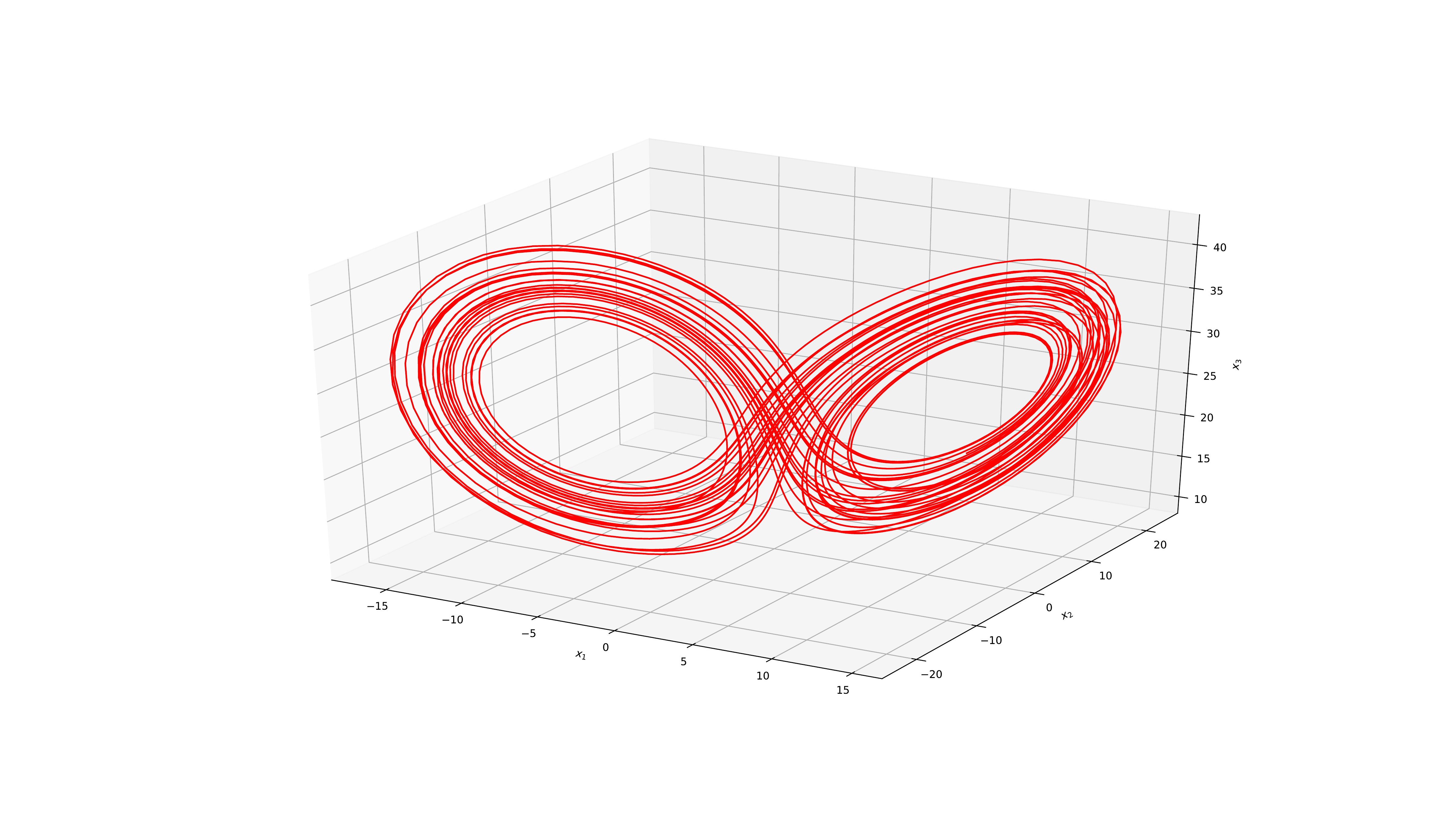}
		\includegraphics[width=\bwidth,clip, trim=100mm 40mm 80mm 40mm]{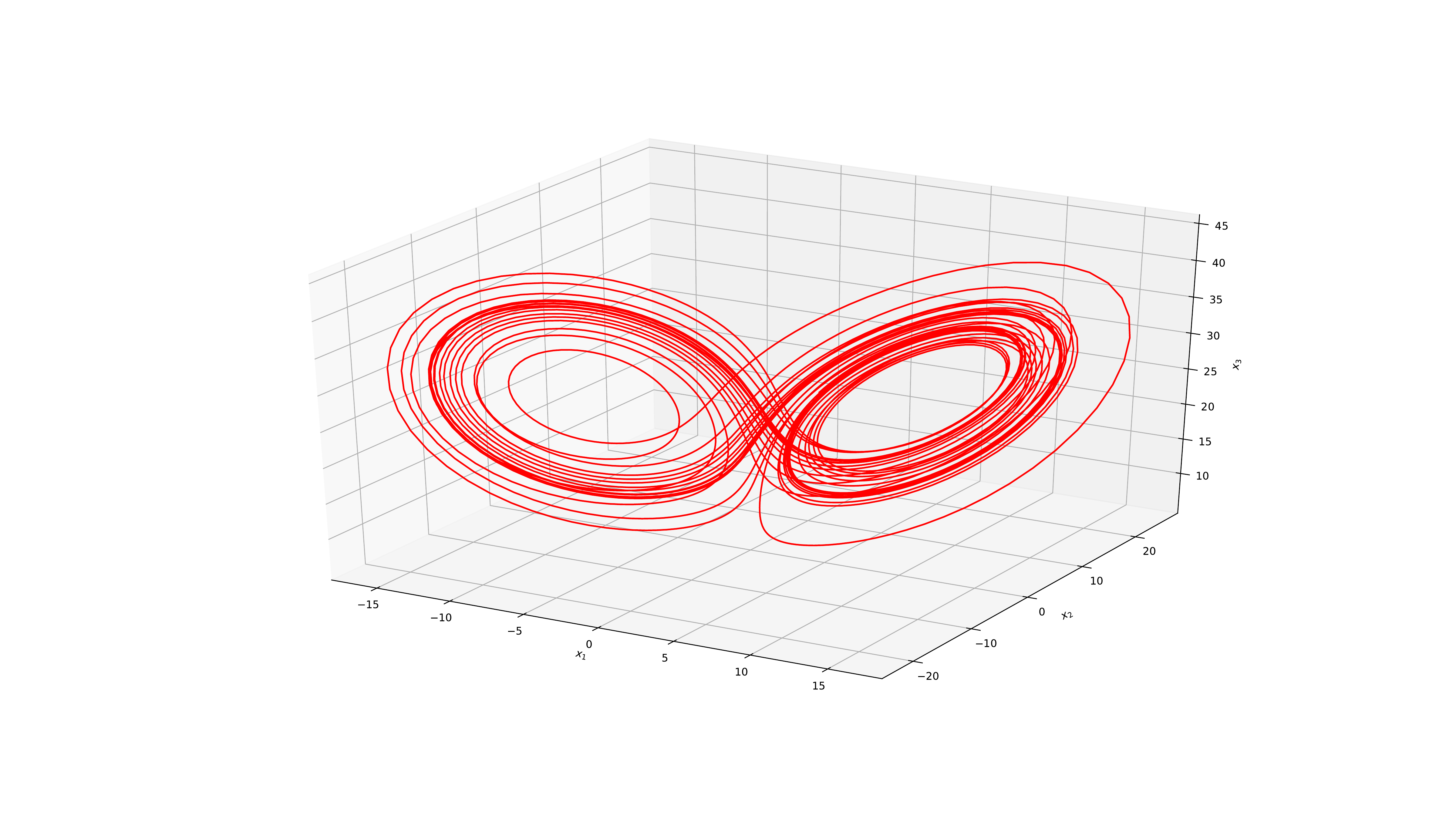}
		\includegraphics[width=\bwidth,clip, trim=100mm 40mm 80mm 40mm]{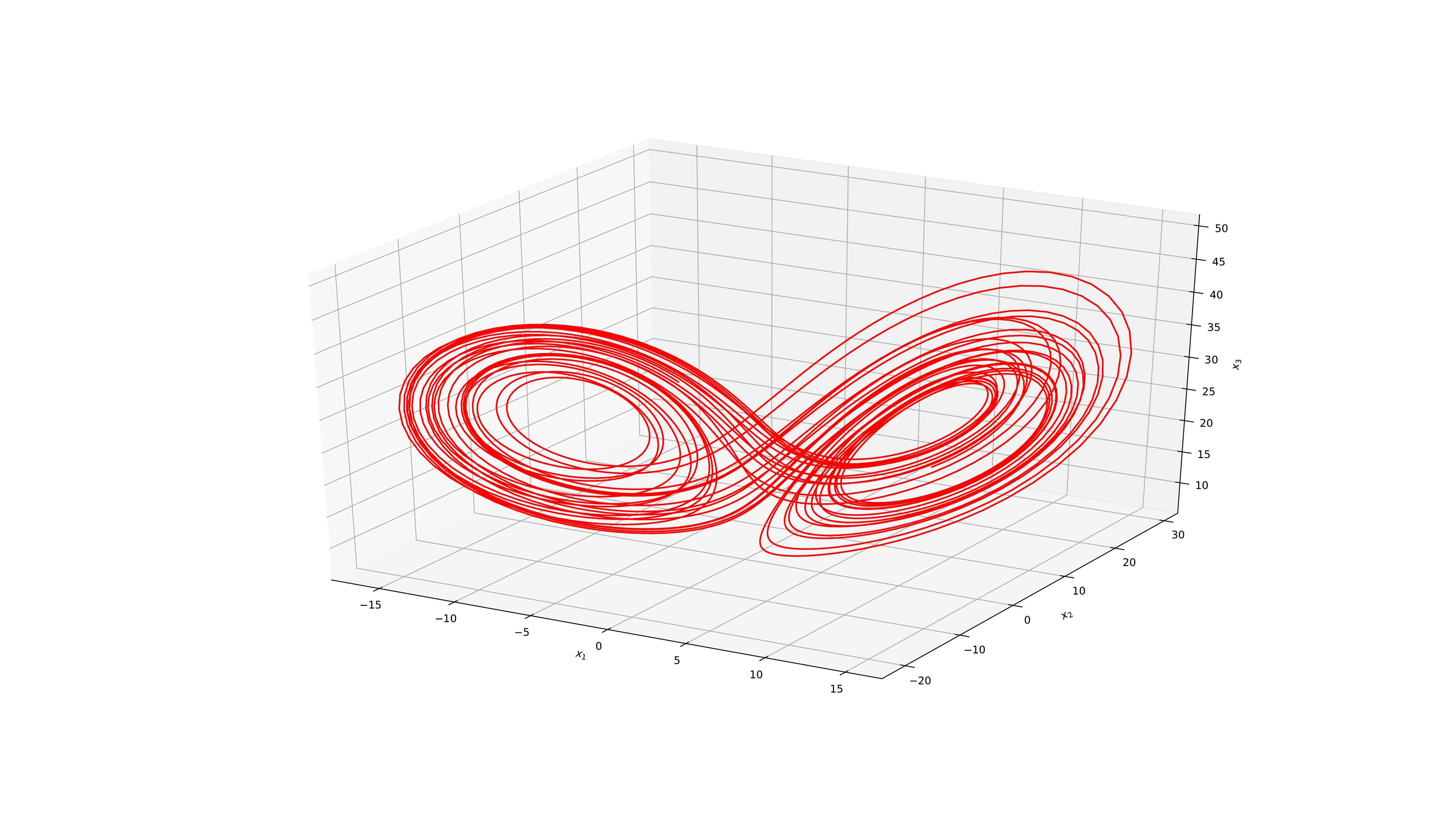}
	\end{subfigure}%
	
	\begin{subfigure}[b]{0.04\linewidth}
	    \rotatebox[origin=t]{90}{\scriptsize VODEN\_S1}\vspace{0.4\linewidth}
	\end{subfigure}%
	\begin{subfigure}[t]{0.96\linewidth}
		\centering
		\includegraphics[width=\bwidth,clip, trim=100mm 40mm 80mm 40mm]{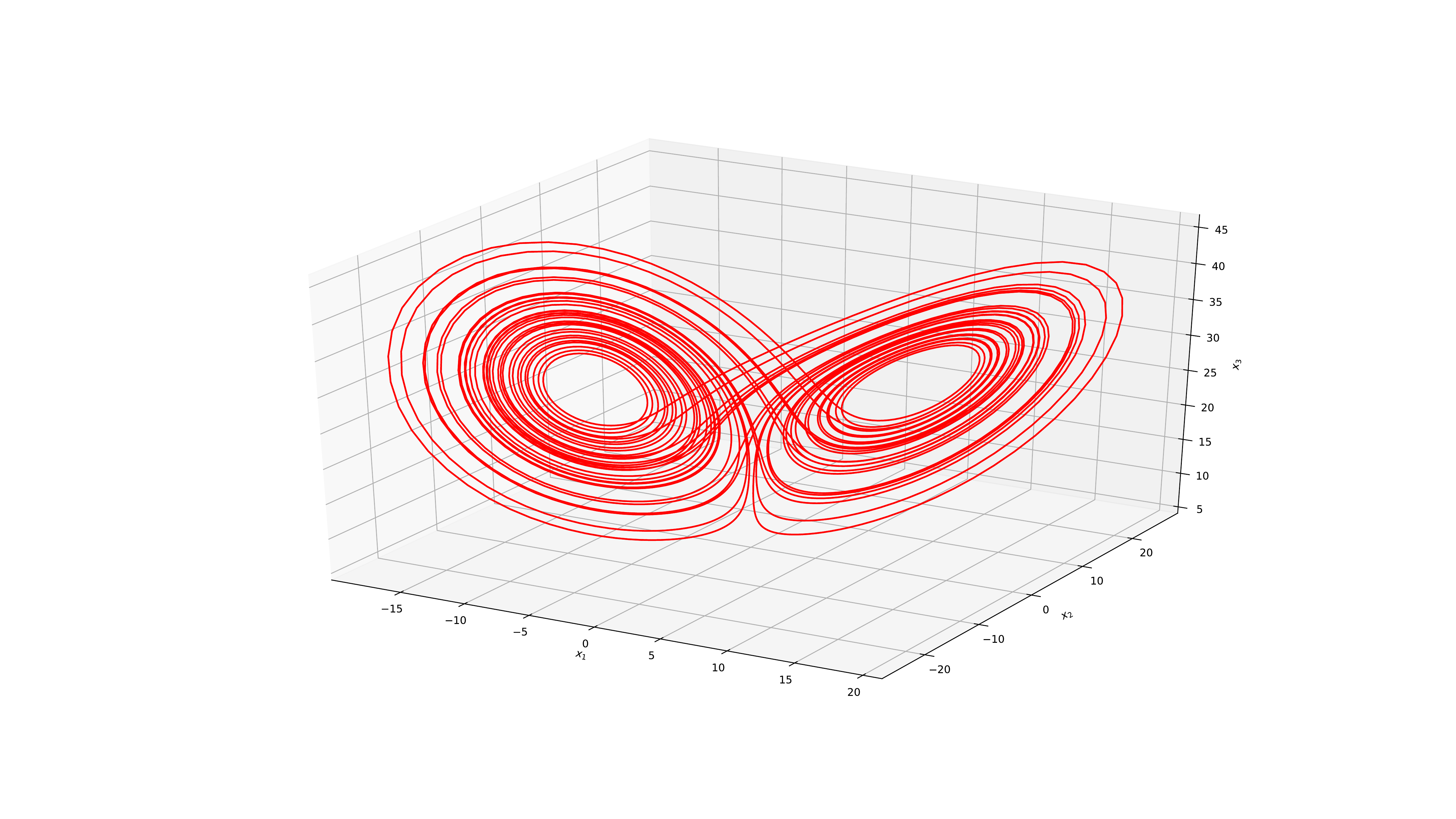}
		\includegraphics[width=\bwidth,clip, trim=100mm 40mm 80mm 40mm]{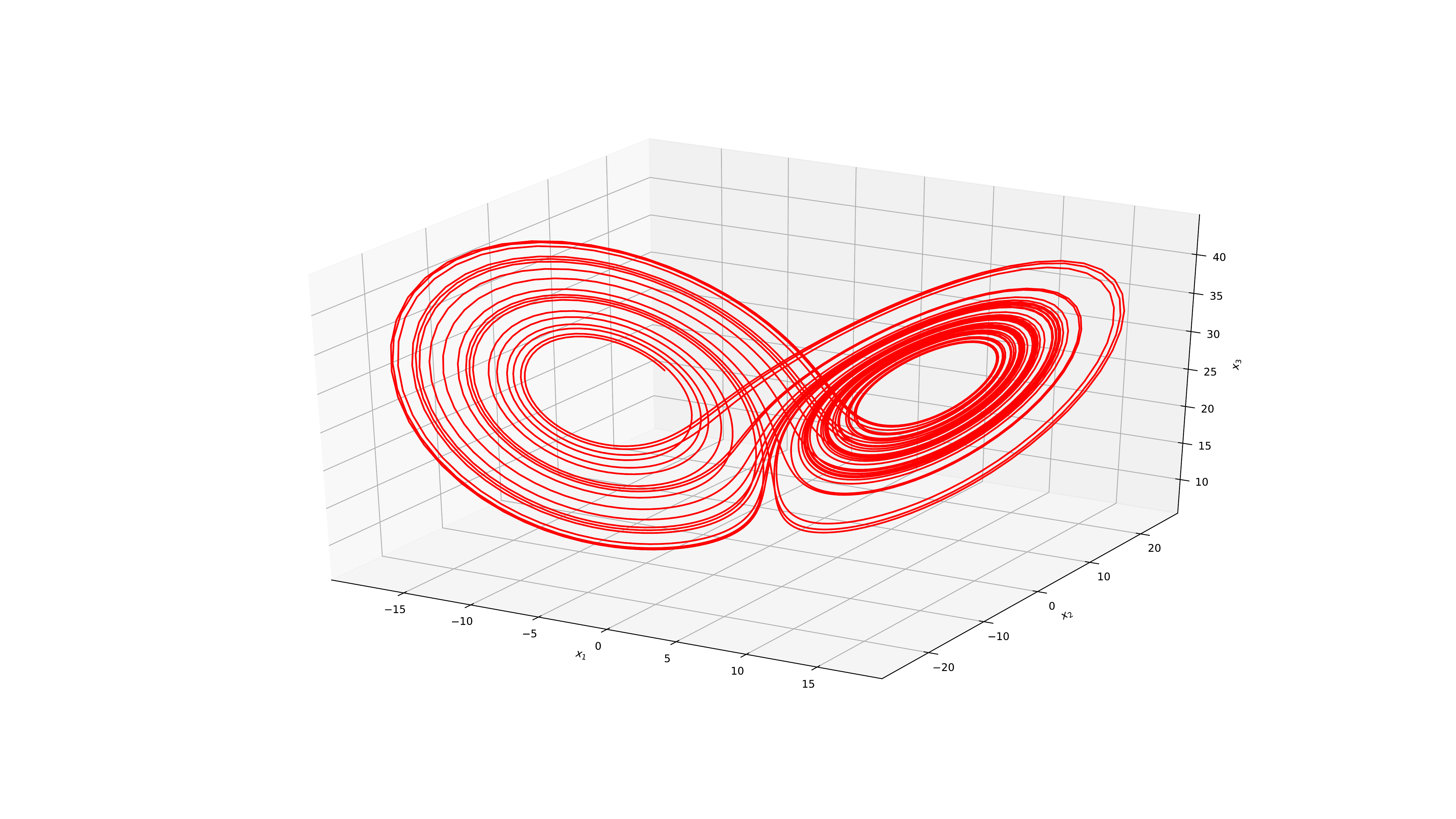}
		\includegraphics[width=\bwidth,clip, trim=100mm 40mm 80mm 40mm]{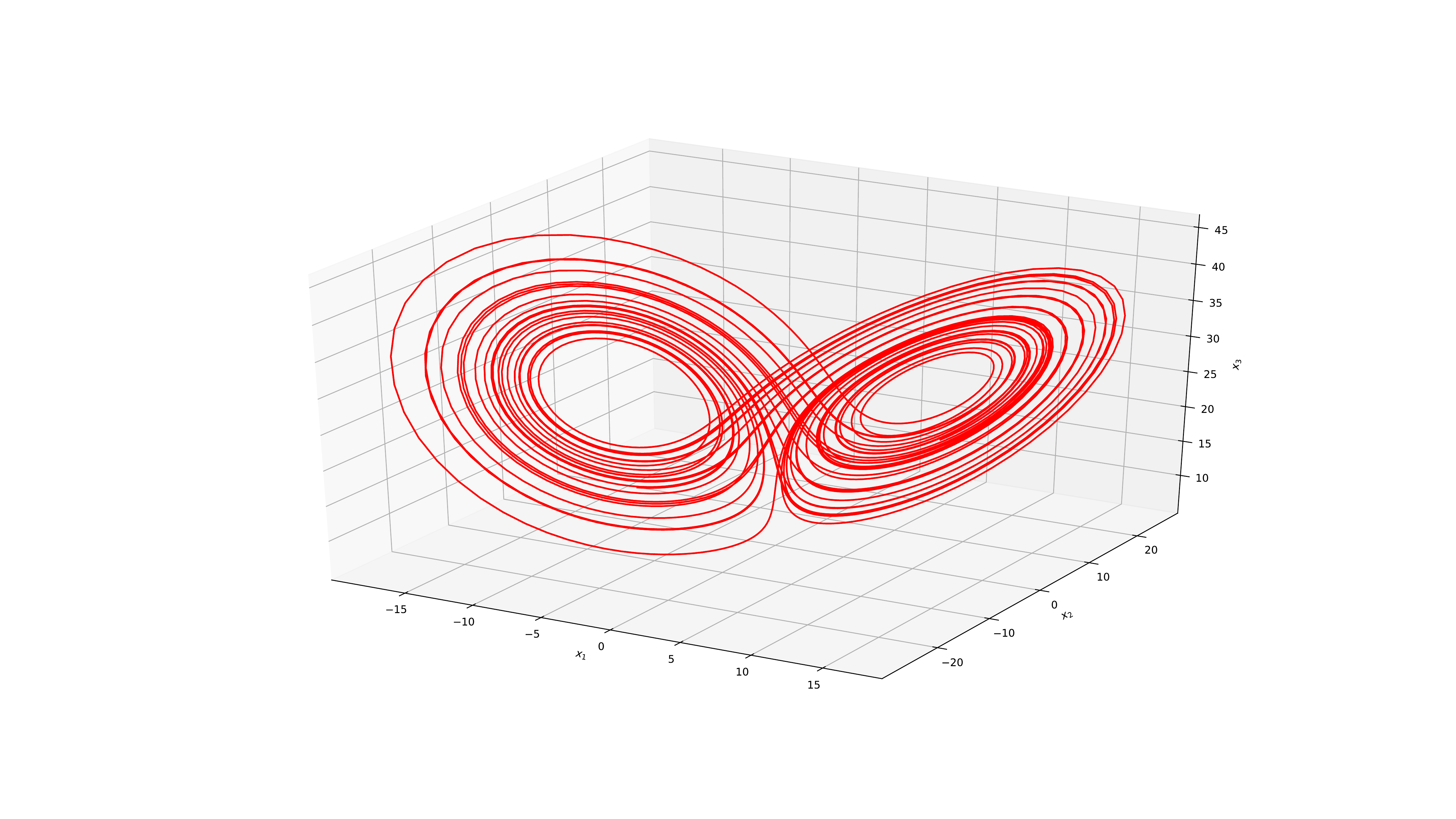}
		\includegraphics[width=\bwidth,clip, trim=100mm 40mm 80mm 40mm]{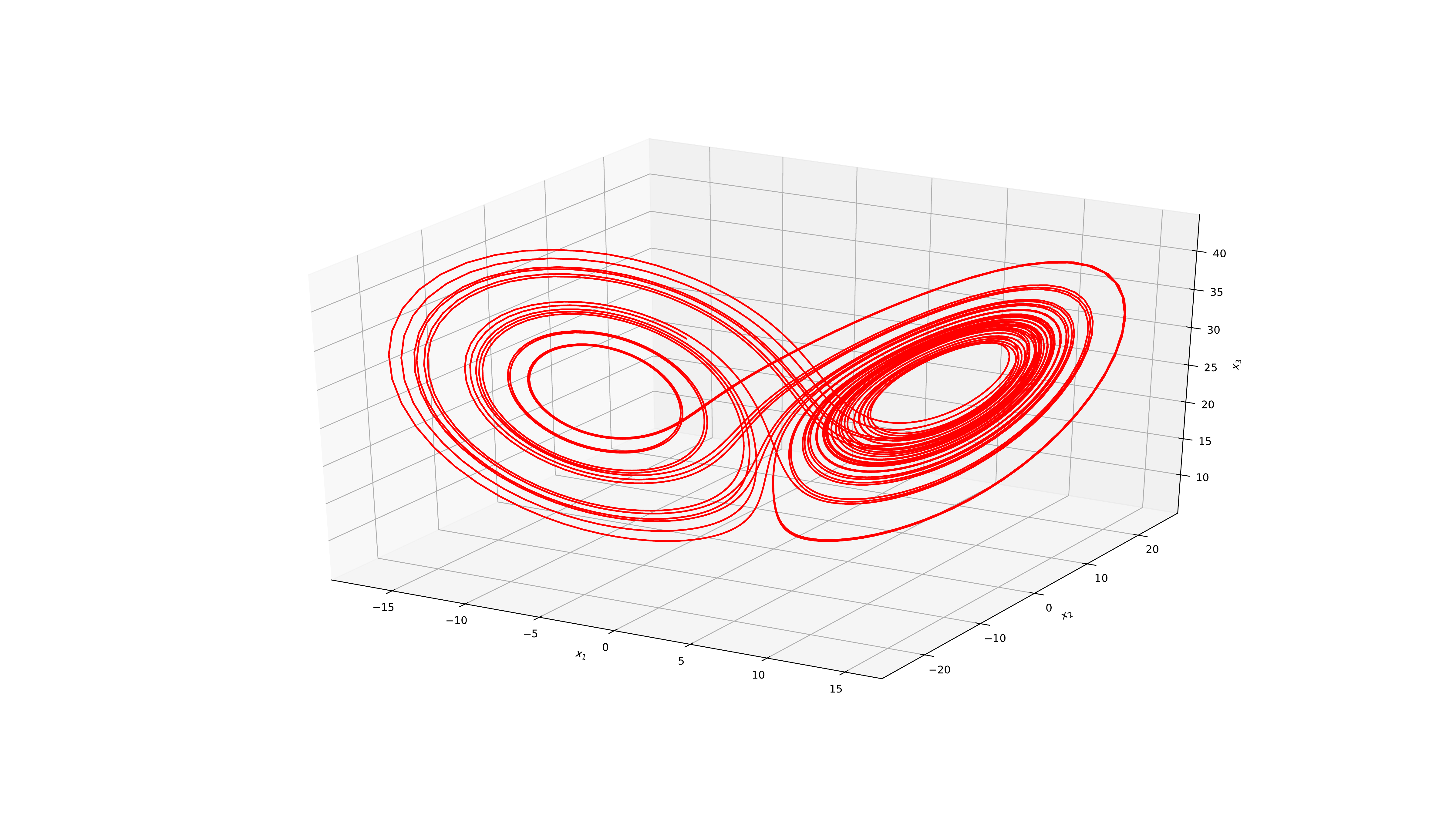}
		\includegraphics[width=\bwidth,clip, trim=100mm 40mm 80mm 40mm]{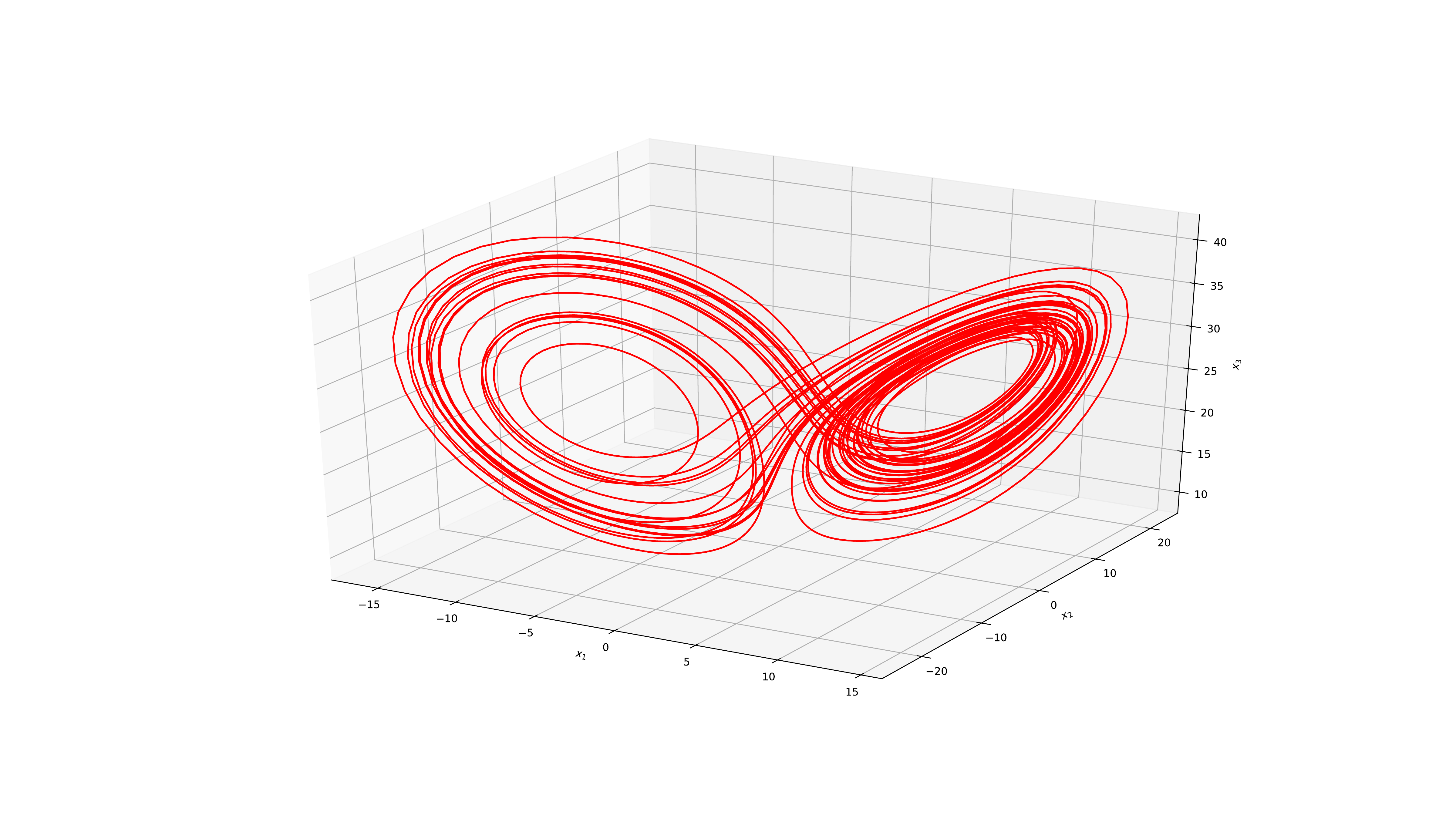}
		\includegraphics[width=\bwidth,clip, trim=100mm 40mm 80mm 40mm]{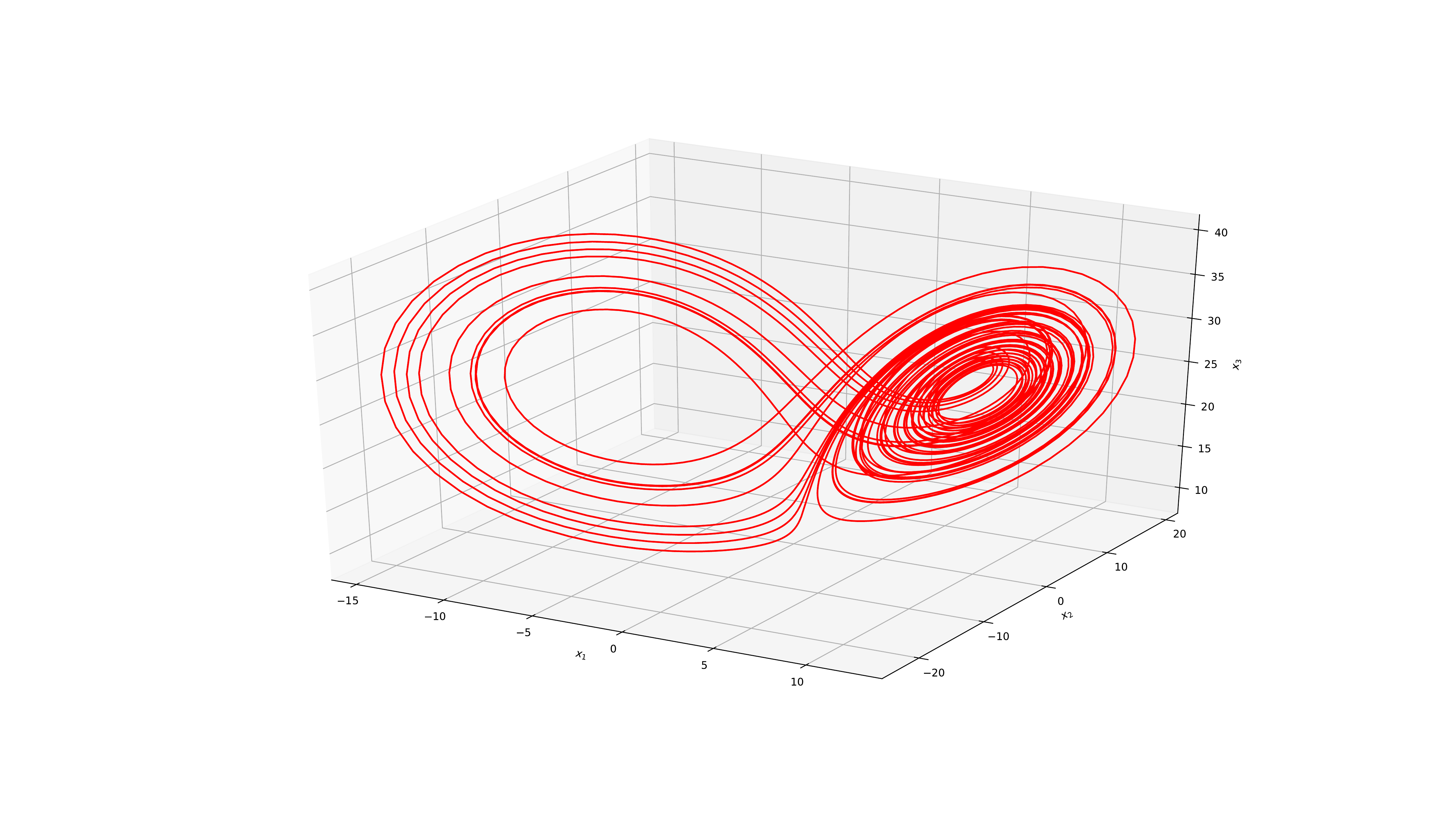}
	\end{subfigure}%
	
	\begin{subfigure}[b]{0.04\linewidth}
	    \rotatebox[origin=t]{90}{\scriptsize EnKS-EM\_S2}\vspace{0.2\linewidth}
	\end{subfigure}%
	\begin{subfigure}[t]{0.96\linewidth}
		\centering
		\includegraphics[width=\bwidth,clip, trim=100mm 40mm 80mm 40mm]{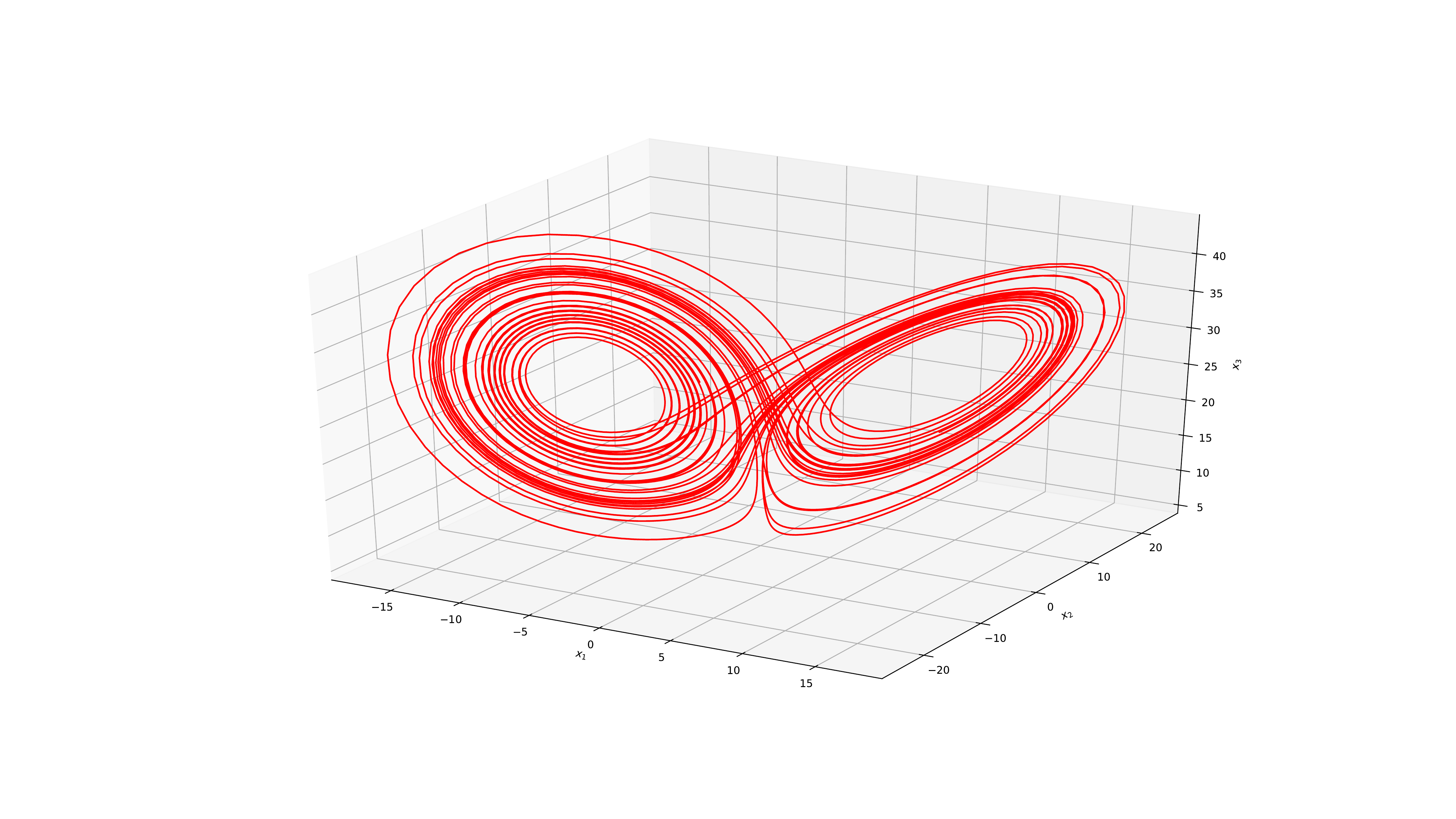}
		\includegraphics[width=\bwidth,clip, trim=100mm 40mm 80mm 40mm]{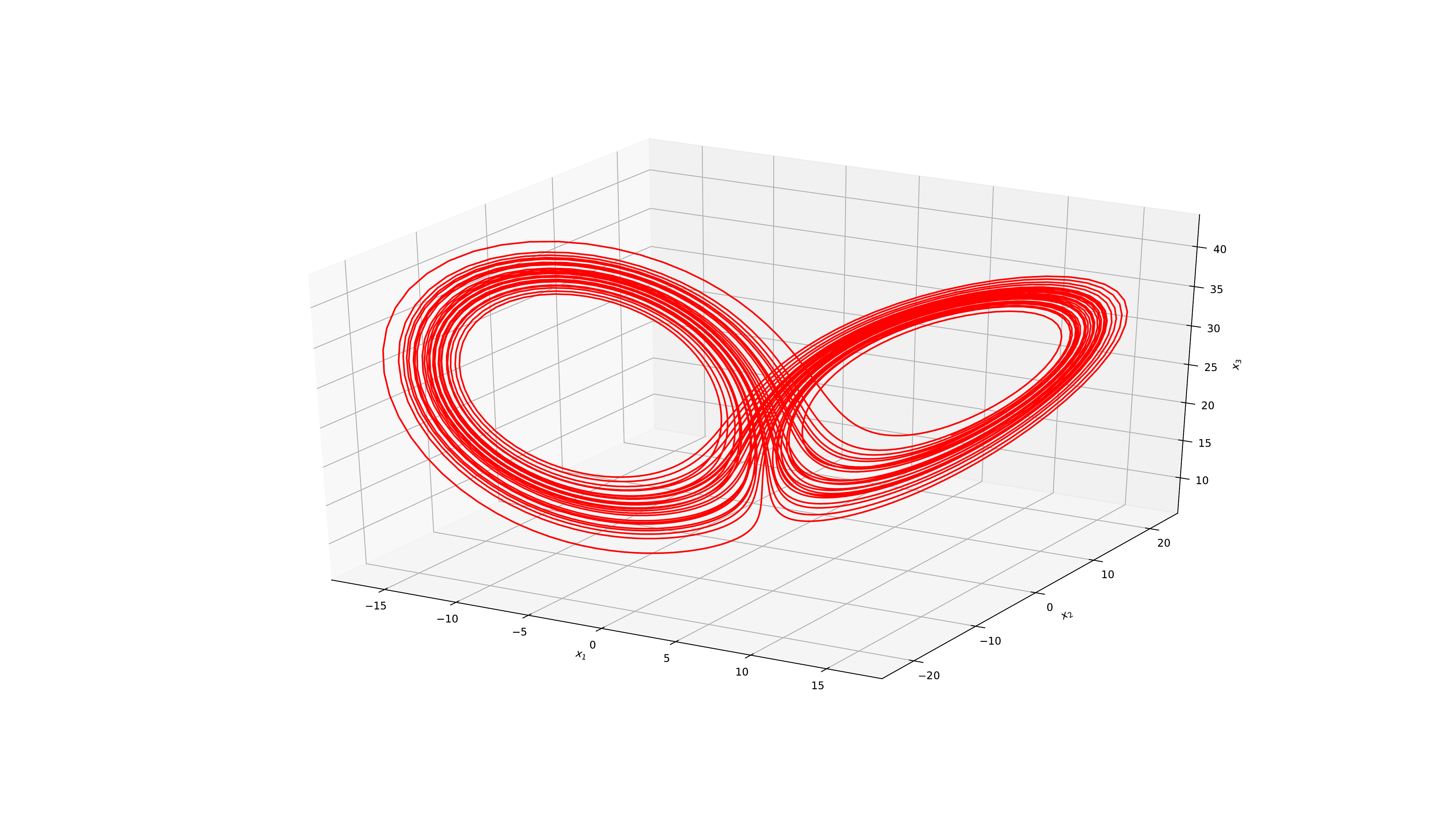}
		\includegraphics[width=\bwidth,clip, trim=100mm 40mm 80mm 40mm]{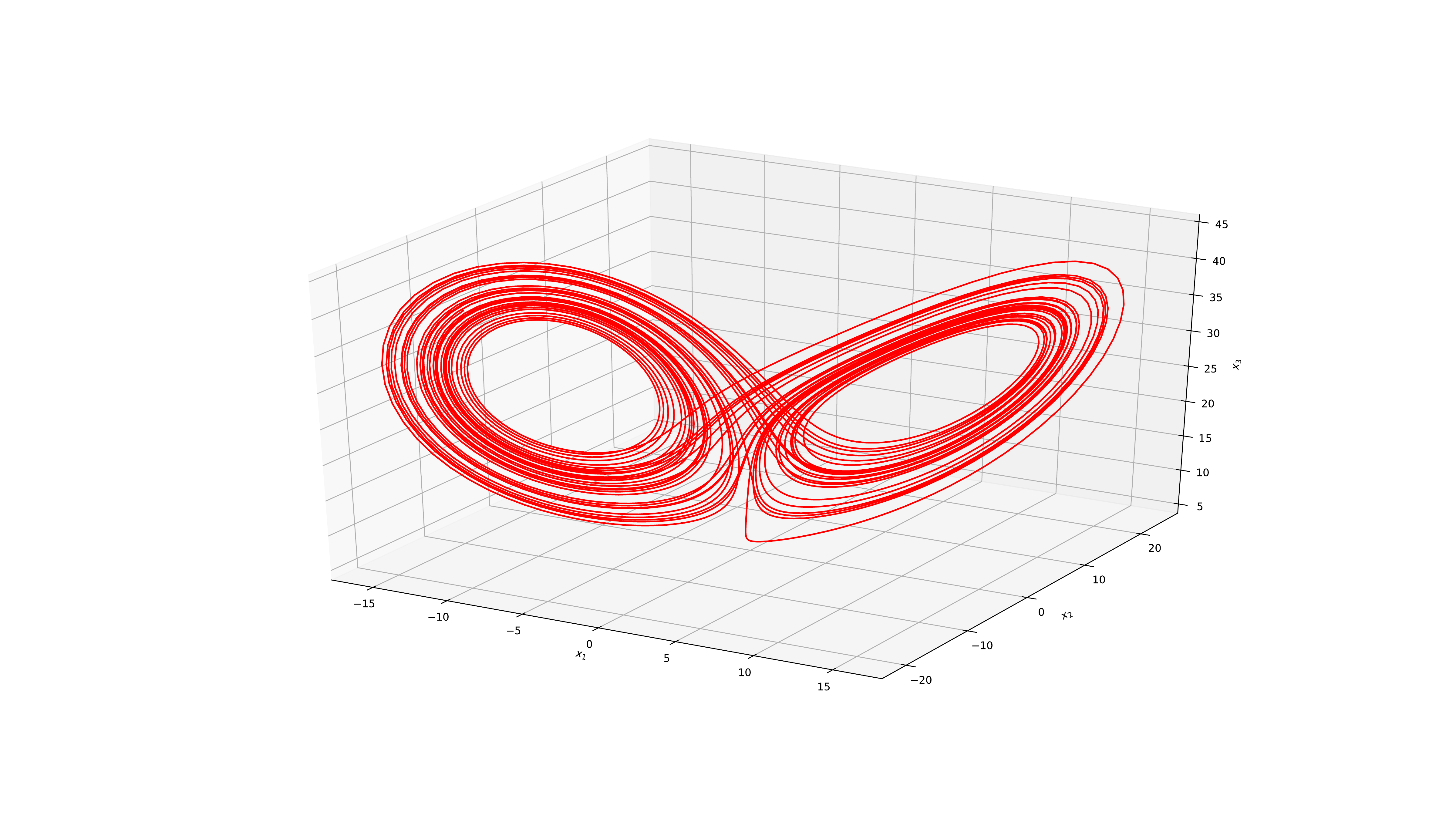}
		\includegraphics[width=\bwidth,clip, trim=100mm 40mm 80mm 40mm]{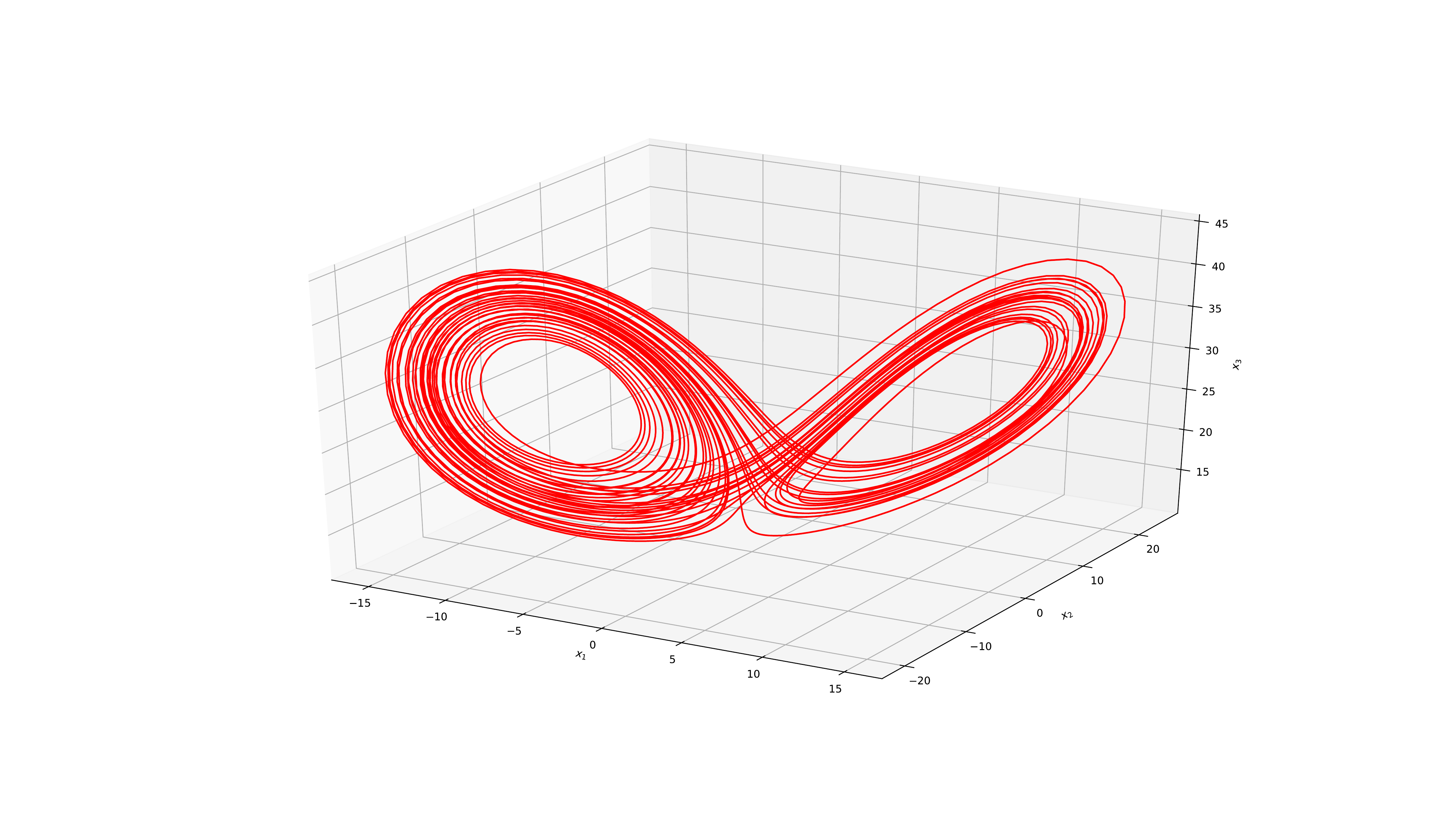}
		\includegraphics[width=\bwidth,clip, trim=100mm 40mm 80mm 40mm]{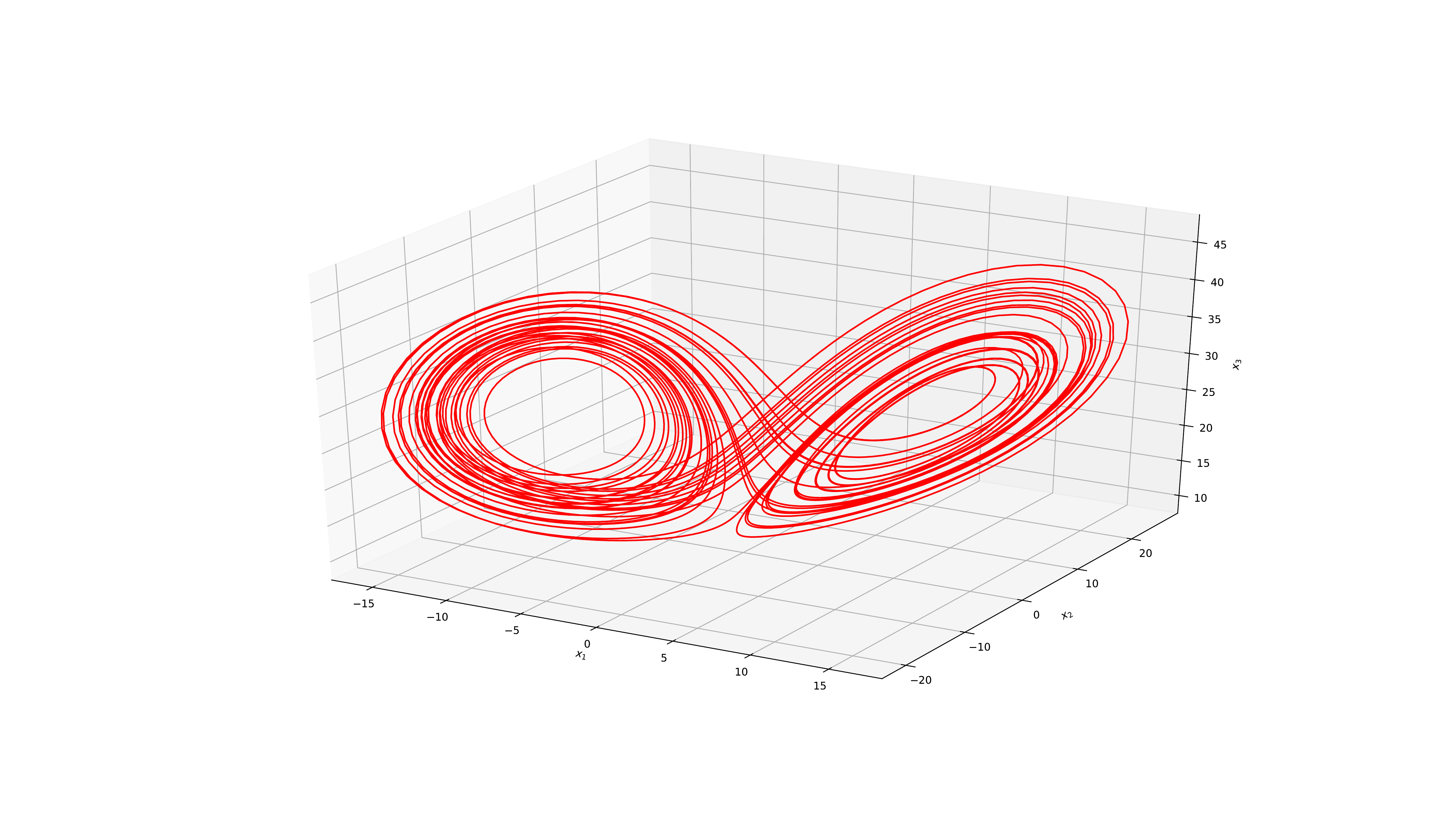}
		\includegraphics[width=\bwidth,clip, trim=100mm 40mm 80mm 40mm]{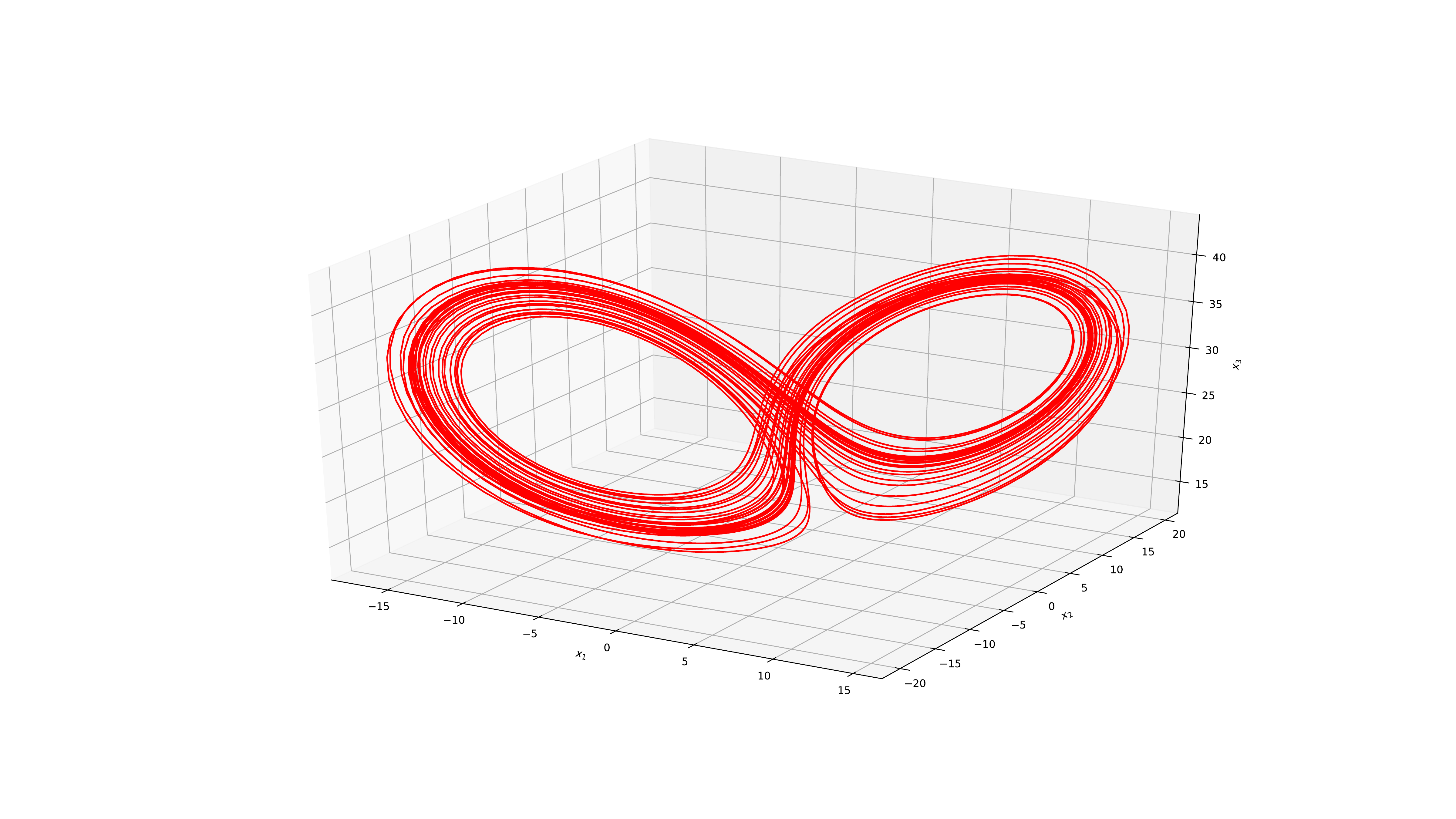}
	\end{subfigure}%
	
	\begin{subfigure}[b]{0.04\linewidth}
	    \rotatebox[origin=t]{90}{\scriptsize VODEN\_S2}\vspace{0.4\linewidth}
	\end{subfigure}%
	\begin{subfigure}[t]{0.96\linewidth}
		\centering
		\includegraphics[width=\bwidth,clip, trim=100mm 40mm 80mm 40mm]{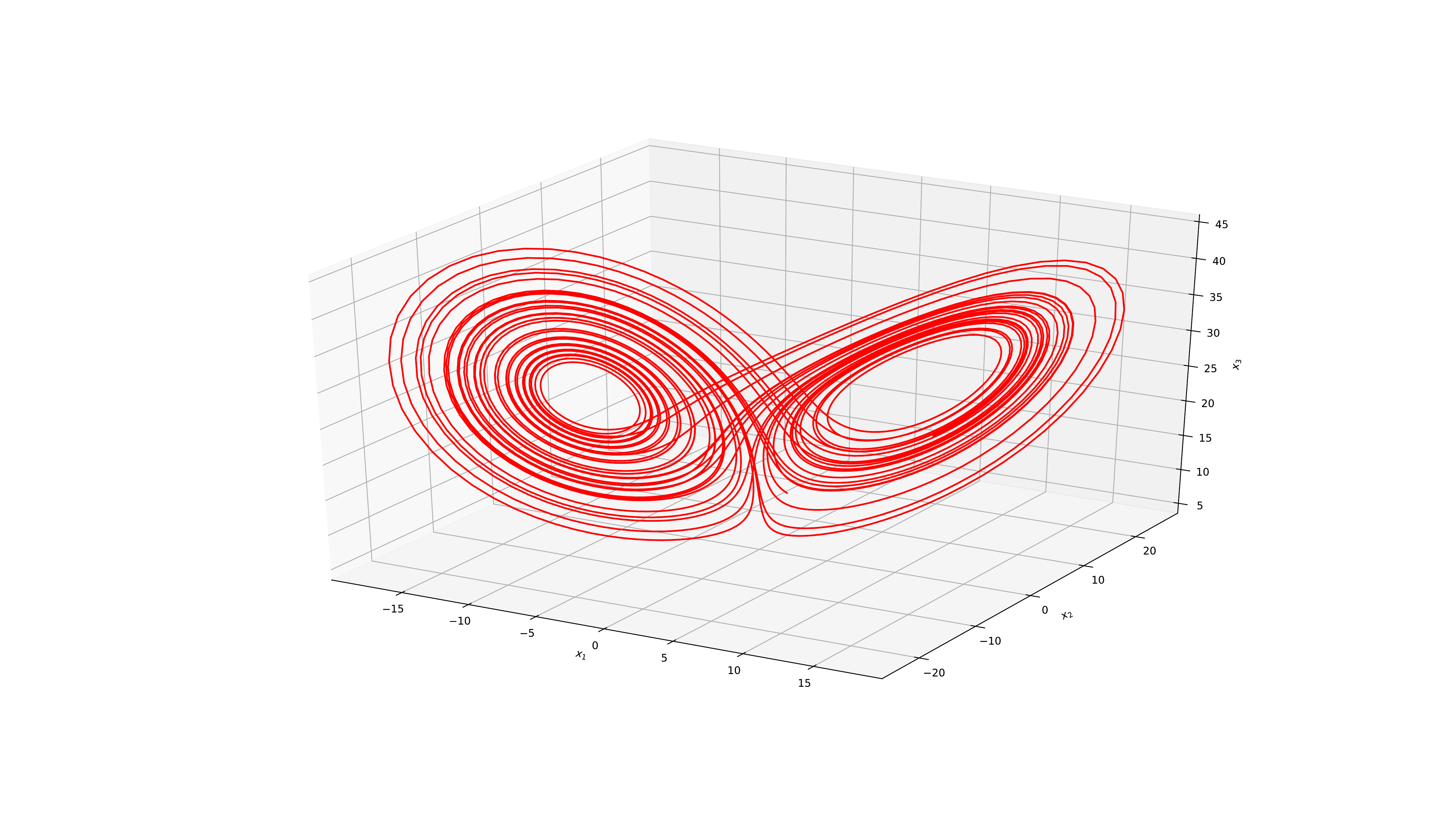}
		\includegraphics[width=\bwidth,clip, trim=100mm 40mm 80mm 40mm]{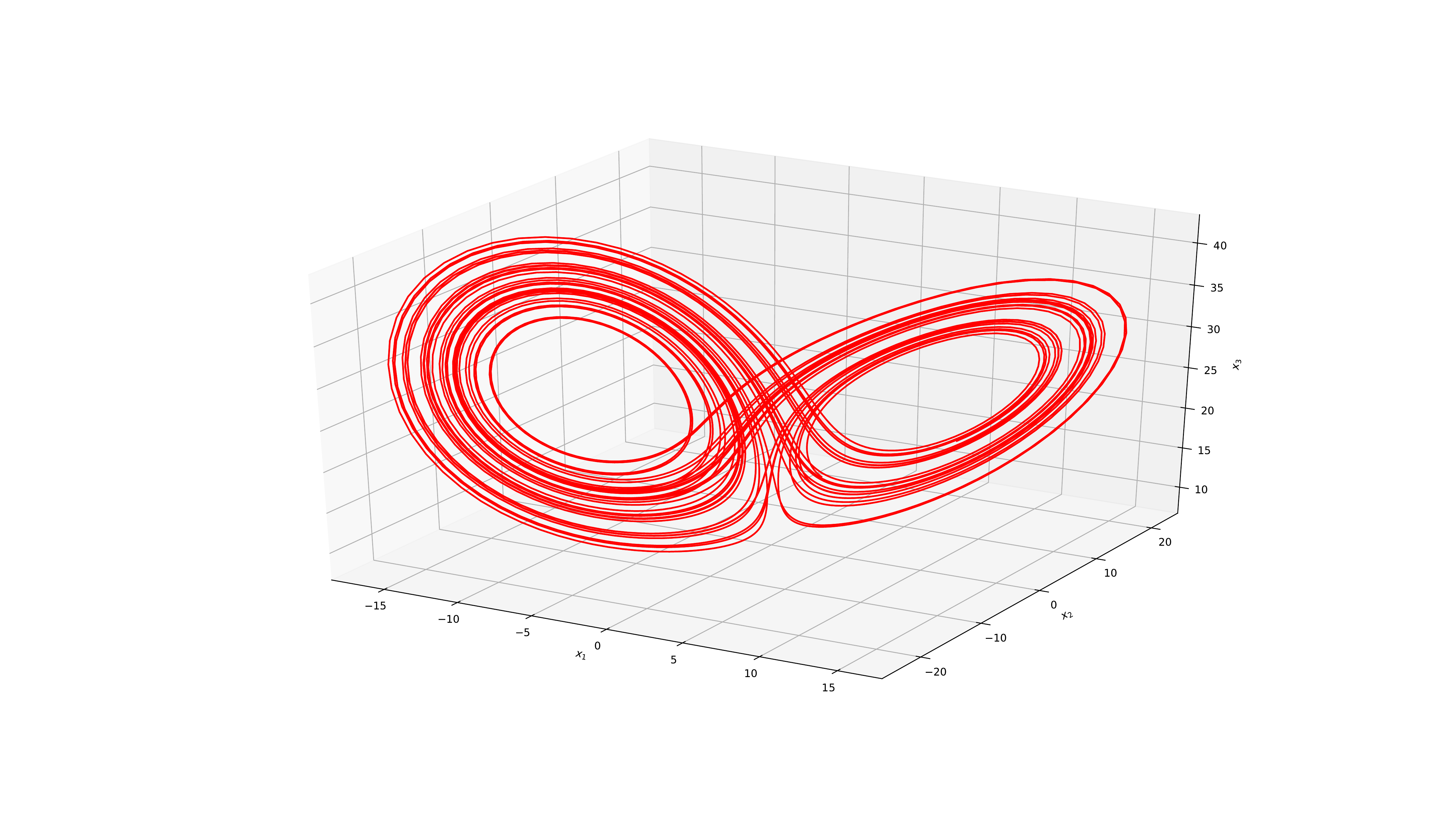}
		\includegraphics[width=\bwidth,clip, trim=100mm 40mm 80mm 40mm]{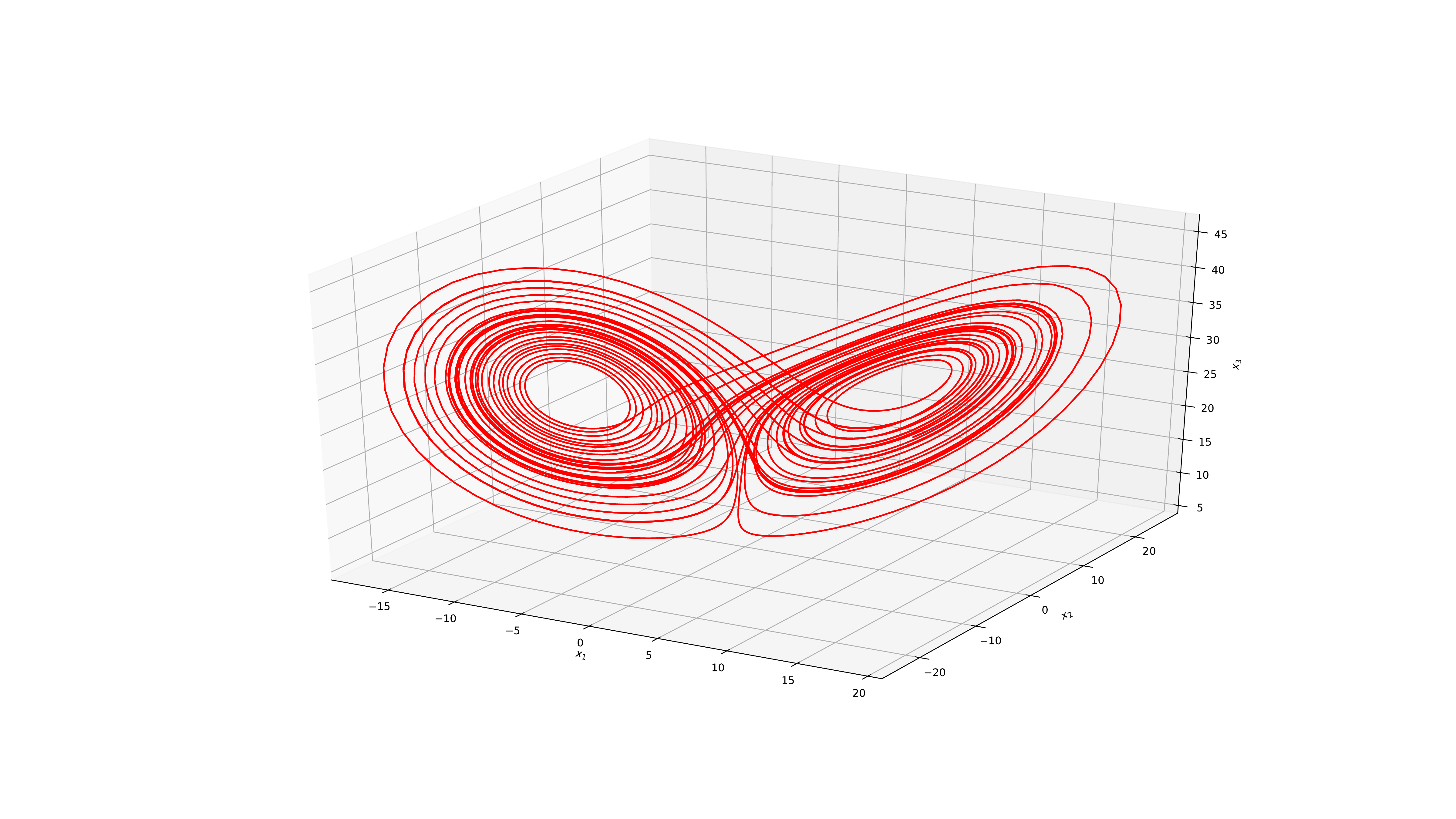}
		\includegraphics[width=\bwidth,clip, trim=100mm 40mm 80mm 40mm]{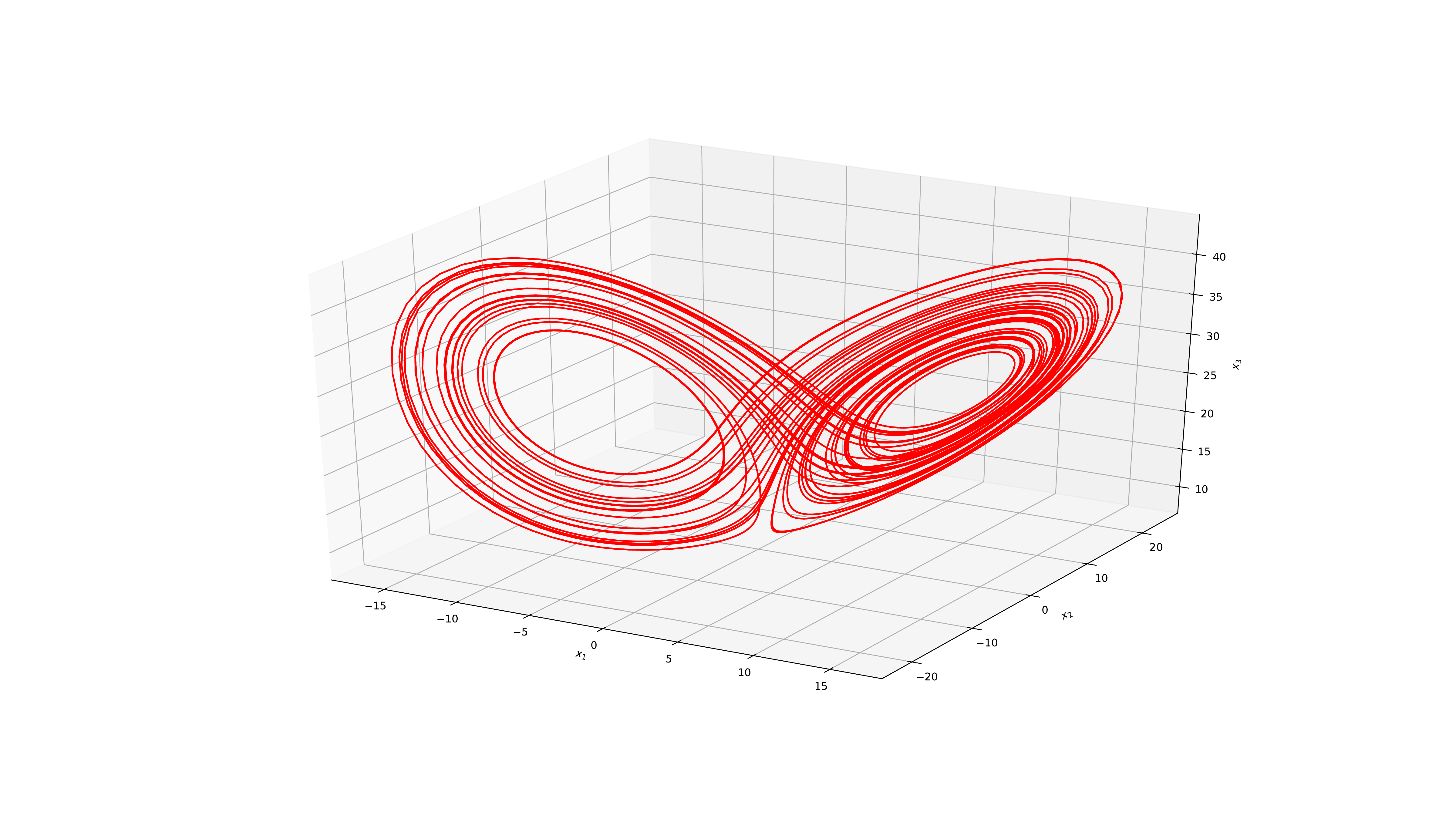}
		\includegraphics[width=\bwidth,clip, trim=100mm 40mm 80mm 40mm]{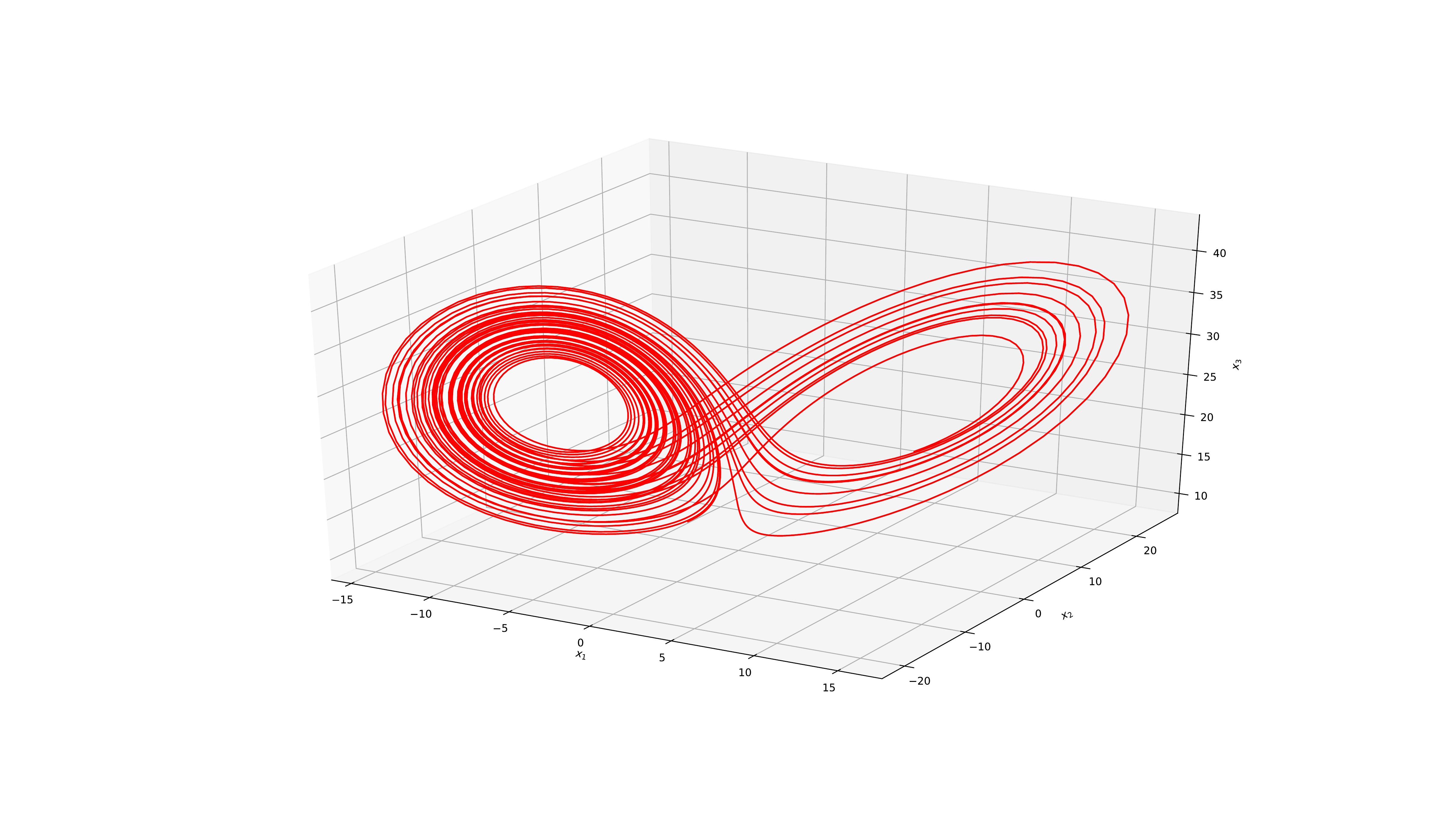}
		\includegraphics[width=\bwidth,clip, trim=100mm 40mm 80mm 40mm]{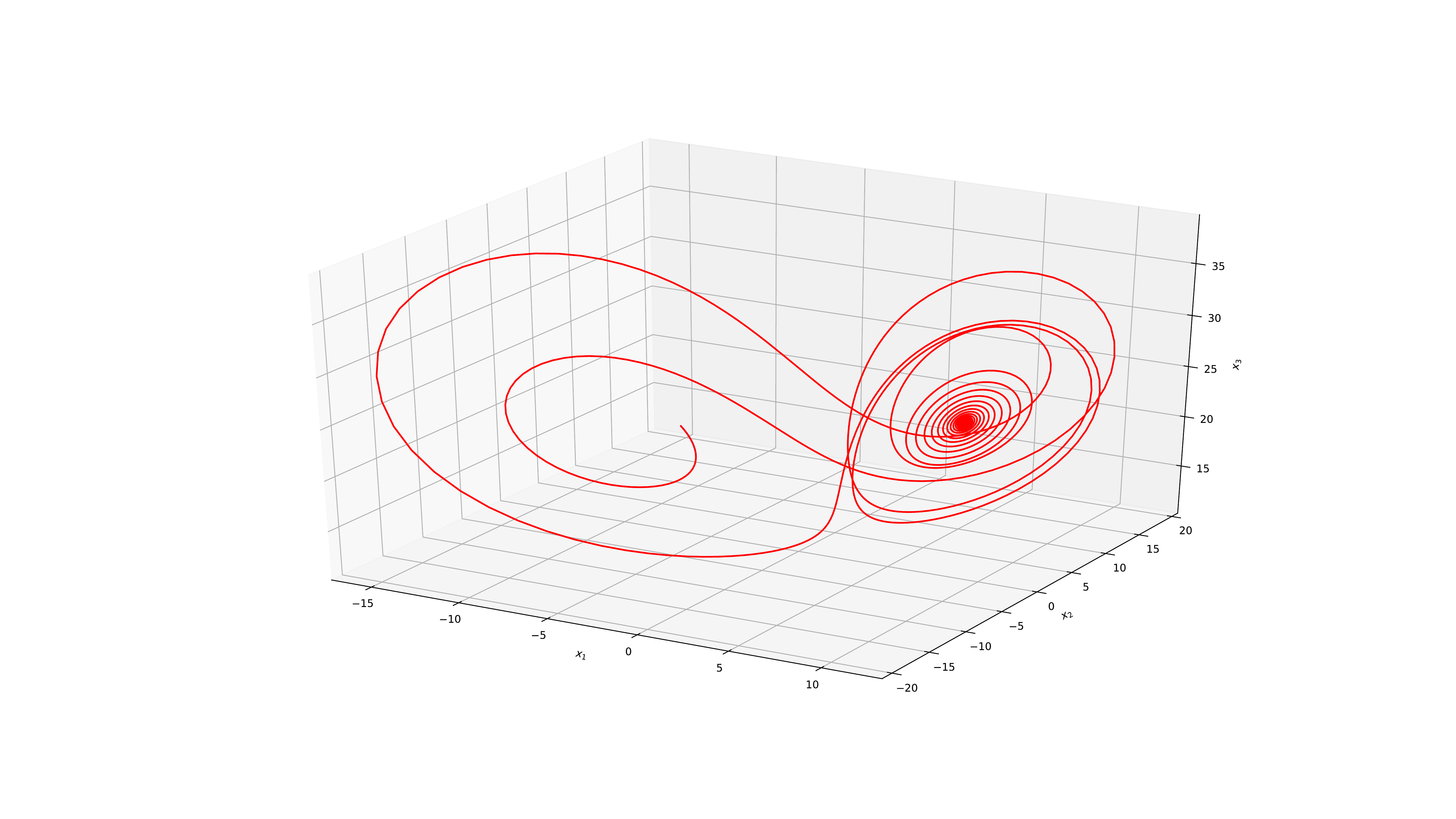}
	\end{subfigure}%
	
	
	\caption{Attractors generated by models trained on partially observed data scenario 1 and 2.}
    \label{fig:attactors_lorenz63_partial}
\end{figure}

We kept the same settings that used in Section \ref{sec:lorenz63_noisy}. The results are shown in Table \ref{tab:lorenz63_partial}, Fig. \ref{fig:seq_comparision_voden_lorenz63_s2_0.5} and Fig. \ref{fig:attactors_lorenz63_partial}. For all cases, the performance of both models was decreased. This again highlights the importance of the sampling frequency for learning dynamical systems. As expected, the EnKS-EM works better on irregular data, because EnKS naturally provides a straightforward tool to deal with irregularity.

\begin{table}
    \caption {Short-term forecasting error and very-long-term forecasting topology of data-driven models learned on noisy and partial Lorenz-63 data.}
    \label{tab:lorenz63_partial}

    \centering
    \begin{tabular}{ll*{6}c}
    \toprule
    \multicolumn{2}{c}{\multirow{2}{*}{Model}} & \multicolumn{6}{c}{$\sigma^2$} \\
    & & 0.5 & 2 & 4 & 8 & 16  & 32 \\
    \midrule \midrule 
    \multirow{3}{*}{EnKS-EM\_s1}
    & $t_0+h$        & 0.005 & 0.013 & 0.015 & 0.016 & 0.061 & 0.067 \\
    & $t_0+4h$       & 0.015 & 0.038 & 0.050 & 0.066 & 0.183 &0.186\\ 
    &  $\lambda_1$   & 0.903 & 0.896 & 0.884 & 0.744 & 0.691 &0.894\\
    \midrule
    \multirow{3}{*}{VODEN\_s1}
    & $t_0+h$           & 0.012 & 0.008 & 0.008 & 0.032 & 0.058 & 0.044 \\
    & $t_0+4h$          & 0.035 & 0.028 & 0.022 & 0.103 & 0.178 & 0.137 \\ 
    &  $\lambda_1$      & 0.862 & 0.848 & 0.877 & 0.705 & 0.772 & 0.648 \\
    \midrule
    \multirow{3}{*}{EnKS-EM\_s2}
    & $t_0+h$        & 0.022 & 0.023 & 0.033 & 0.070 & 0.031 & 0.060\\
    & $t_0+4h$       & 0.065 & 0.075 & 0.101 & 0.156 & 0.072 & 0.149\\ 
    &  $\lambda_1$   & 0.894 & 0.758 & 0.803 & 0.475 & 0.905 & 0.658\\
    \midrule
    \multirow{2}{*}{VODEN\_s2}
    & $t_0+h$           & 0.038 & 0.015 & 0.039 & 0.049 & 0.058 & 0.115\\
    & $t_0+4h$          & 0.115 & 0.047 & 0.132 & 0.138 & 0.174 &
    0.317\\ 
    &  $\lambda_1$      & 0.894 & 0.869 & 0.916 & 0.696 & 0.705 &
    0.164\\
    \bottomrule
    \end{tabular}

\end{table}

\section{Related work and Discussion}
\label{sec:related_work}

Learning dynamical systems is a topic that has been studied for decades. Before the era of deep learning, almost all methods used EM-like iterative learning procedures \cite{ghahramani_parameter_1996, ghahramani_learning_1999, voss_nonlinear_2004}. These models use one method in the family of Kalman-based data assimilation schemes in the E-step to estimate $p_\theta(\vect{x}_t|\vect{y}_{1:T})$. The maximization in the M step is also performed analytically. However, to do so, these methods can use only simple distributions and processes whose analytic form is known. Therefore, the systems that can be learned are very restricted.

Recently, the influence of deep learning \cite{lecun_deep_2015} has spread to every domain, including model identification. \cite{raissi_multistep_2018} used DenseNet, \cite{yeo_deep_2019} used LSTM, \cite{qin_data_2018} used ResNet to identify the nonlinear dynamical systems by minimizing the short-term prediction error. They try to exploit the power of neural networks to overcome the difficulties of modelling the nonlinearities. The problems of these methods appear when the observations are partial, irregular or when a high level of noise is present. Neural networks, in general, do not have an efficient way to deal with irregularity. When the observations are highly damaged by noise, using the short-term prediction error as the objective function would very likely to make the network overfit the data. These methods can be considered as a specific case of our methodology, when $q_\theta(\vect{x}_t|\vect{y}_{1:T})$ collapse to $y_t$. 

Two special methods that do not fall into the two classes above is the Analog Forecasting (AF) \cite{lguensat_analog_2017} and the Sparse Regression (SR) \cite{brunton_discovering_2016}. The analog forecasting is a non-parametric model that "learns by heart" the dynamics in the catalog. For each new observation $\vect{y}_t$, AF looks up its catalog and find the most similar points. It then predicts the next observation $\vect{y}_{t+1}$ by averaging the evolution of these points in the catalog. Since AF is a k-NN based method, it does not work well in high-dimensional spaces. The sparse regression finds the analytic form of the dynamics by performing a regression on a basis formed by many possible functions of each component of the state. This method works extremely well when the observations are complete and clean. When the observation is noisy, partial or irregular, SF fails.

Our proposed EM-like methodology unifies the classical data assimilation schemes and the recent advances in neural networks for the problem of learning non-linear dynamical systems. Different communities may value our contributions for different aspects. For the data assimilation community, we introduce neural networks as a means to go beyond the limit of using analytic functions/processes, such as the nonlinearity. For deep learning practitioners, we has shown that the classical EM procedure might be a way to handle noise and irregularity. 

This framework may be applied to improve the modeling capacity in numerous fields. For example, satellite data are noisy (interfered by unknown factors in the atmosphere) and irregular (constrained by the revisit time of the satellite). We demonstrated here two models that follow this methodology---the EnKS-EM and the VODEN---on the Lorenz system. In general, the methodology is expected to improve the performance of any existing dynamical learning model, by using an appropriate inference scheme. 

A number of open problems remain for future work. The two simplifications could be relaxed for a better objective function. We used simple linear interpolation for the partial cases, but a more sophisticated interpolation clearly might be applied. How to learn a dynamical system whose latent states have some components that have never been observed is still a challenge.

\section{Acknowledgments}

This work was supported by GERONIMO project (ANR-13-JS03-0002), Labex Cominlabs (grant SEACS), Region Bretagne, CNES (grant OSTST-MANATEE), Microsoft (AI EU Ocean awards) and by MESR, FEDER, Région Bretagne, Conseil Général du Finistère, Brest Métropole and Institut Mines Télécom in the framework of the VIGISAT program managed by "Groupement Bretagne Télédétection" (BreTel).

\bibliographystyle{IEEEtran}
\bibliography{Zotero}
\end{document}